\documentclass[dvipsnames]{article}

\usepackage[final]{neurips_2021}

\usepackage[utf8]{inputenc} %
\usepackage[T1]{fontenc}    %
\usepackage{url}            %
\usepackage{booktabs}       %
\usepackage{amsfonts}       %
\usepackage{nicefrac}       %
\usepackage{booktabs} 
\usepackage{multirow}
\usepackage{wrapfig}
\usepackage{amssymb}
\usepackage{epigraph}
\usepackage{makecell}

\usepackage[labelfont=bf]{caption} %

\usepackage[ruled,vlined]{algorithm2e}
\usepackage{color, soul}
\usepackage{microtype}      %
\usepackage{xcolor}         %
\usepackage{geometry}
\usepackage{times}
\usepackage{ragged2e}
\usepackage{amsmath}
\usepackage{bm}
\usepackage{latexsym}
\usepackage{subfigure}
\usepackage{graphicx}
\usepackage{float}
\usepackage{longtable}
\usepackage{booktabs} 
\usepackage{multirow}
\usepackage{amssymb}
\usepackage{longtable}
\usepackage{tabu}
\usepackage{bbding}
\usepackage{awesomebox} %
\usepackage{bbding}
\usepackage[most]{tcolorbox}
\usepackage{etoolbox}
\usepackage{fancyhdr}

\usepackage{colortbl}
\usepackage{xcolor}
\usepackage{tikz}

\usepackage{caption} 

\usepackage[colorlinks, linkcolor=RoyalBlue, anchorcolor=BrickRed, citecolor=RoyalBlue, urlcolor=RoyalBlue]{hyperref}

\usepackage{cleveref}
\crefname{section}{§}{§§}
\Crefname{section}{§}{§§}

\newcommand\refsec[1]{\hyperref[sec:#1]{§\ref{sec:#1}:~\textsc{#1}}}
\newcommand\refsecs[2]{\hyperref[sec:#1]{§\ref{sec:#1}:~\textsc{#1}}, \hyperref[sec:#2]{§\ref{sec:#2}:~\textsc{#2}}}

\newtheorem{thm}{Theorem}[section]

\newtheorem{definition}[thm]{Definition}
\newtcolorbox{myboxnote}[1][]{
  breakable,
  title=#1,
  colback=cyan!0,
  colbacktitle=cyan!0,
  coltitle=black,
  fonttitle=\bfseries,
  bottomrule=0pt,
  toprule=0pt,
  leftrule=1.5pt,
  rightrule=1.5pt,
  titlerule=0pt,
  arc=0pt,
  outer arc=0pt,
  colframe=lightgray,
}
\newenvironment{itemize*}%
 {\leftmargini=20pt\begin{itemize}%
  \setlength{\itemsep}{3pt}%
  \setlength{\parskip}{0pt}%
  }%
 {\end{itemize}}
\newenvironment{enumerate*}%
 {\begin{enumerate}%
  \setlength{\itemsep}{0pt}%
  \setlength{\parskip}{0pt}}%
 {\end{enumerate}}

\usepackage{fancyhdr} %
\usepackage{blindtext} %

\title{Evaluating Intelligence via Trial and Error}

\author{%
  Jingtao Zhan$^{1}$, Jiahao Zhao$^{2}$, Jiayu Li$^{1}$, Yiqun Liu$^{1}$\thanks{Corresponding author.} , Bo Zhang$^{1}$, \\ 
  \textbf{\vspace{0.1cm} Qingyao Ai$^{1}$, Jiaxin Mao$^{2}$, Hongning Wang$^{1}$, Min Zhang$^{1}$, Shaoping Ma$^{1}$} \\
   {\vspace{0.1cm} $^{1}$Tsinghua University, $^{2}$Renmin University of China} \\ 
 {\texttt{\{zhanjt20@mails., yiqunliu@\}tsinghua.edu.cn}
 }
}

\begin{document}

\maketitle

\begin{abstract}
How does intelligence emerge? We propose that intelligence is not a sudden gift or random occurrence, but rather a crucial trait for species to survive through Natural Selection. Natural Selection requires a species to find solutions within a limited number of trial-and-error attempts. Building on this idea, we introduce \textit{Survival Game} as a framework to evaluate intelligence based on the number of failed attempts in a trial-and-error process. Fewer failures indicate higher intelligence. When the expectation and variance of failure counts are both finite, it signals the ability to consistently find solutions to new challenges, which we define as the Autonomous Level of intelligence. Using \textit{Survival Game}, we comprehensively evaluate existing artificial intelligence (AI) systems. Our results show that while AI systems achieve the Autonomous Level in simple tasks, they are still far from it in more complex tasks, such as vision, search, recommendation, and language. While scaling current AI technologies might help, this would come at an astronomical cost. Projections suggest that achieving the Autonomous Level for general tasks would require $10^{26}$ parameters. To put this into perspective, loading such a massive model requires so many H100 GPUs that their total value is $4 \times 10^{7}$ times that of Apple Inc.'s market value. Even with Moore's Law, supporting such a parameter scale would take $70$ years. This staggering cost highlights the complexity of human tasks and the inadequacies of current AI technologies. To further investigate this phenomenon, we conduct a theoretical analysis of \textit{Survival Game} and its experimental results. Our findings suggest that human tasks possess a ``criticality'' property. As a result, Autonomous Level requires a deep understanding of the task's underlying mechanisms. Current AI systems, however, do not fully grasp these mechanisms and instead rely on superficial mimicry, making it difficult for them to reach an autonomous level. We believe \textit{Survival Game} can not only guide the future development of AI but also offer profound insights into human intelligence.
\end{abstract}

\vspace{0.5cm}
\hspace{130pt}\parbox[b]{0.6\textwidth}
{
\epigraph{\textit{``... But Natural Selection, as we shall hereafter see, is a power incessantly ready for action, and is as immeasurably superior to man's feeble efforts, as the works of Nature are to those of Art. ...''}}{--- Charles Darwin, \textit{Origin of Species}, 1859.}
}

\newpage
{
  \hypersetup{linkcolor=RoyalBlue, linktoc=page}
  \renewcommand{\baselinestretch}{0.9}  %
  \tableofcontents
}

\newpage

\section{Introduction}

\textit{How does intelligence emerge?} We believe that intelligence is not an innate gift but rather a necessity shaped by \textit{Natural Selection}. The diversity of life forms we see today, including humans, animals, and plants, has addressed countless challenges imposed by the natural world.
Natural Selection can be viewed as a trial-and-error process, where subjects shall persistently explore, seek solutions in the face of uncertainty, and ultimately prevail.  
The subjects must strive to solve the challenges, experimenting again and again until they succeed. If they cannot find a solution, they fail the test and thus do not survive.

Inspired by Natural Selection, we propose \textit{Survival Game} to evaluate intelligence. Similar to how species find a way to survive through trial and error in Natural Selection, \textit{Survival Game} evaluates intelligence by counting the number of failures before finding correct solutions in a trial-and-error process. Fewer failures correspond to higher intelligence. The number of failures is a discrete random variable, and smaller expectations and variances of the failure count indicate higher intelligence.
If expectations and variances are infinite, the subjects can never find the correct solutions and thus do not survive in the game.
Based on the convergence of the expectations and variances, \textit{Survival Game} divides intelligence into three levels: Limited, Capable, and Autonomous. If both the expectation and variance diverge, the subject is at the Limited Level. At this level, the subject is comparable to blindly enumerating possible solutions. If both the expectation and variance converge, the subject reaches the Autonomous Level. At this level, the subject can stably find the correct solution with only a few trials, thereby being able to autonomously operate at an affordable cost.
As we can see, the results of the \textit{Survival Game} have clear physical meaning about the subject's intelligence level.

The \textit{Survival Game} can be applied to any task and any species. In this paper, we are particularly interested in artificial intelligence~(AI) systems. Therefore, we conduct \textit{Survival Game} on state-of-the-art AI systems available today. The results demonstrate that a system with better modeling of the task can reach a higher level of intelligence. Current AI technologies can reach the Autonomous Level on simple tasks like handwritten digit recognition. However, they are mostly at Limited Level on more complex tasks, including vision, search, recommendation, and language. This indicates that most AI systems are at a preliminary stage: they are unable to substantially narrow down the range of possible answers and their performance is comparable to brute-force enumeration. This indicates that directly applying these AI technologies can result in very high costs and serious errors, so they cannot operate autonomously, and human supervision is essential. 
These findings challenge conclusions from previous studies~\citep{biever2023chatgpt, aharoni2024attributions, mei2024turing}, which suggest that AI has already reached a very high level of intelligence.

\begin{figure}[h]
    \subcapraggedrighttrue
    \subcaphangtrue
        \centering
        \subfigure{\includegraphics[width=0.49\textwidth]{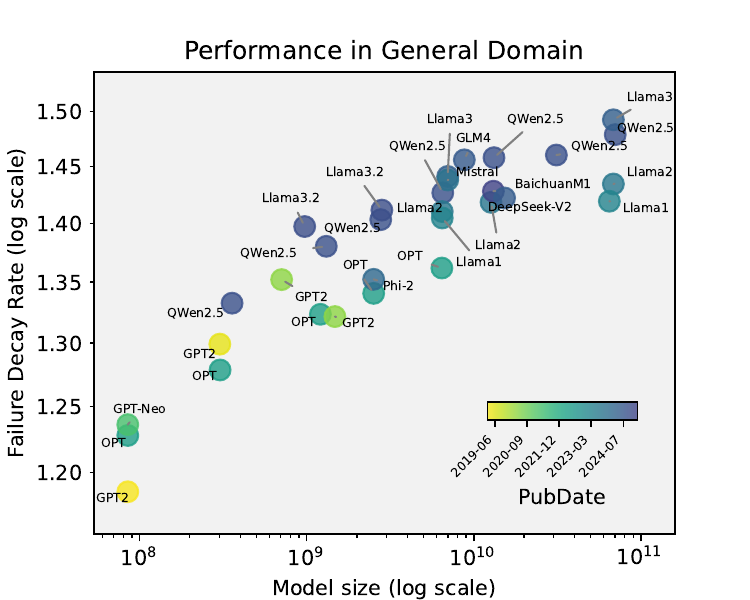}} \hfill
        \subfigure{\includegraphics[width=0.49\textwidth]{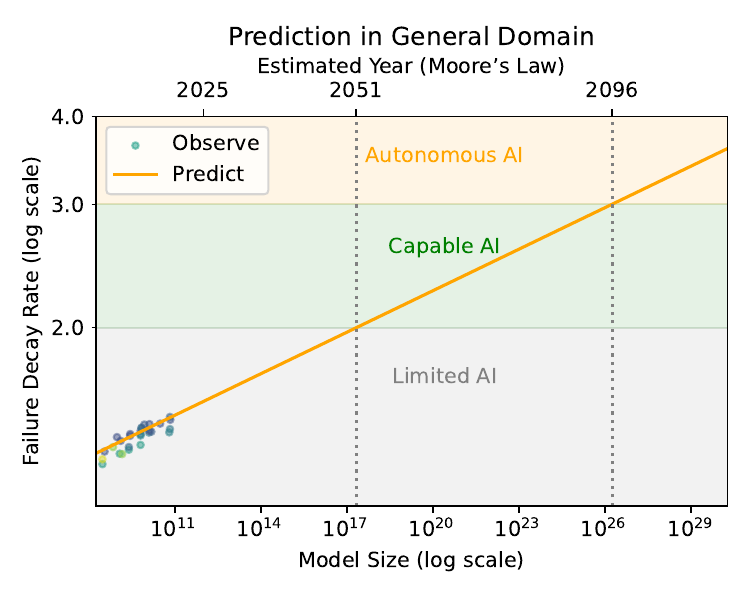}}
        \caption{Experimental Results of \textit{Survival Game} in General Domain. Left: Results suggest that larger models achieve better performance.
         Right: Results suggest that achieving Autonomous-Level Intelligence requires an unimaginable parameter scale.}
        \label{fig:full_perf}
    \end{figure}

In \textit{Survival Game}, the intelligence score exhibits a log-linear relationship with the scale of AI systems. If we assume this relationship continues to hold, we can predict the scale required to achieve Autonomous-Level intelligence, as shown in Figure~\ref{fig:full_perf}. The projection suggests that, for general language tasks, an AI system would need a parameter size of $10^{26}$ to reach the Autonomous Level.
To put this scale into perspective, this is equivalent to $10^5$ times the total number of neurons in all of humanity’s brains combined. Loading a model of this size onto H100 GPUs would necessitate $5 \times 10^{15}$ H100 cards, a cost equivalent to $4 \times 10^7$ times the market value of Apple Inc. If hardware development continues to follow Moore's Law, it would take 70 years of progress to support the development of such a large model.
These results suggest that attempting to solve human tasks with current AI technology is extremely difficult, if not impossible.

Why is the Autonomous Level so difficult to achieve for current AI systems? We conduct a theoretical analysis and demonstrate that the root cause lies in the complexity of human tasks and the inadequacies of current AI technologies. Specifically, we leverage self-organized criticality (SOC)~\citep{bak2013nature} to analyze \textit{Survival Game}. Results suggest that many human tasks exhibit a criticality property: even slight changes in the environment require entirely different responses. To successfully operate these tasks, it is important to fully understand their mechanisms. However, current AI systems do not fully grasp this complex mechanism and instead leverage superficial imitation: They memorize answers to some questions and attempt to solve new questions through exploration. Although scaling AI systems can make the exploration more effective and improve imitation performance, a lack of a full understanding of the underlying mechanism results in unimaginable costs.

The structure of this paper is as follows. In Section~\ref{sec:related_work}, we review related works on intelligence evaluation to provide a broad context for our method. In Section~\ref{sec:define_intelligence_test}, we present \textit{Survival Game} and show how it measures intelligence and categorizes it into three levels. In Section~\ref{sec:evaluate_current_ai}, we extensively evaluate existing AI systems using \textit{Survival Game}. Section~\ref{sec:future_prediction} empirically explores how scaling improves Intelligence. In Section~\ref{sec:theory}, we provide a theoretical analysis of \textit{Survival Game} to gain a deep understanding of the nature of human tasks and current AI. Finally, in Section~\ref{sec:conclusion}, we conclude the paper and outline potential directions for future research.

\section{Related Work}
\label{sec:related_work}

Defining a test for intelligence is a fundamental issue. For AI research, it allows us to understand, apply, and develop AI technology. More broadly, it enables us to gain a deep understanding of intelligence, leading to a profound insight into both humanity and the natural world.

In 1950, Alan Turing proposed the Imitation Game to test whether a machine can possess human intelligence. Since then, it has been highly influential in the AI field. Many researchers have developed methods to practically implement or further improve Imitation Game. In this section, we review some of the most influential approaches. 

\begin{itemize}
	\item Imitation Game, aka Turing Test~\citep{turing1950computing}: Intelligence is the ability to imitate human responses convincingly in a text-based conversation. If a human evaluator cannot reliably distinguish between a machine and a human based on their answers, the machine is considered intelligent.
	\item Total Turing Test~\citep{harnad1991other}: It is an extended version of the Turing Test that assesses a machine's ability to interact with the world in a human-like way. It goes beyond text-based conversations to include physical interaction and sensory perception.
	\item Chinese Room Argument~\citep{searle1999chinese}: It argues that the Imitation Game only evaluates syntactics and yet AI should also understand semantics, such as knowing the actual meaning of each word.
	\item Lovelace Test~\citep{bringsjord2003creativity}: It argues that intelligence is about creativity. For example, AI should be able to originate art, music, or poetry.
	\item Reverse Turing Test~\citep{baird2003pessimalprint}: Instead of asking whether a machine can act like a human, it asks whether an AI can differentiate between humans and machines. 
	\item Universal Intelligence~\citep{legg2007universal}: Beyond the conversation task in Imitation Game, it measures an agent’s ability to achieve goals in a wide range of environments.
	\item Winograd Schema Challenge~\citep{levesque2012winograd}: It tests whether AI can identify the antecedent of an ambiguous pronoun in a statement. It requires world knowledge and contextual understanding. 
	\item General intelligence~\citep{goertzel}: It defines intelligence as the ability to achieve a wide range of goals and handle new problems in different contexts and environments.
	\item Visual Turing Test~\citep{geman2015visual}: It adds the visual understanding ability to the Imitation Game. It tests whether AI can answer complex questions about images.
	\item Economical Value~\citep{openAICharter}: It tests whether AI can be a highly autonomous system that outperforms humans at most economically valuable work.
	\item The Modern Turing Test~\citep{suleyman2023coming}: It argues intelligence is to make a meaningful impact to the real world. It tests whether AI can make \$1 million on a retail web platform in a few months with just a \$100,000 investment.
	\item Outperforming Humans~\citep{morris2024levels}: It defines different levels of intelligence by how many humans AI can outperform. For example, a Competent AI outperforms $50\%$ skilled adults and a virtuoso AI outperforms $99\%$ skilled adults.
\end{itemize}

We observe that almost all previous works attempted to define intelligence by determining which human-like tasks that a machine must accomplish in order to be considered intelligent. However, there is significant variation in the choice of the tasks, as different researchers hold different perspectives on what constitutes intelligence. We can see that these approach approaches are inherently \textit{subjective}, which manifests in three key ways:
\begin{itemize}
	\item \textit{Subjective} (Human-Centric) View of Intelligence: Many of these tests utilize human intelligence as an upper bound for AI and evaluate whether AI can approach this bound. For example, Imitation Game~\citep{turing1950computing} evaluates a machine’s ability to replicate human behavior; \cite{openAICharter} defines intelligence as outperforming humans at economically valuable work; \cite{morris2024levels} defines intelligence level by how many humans AI can outperform. Nevertheless, if AI surpasses humans in certain tasks, these evaluation methods are no longer applicable. 
	\item \textit{Subjective} Choice of Tasks: These researchers believe that intelligence is a general property rather than something tied to specific tasks. Researchers have sought to define tasks that best reflect intelligence, making these tasks increasingly complex to measure ever more sophisticated forms of intelligence. However, this approach is inherently subjective: different researchers emphasize different aspects of intelligence, preventing consensus. For instance, \citet{harnad1991other} chooses physical tasks; \citet{bringsjord2003creativity} argues creative tasks; \citet{suleyman2023coming} adopts economic tasks; \citet{morris2024levels} suggest cognitive tasks. The belief that intelligence is independent of tasks, yet simultaneously trying to define it through a single universal task, leads to contradictions.
	\item \textit{Subjective} Evaluation Framework: These tests rely heavily on subjective measures of how well an AI system imitates human behavior. However, defining what constitutes ``good imitation'' and the threshold at which intelligence emerges is highly ambiguous. For example, Winograd Schema Challenge~\citep{levesque2012winograd} is considered defeated because AI achieved $90\%$ accuracy~\citep{kocijan2023defeat}; Imitation Game~\citep{turing1950computing} is considered defeated because current chatbot successfully fooled human evaluators $40\%$ of the time~\citep{biever2023chatgpt}; Modern Turing Test~\citep{suleyman2023coming} will be defeated if AI makes \$1 million. However, these thresholds are not well-defined and may differ among researchers. Since there is no universally accepted standard, the conclusions will be inconsistent. This defect makes it difficult to translate these tests into reliable evaluation methods for real-world AI applications.
\end{itemize}

In contrast, our proposed \textit{Survival Game} has a clear physical meaning and a well-defined statistical basis. It is inherently an \textit{objective} way to evaluate intelligence. Before we further elaborate on the differences between \textit{Survival Game} and related studies, we will first introduce \textit{Survival Game} in Section~\ref{sec:define_intelligence_test} and then continue the comparison in Section~\ref{sec:comparison_with_related_work}.

\section{Methodology}
\label{sec:define_intelligence_test}

We propose \textit{Survival Game}, a framework to evaluate intelligence via a trial-and-error process.
Its core concept is to test how well a subject can autonomously explore and find solutions.
In the following subsections, we first revisit Natural Selection and formalize it as a \textit{Survival Game}. Then, we extend Survival Game as \textit{Survival Game} to quantify intelligence at any task. Next, we interpret the results of \textit{Survival Game} into three levels of intelligence. Furthermore, we propose an approximation method to apply \textit{Survival Game} in costly tasks.
Finally, we discuss how \textit{Survival Game} differs from previous studies.

\subsection{Natural Selection as a Trial-and-Error Test}

Natural Selection is an intuitive way to test intelligence. If a subject passes Natural Selection, it signifies that this subject possesses the intelligence to operate autonomously and can sustain itself without external guidance.
The process of Natural Selection is extraordinarily complex, involving competition between species, genetic mutations, etc. Rather than delving into these intricate details, we simplify Natural Selection into a trial-and-error test as follows:

\begin{definition}
Imagine a species with a sufficiently large population. Its individuals stand in line outside a room. A sign at the entrance warns them that once inside, they will face a critical question. One by one, the individual enters the room and gives answers. An incorrect answer makes the individual vanish, while a correct answer lets it survive. One survivor can mark the species as having passed the test.
\end{definition}

Despite the simplification, this trial-and-error test captures the essence of Natural Selection. Throughout history, nature has posed countless challenges to humankind. When asked how to survive predators, the intelligent among us answered fire and tools. When faced with the threat of starvation, the intelligent among us developed agriculture. When confronted with disease, the intelligent among us advanced medicine. Civilization itself has been forged through these relentless trials and errors. 

Based on the description of the above trial-and-error process, we can translate it into mathematical terms to make it clearer.

\begin{definition}
Let $N$ be the population size of a species. Let \( X \) represent the number of individuals who fail before the correct answer is found. \( X \) takes values in the range of $0 \leq X \leq N$, where $X=0$ means the first individual answers correctly, while $X=N$ means that all individuals fail. If at least one individual succeeds, i.e., \( X < N \), the species passes the game.
\end{definition}

We can see that the number of failures, $X$, is a direct measure of a species' survival intelligence. The smaller the value of $X$, the less effort the species needs to solve problems. Inspired by this, our proposed Survival Game will similarly measure intelligence.

\subsection{Measuring Intelligence with Survival Game}

Based on the trial-and-error process in Natural Selection, we introduce \textit{Survival Game}, which evaluates intelligence by the number of failure attempts in this process. To ensure a robust evaluation result, \textit{Survival Game} models failure counts as a discrete random variable and uses statistical metrics for evaluation. The modifications are two-fold:
\begin{itemize}
	\item Modeling Failure Count as a Discrete Random Variable: One task can involve numerous variations, and failure counts may be very different across these variations. For instance, consider testing a subject's ability to solve mathematical problems. A small change in the numbers or the context of the problem could lead to a significant shift in the subject’s failure counts. Similarly, when a task is classifying images, different pictures can result in substantial fluctuations in performance. Therefore, the variability within the task can cause the results to be unstable. To account for this variability, \textit{Survival Game} models failure count as a discrete random variable, which allows us to handle the variations across task variants effectively.
	\item Statistical Criteria for Evaluation: Population size $N$ serves as a threshold value in the trial-and-error test for Natural Selection. It directly affects the conclusion. The larger the value of $N$, the more attempts are available to the subject, and consequently, the higher the likelihood of success. Yet, for tasks other than survival, the notion of what constitutes an ``appropriate'' $N$ can vary from one researcher to another. This variability in determining an appropriate N leads to inconsistencies in the conclusions. Therefore, \textit{Survival Game} does not use a pre-defined threshold for measurement. It quantifies intelligence as the distribution of failure count. A lower probability of a large failure count suggests higher intelligence. 
\end{itemize}

With these statistical improvements, we formally define \textit{Survival Game} as follows:
\begin{definition}[Survival Game]
Let a subject perform a certain task through continuous trial and error until finding the correct solution. \( X \) is a random variable representing the number of failures before the subject finds the correct solution. Then, $X$ serves as the measure of this subject's intelligence on the task. Smaller expectations and variances of $X$ correspond to higher intelligence.
\end{definition}
Smaller expectations and variances indicate that the subject can achieve success with fewer failures and thus is more intelligent.
This definition allows us to assess intelligence in any given task. We can choose to evaluate intelligence in narrow tasks such as answering domain-specific questions, or we can test a subject across diverse and complex tasks to determine whether it exhibits general intelligence, such as memorizing every information on the Internet.
The measurement of \textit{Survival Game} has a clear physical meaning: it signifies how well a subject can reliably find solutions for a given task on its own.

It is worth noting that \textit{Survival Game} assumes that subjects must keep trying until they succeed even if the cases are very difficult. For easy cases where subjects can answer correctly without trial and error, the contribution to the failure count is zero. We can see that \textit{Survival Game} essentially ignores easy cases where subjects can answer correctly right away and instead focuses on difficult cases that require repeated trials and errors.
For example, in image classification, \textit{Survival Game} focuses on images that the AI model initially misclassifies. It examines how many trial-and-error attempts are needed before achieving the correct classification. This emphasis on trial and error differentiates the Survival Game from existing evaluation methods based on accuracy.
In real-world applications, if a task is highly sensitive to errors, such as high-risk decision-making scenarios like autonomous driving, \textit{Survival Game} provides a better reflection of whether an AI model can be trusted. Additionally, in highly intellectual tasks that require AI to go through trial and error to find a solution, such as proving mathematical theorems or optimizing agent workflows, \textit{Survival Game} metric directly corresponds to computational cost and shall better reflect AI's applicability.

\subsection{Classifying Intelligence into Three Levels}

In this subsection, we analyze the distribution of failure counts obtained from the \textit{Survival Game} to gain a clear understanding of the subject's level of intelligence. First, we introduce an Infinity Assumption to define the least intelligent scenario. Based on this, we then propose three levels of intelligence. Finally, we explain how to classify subjects into these three intelligence levels based on the distribution of their failure counts.

\subsubsection{Infinity Assumption}

What situation represents a subject having almost no intelligence related to the task? Imagine a scenario in an \textit{Survival Game} where a monkey sits in front of a computer and types to see if it can produce Shakespeare’s works. If it deviates from Shakespeare’s works, we let it attempt again. The failure count refers to the number of attempts before success. The monkey has no understanding of human language and just types randomly. In theory, since the human vocabulary is finite and Shakespeare’s works are also of limited length, the monkey could use an enumeration method, blindly trying all possible combinations of words. Even though most of these combinations are completely nonsensical to us, the monkey can eventually type out Shakespeare's works. However, this blind, exhaustive enumeration shows that the subject lacks any real intelligence. It is also disconnected from practical reality because the cost of such an exhaustive search would far exceed any reasonable resource limitations, much like how it is completely unrealistic to expect a monkey to eventually produce Shakespeare’s works. Shakespeare did not create his works by randomly typing and waiting for greatness to emerge. Instead, he produced the masterpieces through intentional creativity within the limitations of human life. Therefore, when the failure count approaches the cost of exhaustive enumeration, it almost certainly indicates that the subject has no intelligence related to the task.

We note that the high cost of blind enumeration closely resembles the mathematical concept of infinity. In mathematics, infinity describes a scenario where a quantity is beyond the scale we can measure or endure. For example, when measuring objects on Earth, we can assume the distance from the Sun to the Earth is infinite, and based on this assumption, we treat sunlight as parallel rays. This is because, compared to the size of objects on Earth, the distance between the Sun and Earth is so vast that it can be approximated as infinity. This allows us to use the property of parallel sunlight to help with measurement tasks. The concept of infinity in mathematics is a way of thinking in terms of limits and approximations. While infinity does not directly exist in the physical world, it helps us understand and describe extremely large quantities and allows us to handle them more conveniently.
In the case of the \textit{Survival Game}, the characteristic of blind enumeration aligns closely with the concept of infinity. In theory, blind enumeration can eventually lead to the correct solution, but the cost of doing so far exceeds the available resources or our willingness. Therefore, we can model the cost of blind enumeration in the \textit{Survival Game} as infinity and thus can better interpret the results of the test.

We propose the following Infinity Assumption: Failure count approaches infinity if it approaches the cost of blindly enumerating all possibilities. In other words, failure count is finite if it is much smaller than the cost of exhaustive enumeration.
Under this mathematical assumption, infinity serves as a clear criterion for determining whether intelligence is present. When the failure count is finite, it indicates that the subject has excluded many possibilities in advance and is consciously engaging in trial and error, ultimately achieving success. At this point, the subject truly demonstrates intelligence in this task. This mathematical assumption allows us to clearly distinguish between different levels of intelligence.

\subsubsection{Three Intelligence Levels}

The above Infinity Assumption links intelligence with infinity. It enables us to clearly define different levels of intelligence in mathematical terms. Based on this, we compare the statistical measures of failure count with infinity and define three levels of intelligence:

\begin{itemize}
	\item \textbf{Limited Level}: A subject belongs to this category if the expectation of failures is infinite: \( E(X) \rightarrow \infty \). At this intelligence level, the subject is comparable to blindly enumerating all possible outcomes. The cost for the subject to autonomously solve the task is unacceptable in real-world scenarios. It requires external supervision to improve itself and reliably operate within the task.
	\item \textbf{Capable Level}: A subject belongs to this category if the expectation of failures is finite, but the variance remains infinite: \( E(X) < \infty , \text{Var}(X) \rightarrow \infty \). At this intelligence level, the subject is, in principle, capable of solving the given task. However, the number of failures vary drastically across different cases. Its performance is highly unpredictable and failures can still occur frequently. As a result, autonomous operation is risky, and external supervision remains necessary to ensure reliability.
	\item \textbf{Autonomous Level}: A subject belongs to this category when both the expectation and variance of failures are finite: \( E(X) < \infty , \text{Var}(X) < \infty \). Subjects at this level can reliably find solutions for the given task. They may operate autonomously without relying on external supervision.
\end{itemize}

If a subject reaches the Autonomous Level, it can reliably find solutions with affordable trials and errors. If we imagine that the subject will use the correct solutions as supervision signals to improve itself, the Autonomous Level implies that the subject no longer requires external supervision to provide correct answers. Instead, it can rely solely on their attempts to find the solution. In this way, the subject can independently generate supervision data and improve itself to further reduce the failure counts. In AI, this process is similar to reinforcement learning, where the system autonomously explores solutions and uses the results to update itself. If the subject has not reached the Autonomous Level, it is almost infeasible to find solutions on its own. More precisely, subjects at the Limited Level require an infinite number of attempts, which is completely beyond reasonable limits, while subjects at the Capable Level are very unstable in finding the solution. These factors make it challenging for the system to autonomously explore solutions and instead necessitate external supervision.

\subsubsection{Decay Rate Classification}
\label{sec:decay_rate_classification}

Before presenting how to practically determine intelligence levels, let us revisit the Infinity Assumption.
Although infinity does not exist in the physical world, this does not prevent us from treating certain quantities, which far exceed our capacity to measure or endure, as if they were infinite. 
In the case of the \textit{Survival Game}, the total number of possible solutions is finite, but as described in the Infinity Assumption, we lack the resources or willingness to blindly enumerate all of them. Therefore, the Infinity Assumption treats the number of possible solutions as if it were infinite. 

Based on the Infinity Assumption, the distribution of the failure count can be seen as extending from 0 to infinity. Therefore, we can assess the convergence of the expectation and variance according to the distribution of the failure count.
Note that the convergence of expectation and variance is determined by the tail behavior of the probability density function. Let $X$ be a discrete random variable, and \( P(X) \) be the discrete probability density function. The convergence of \( E(X) \) and \( \text{Var}(X) \) completely rely on how fast \( P(X) \) decays at the tail. Only if \( P(X) \) is sufficiently small for big \( X \) values will the expectation and variance be finite.

Since the decay rate of failure count determines the convergence of its expectation and variance, it also directly determines the subject's intelligence level. In this way, we connect the intelligence level to the decay rate of failure count.
To examine the decay rate, we introduce power law as reference distributions for comparison. Power law \( 1/x^\alpha \) has the following properties:

\begin{itemize}
    \item When \( \alpha \leq 2 \), both expectation and variance are infinite.
    \item When \( 2 < \alpha \leq 3 \), expectation is finite but variance is infinite.
    \item When \( \alpha > 3 \), both expectation and variance are finite.
\end{itemize}

Therefore, we compare the decay rate of failure count \( P(X) \) with \( x^{-2} \) and \( x^{-3} \), and propose the following classification method to determine the intelligence level:

\begin{itemize}
	\item If \( P(X) \) decays more slowly than \( x^{-2} \), both expectation and variance are infinite. The subject is at the Limited Level.
	\item If \( P(X) \) decays faster than \( x^{-2} \) but more slowly than \( x^{-3} \), expectation is finite but variance is infinite. The subject is at the Capable Level.
    \item If \( P(X) \) decays faster than \( x^{-3} \), both expectation and variance are finite. The subject is at the Autonomous Level.
\end{itemize}

\begin{figure}[t]
  \centering
  \includegraphics[width=0.4\textwidth]{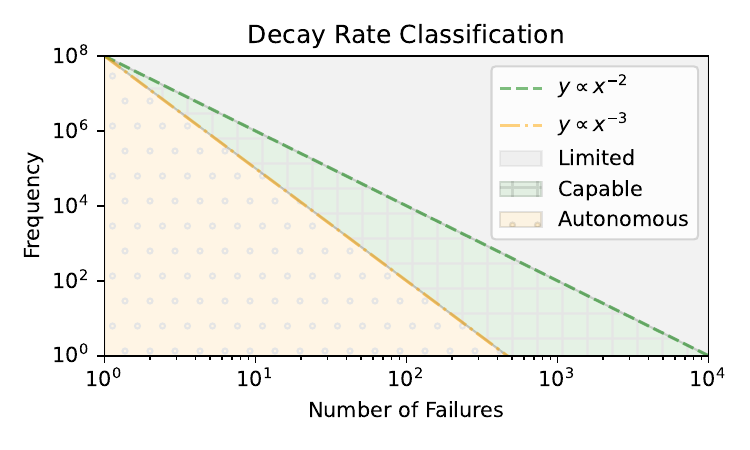}
  \caption{Decay Rate Classification: Log-log plot of failure counts (x-axis) vs. probability (y-axis). The intelligence level is determined based on which region the distribution of the subject falls in.}
  \label{fig:decay_rate_classify}
\end{figure}

A practical way to visualize this comparison is to plot \( P(X) \) alongside these two reference power-law functions on a log-log scale. On such a plot, the reference functions appear as straight lines, allowing for an intuitive comparison of decay rates. As shown in Figure~\ref{fig:decay_rate_classify}, the two reference distributions divide the graph into three distinct regions, corresponding to Limited Level, Capable Level, and Autonomous Level, from top to bottom. We can easily determine the intelligence level of the subject by examining which region \( P(X) \) falls in.

\subsection{Approximation with Reference Answers}

Note that \textit{Survival Game} requires to determine whether each attempt made by a subject is correct. Yet verifying correctness for every attempt can be expensive in some tasks. Consider a task where the test subject is to prove a mathematical theorem. The subject provides proof with each attempt. However, for complex mathematical theorems, the proofs can be very long and intricate, and the cost of verifying the correctness of each proof is extremely high. In such cases, directly applying \textit{Survival Game} may make the process prohibitively expensive.

Therefore, we propose an approximation method to address this problem. We will first introduce the underlying assumption and then formalize the approximation method.

\subsubsection{Scale-Invariance Assumption}

In situations where it is difficult to verify the correctness of each trial, we can adopt an alternative approach: counting the failure attempts before the subject arrives at a \textbf{predefined reference answer}.
For example, when testing whether a subject can prove mathematical theorems, we do not evaluate whether each of its outputs constitutes a valid new proof. Instead, we check whether it can produce a known proof.

To support the validity of such reference-based evaluation, we propose a Scale-Invariance Assumption:
\begin{itemize}
	\item The number of failures before finding any solution follows a power-law distribution.
	\item The number of failures before finding any solution is linearly related to the number of failures before finding a particular solution. 
\end{itemize}

Under the given assumption, we can prove that the number of failures before finding any solution and the number of failures before finding a particular solution have the same failure decay rate. Therefore, they yield the same result in our Decay Rate Classification. The proof is as follows:
\begin{equation}
    P(X = x) = C x^{-\alpha}, \quad x \geq x_{\text{min}}.
\end{equation}
\begin{equation}
    P(kX = x) = P(X = x/k) = C (x/k)^{-\alpha} = C k^{\alpha} x^{-\alpha} \propto x^{-\alpha}.
\end{equation}
Thus, the power-law distribution retains its functional form under linear scaling.
If the failure count follows the power law, a linear transformation does not change the power-law formulation and the exponent.

The first part of the Scale-Invariance Assumption is empirically and theoretically supported.
Specifically, Section~\ref{sec:future_prediction} will show that the failure count is close to a power-law distribution. In Section~\ref{sec:theory}, we theoretically analyze the cause of such a phenomenon. We demonstrate that this is because human tasks exhibit criticality property.

The second part of the Scale-Invariance Assumption requires further investigation. 
Whether it is linearly correlated depends on the relationship between references, tasks, and subjects.
In our experiments, we use human-written answers as references. For example, when assessing whether the model can write mathematical proofs, we use human-written proofs as references. When evaluating whether the model can generate high-quality legal opinions, we use legal opinions written by human judges. Similarly, when assessing whether the model can produce excellent literary works, we use human literary works as references.
In these cases, we assume that the number of failed attempts to arrive at a feasible solution is linearly related to the number of attempts to reach these reference solutions. 
We have not verified its correctness for now and will verify it in the future.

\subsubsection{Survival Game with References}

Based on the Scale-Invariance Assumption, we propose a variation named \textit{Survival Game with References}. It avoids the need for direct correctness verification while keeping the core of \textit{Survival Game}. In those tasks where the cost of correctness verification is high, it uses a reference answer and measures the number of failed attempts before producing the reference answer. The validity of this method is supported by the following theorem:
\begin{thm}[Survival Game with References]
Let \( X^* \) be a discrete random variable representing the number of failure attempts before finding a \textbf{predefined reference answer}.  
\( X^* \) is an upper bound estimation of the real failure counts.
If the Scale-Invariance Assumption holds, the failure decay rate of \( X^* \) is accurate.
\end{thm}

This approach eliminates the need for verifying every attempt and instead examines failure counts until reaching a known reference answer. It is a low-cost realization of \textit{Survival Game} and reflects an upper bound of the subjects' errors.
If the Scale-Invariance Assumption holds, this theorem shows that we can exactly evaluate the intelligence level in an efficient way.

\subsection{Comparison with Related Work}
\label{sec:comparison_with_related_work}

Since we have introduced \textit{Survival Game}, we can pick up our discussion from Section~\ref{sec:related_work}. 
In contrast to the subjective tests in prior studies, \textit{Survival Game} provides an \textit{objective} way to evaluate intelligence:
\begin{itemize}
	\item \textit{Objective}~(Species-Agnostic) View of Intelligence: We define intelligence not by its similarity to humans, but by the ability to pass a test akin to Natural Selection. Any entity that can independently find solutions demonstrates intelligence, regardless of whether it is human, artificial, or another species. Even humans may not necessarily be at the Autonomous Level in some tasks, and the test is always applicable no matter whether AI surpasses humans.
	\item \textit{Objective} Choice of Tasks: We recognize that intelligence is inherently task-dependent. Unlike previous approaches that attempt to define universal intelligence, \textit{Survival Game} does not prescribe any specific task. Instead, it allows researchers to evaluate intelligence in any task of interest, ensuring that the definition of intelligence remains grounded in the actual demands of a given task.
	\item \textit{Objective} Evaluation Framework: The \textit{Survival Game} is mathematically well-defined and does not rely on any hyperparameters. Its conclusions are based on clear statistical criteria rather than subjective assessments. This ensures that evaluations remain consistent across different studies and applications, making it a robust and practical tool for assessing intelligence in real-world settings.
\end{itemize}

It is important to note that while we argue that intelligence is inherently task-dependent and should be evaluated within specific tasks, this does not prevent researchers from using \textit{Survival Game} as a framework for assessing Artificial General Intelligence~(AGI). 
From our point of view, an AGI system should reach the Autonomous Level in at least every basic human task. Therefore, to evaluate general intelligence, researchers can construct a diverse set of tasks and apply \textit{Survival Game} on each of them.

\section{Evaluation with Survival Game}
\label{sec:evaluate_current_ai}

In this section, we evaluate state-of-the-art AI systems with \textit{Survival Game}. We adopt a wide range of tasks, including vision, search, recommendation, and language. 

\textbf{Quantify AI's failures}:
We calculate the number of failures based on the scores output by the AI system. More precisely, for a given task, existing AI systems output a score for each potential answer. For instance, an image classification model assigns a score to each class; a search engine model predicts a relevance score for each document; a recommendation system assigns a score to each product; and a language model outputs a score to each word. A higher score represents a higher possibility the AI system predicts that this is the correct answer. We rank the answers based on the model's output score from highest to lowest. This ranking list is the model's attempt sequence, and the failure count equals the position of the reference answer minus one. 

The following presents the evaluation results of these models across various tasks. We will see that current models only reach the Autonomous Level in simple tasks and are at the Limited Level in most complex tasks. At the end of this section, we revisit existing AI techniques and show that these techniques are exactly developed in the context of Limited-Level intelligence.

\subsection{A Beginner's Task: MNIST}

\begin{figure*}[t]
    \subcapraggedrighttrue
    \subcaphangtrue
        \centering
        \subfigure{\includegraphics[width=0.3\textwidth]{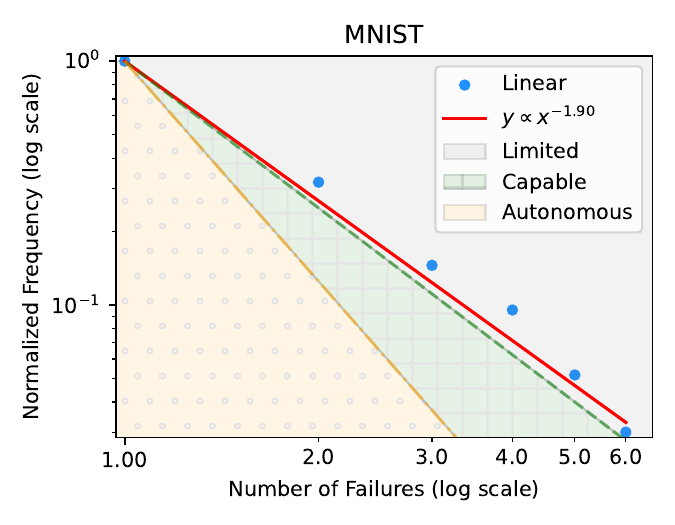}} \hfill
        \subfigure{\includegraphics[width=0.3\textwidth]{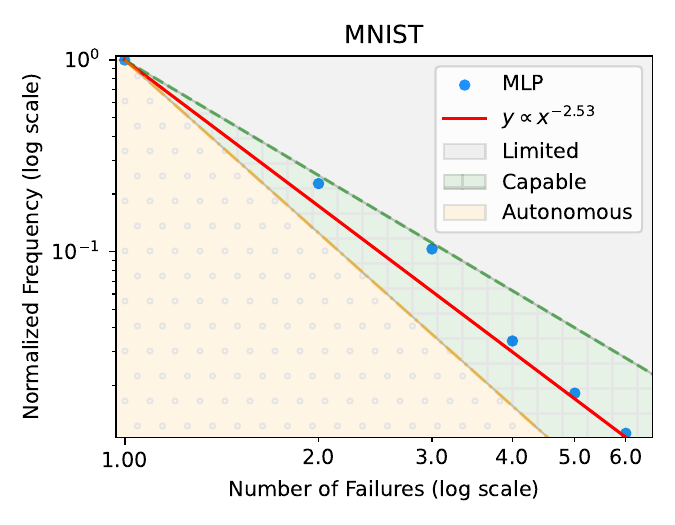}} \hfill
        \subfigure{\includegraphics[width=0.3\textwidth]{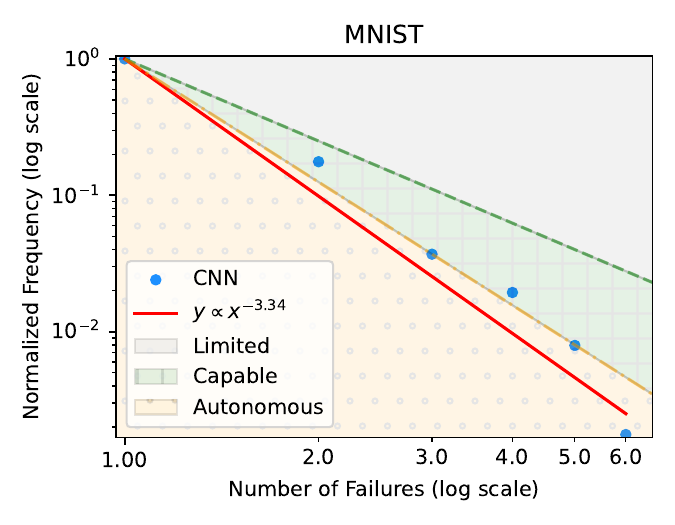}}
        \caption{Experimental Results of \textit{Survival Game} in handwritten digit recognition task (MNIST). The red line is a power law curve drawn based on the distribution of the model's data points. Results suggest that a system reaches a higher-level of intelligence if it better models the task.}
        \label{fig:mnist_result}
    \end{figure*}

MNIST~\citep{deng2012mnist} is a handwritten digit recognition task. It consists of a collection of images depicting the digits 0-9, written by different people. The task is for an AI system to correctly identify the digit in each image. Many people consider MNIST to be a relatively simple task, and it is often used as an introductory challenge for beginners to experiment with various AI algorithms. As such, we also start with this task to test whether \textit{Survival Game} can effectively distinguish different types of AI algorithms.

We used three AI algorithms: a linear classifier, a multilayer perceptron~(MLP) classifier~\citep{haykin1994neural}, and a convolutional neural network~(CNN). Neither the linear classifier nor the MLP classifier takes into account the specific characteristics of the task; they both flatten the 2D image into a 1D vector and perform transformations on this vector to do the classification. The transformation for the linear classifier is linear, while the MLP classifier introduces non-linear activation functions. In contrast, CNN is equipped with a deeper understanding of images. It uses convolution to capture local features and employs multiple layers to extract abstract semantic information. Therefore, from the perspective of task modeling, CNN performs more in-depth modeling compared to both the MLP and the linear classifier. We train the three models on MNIST's training data. To ensure the stability of the results, we run the experiments with $10$ different random seeds and then average the failure count distribution.  

The experimental results are shown in Figure~\ref{fig:mnist_result}. The blue dots are failure count distribution, and the red line is a power law for reference. The gray, green, and yellow regions represent Limited Level, Capable Level, and Autonomous Level, respectively. As we can see, \textit{Survival Game} clearly distinguishes these three different methods: The linear classifier falls within the Linear Level region, the MLP classifier falls within the Capable Level region, and CNN approaches the Autonomous Level region. Therefore, the better the system models the task, the higher its intelligence level. This suggests that the Survival Game is effective at evaluating the intelligence level of an AI system.

\subsection{Vision}

In this subsection, we test whether current AI models can effectively execute complex vision tasks. We select two types of tasks for evaluation. The first is an image classification task. Given an image, the model needs to identify what animal or object is present and categorize it appropriately. For this task, we use a widely recognized dataset, ImageNet-1K~\citep{deng2009imagenet}. The second task is more complex: given a natural language description, the model should find the corresponding image from a large set of images. Compared to image classification, this task requires the model to understand the meaning of a long natural language description and have a deeper understanding of complex images. We use two popular datasets for this task: MS COCO ~\citep{lin2014microsoft} and Flickr30k~\citep{plummer2015flickr30k}.

We evaluate state-of-the-art AI models currently available in the field. In the first image classification task, we use CLIP model~\citep{radford2021learning} and MAE models of various sizes~\citep{he2022masked}. CLIP is widely used for visual tasks, such as text-to-image generation. The MAE models are among the best-performing on ImageNet. For the second task, we select several top-performing models from the relevant task leaderboard~\citep{ilharco2021openclip}, namely DFN-VIT-L, ConvN-XXL, and EVA01-G. These models are not only large in parameter size but also in the size of the training data. They represent the best models in the field.

\begin{figure*}[t]
    \subcapraggedrighttrue
    \subcaphangtrue
        \centering
        \subfigure{\includegraphics[width=0.23\textwidth]{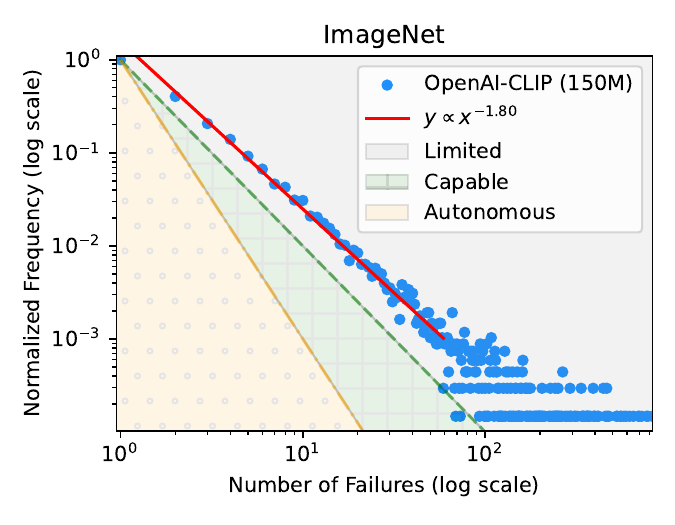}}
        \subfigure{\includegraphics[width=0.23\textwidth]{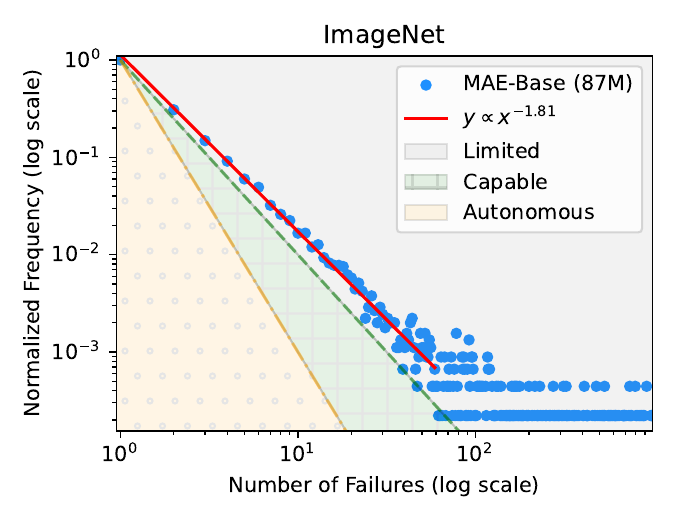}}
        \subfigure{\includegraphics[width=0.23\textwidth]{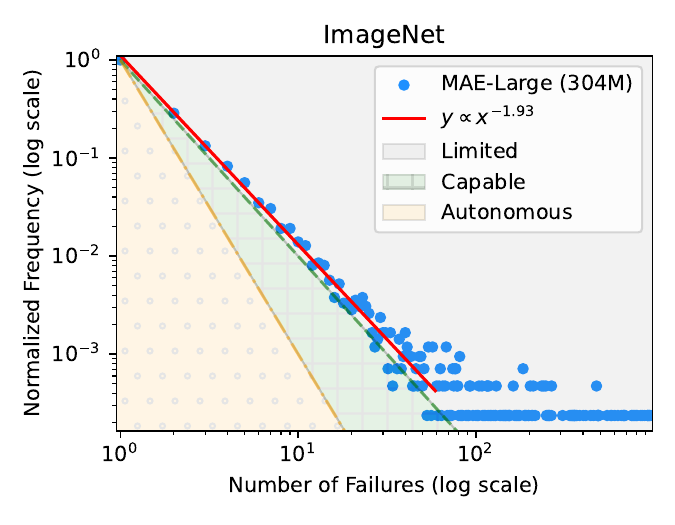}}
        \subfigure{\includegraphics[width=0.23\textwidth]{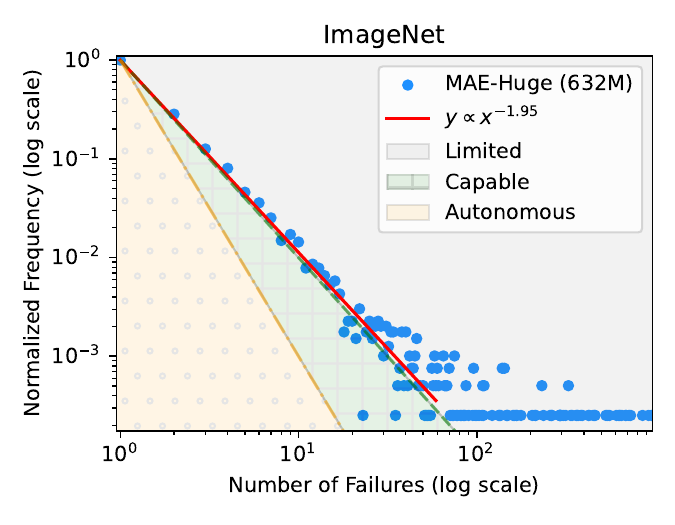}}

        \subfigure{\includegraphics[width=0.23\textwidth]{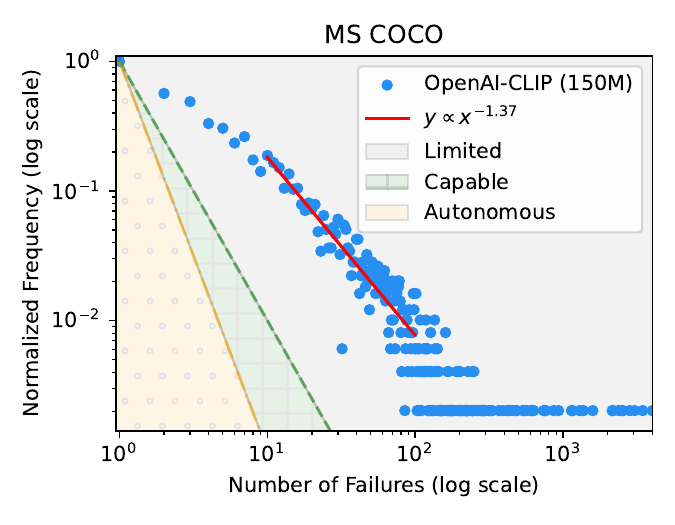}}
        \subfigure{\includegraphics[width=0.23\textwidth]{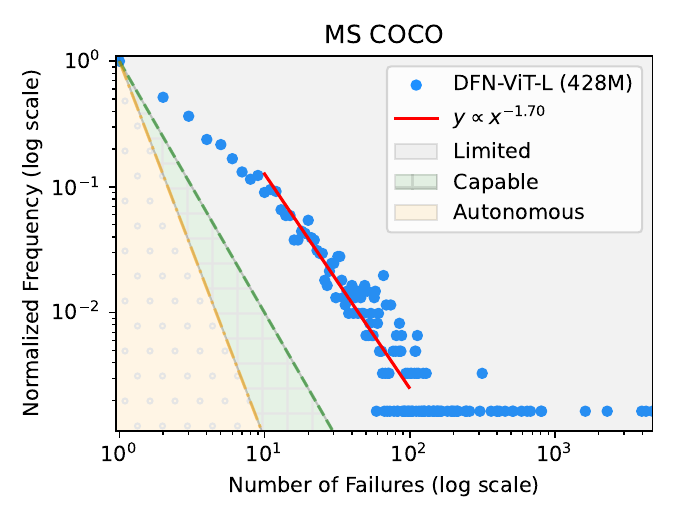}}
        \subfigure{\includegraphics[width=0.23\textwidth]{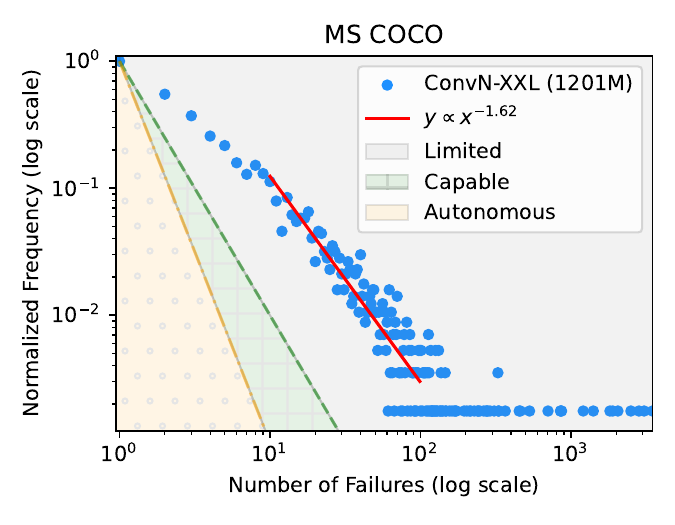}}
        \subfigure{\includegraphics[width=0.23\textwidth]{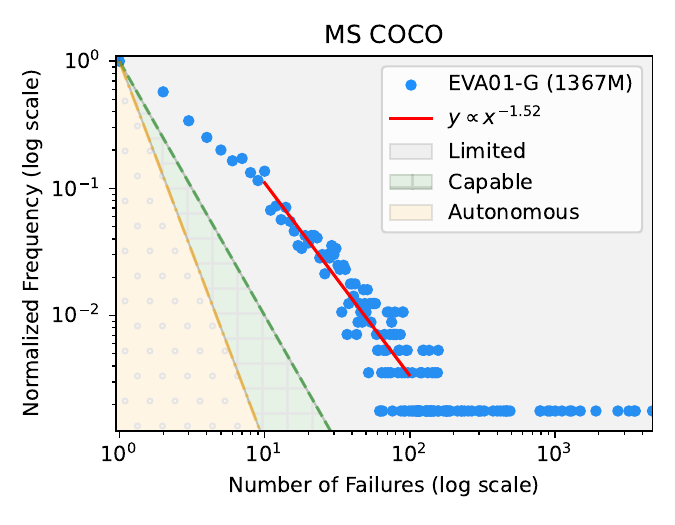}}

        \subfigure{\includegraphics[width=0.23\textwidth]{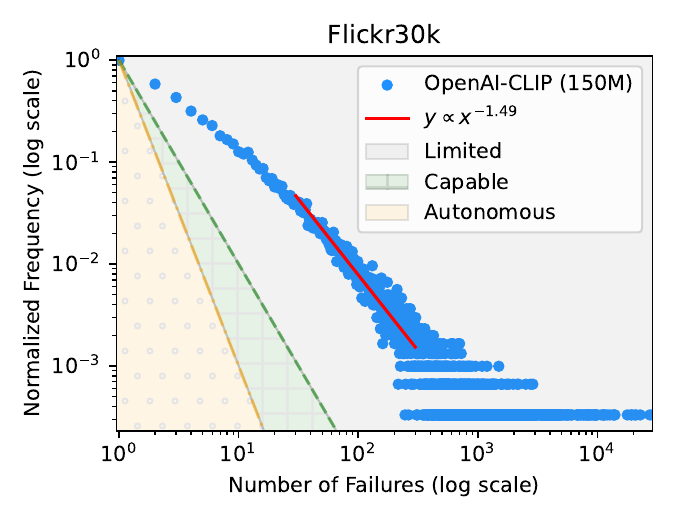}}
        \subfigure{\includegraphics[width=0.23\textwidth]{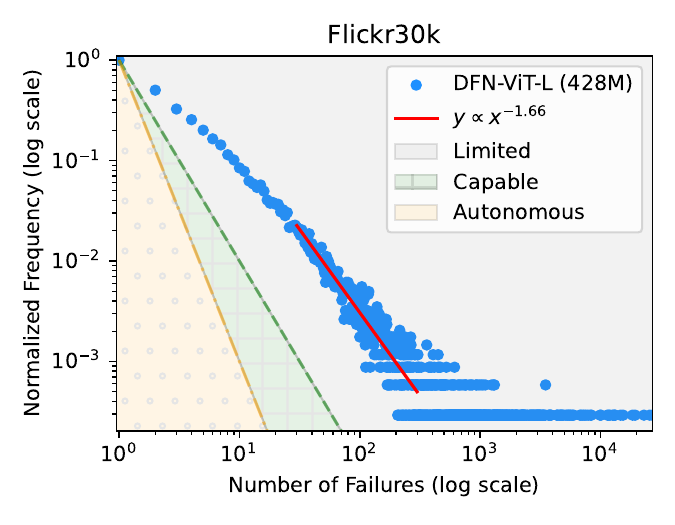}}
        \subfigure{\includegraphics[width=0.23\textwidth]{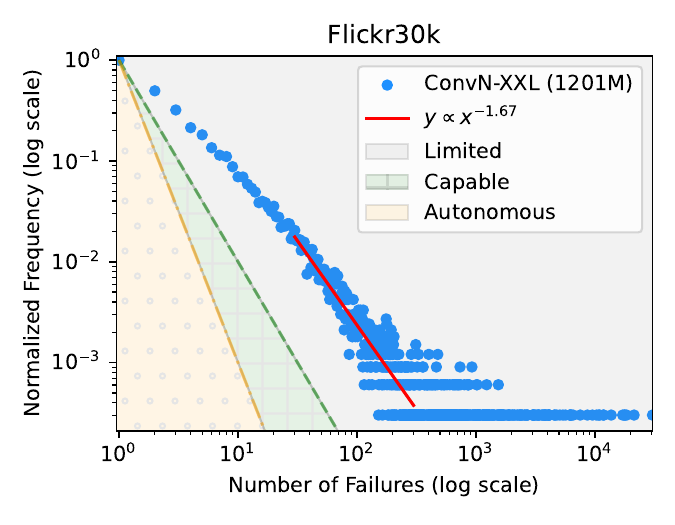}}
        \subfigure{\includegraphics[width=0.23\textwidth]{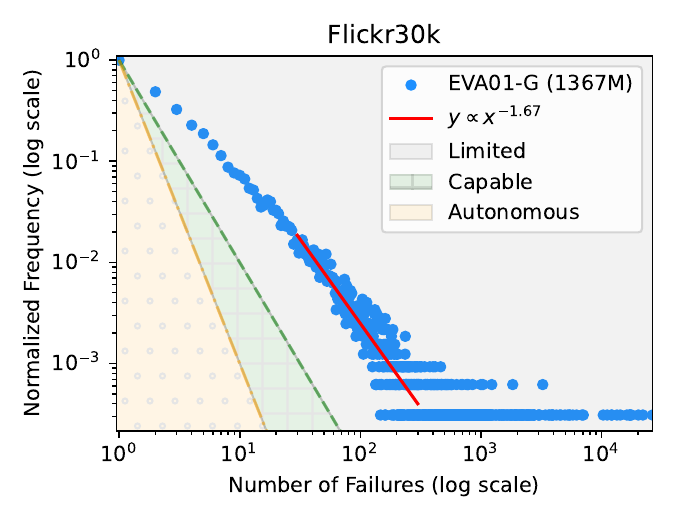}}
        \caption{Experimental Results of \textit{Survival Game} in Computer Vision. The three rows correspond to three different datasets. Figures in different columns correspond to different models. The red line is a power law curve drawn based on the distribution of the model's data points. Its exponent roughly represents the model's failure decay rate. Results show that models are at Limited Level.}
        \label{fig:vision_result}
    \end{figure*}

The experimental results are shown in Figure~\ref{fig:vision_result}. 
The first row shows the results of the image classification task, with different images corresponding to different models. We can see that all models are at Limited Level. As we use a larger MAE model, the decay rate increases and data points gradually approach the Capable Level. In the two subsequent rows, we show the results for the MS COCO and Flickr30k datasets. Different images in the same row correspond to different models. The results show that even the most advanced models today are at Limited Level, with decay rates around $1.7$ or below, far from Capable Level's threshold of $2$. We can also see a similar trend as observed in the first row: the larger the model, the closer it is to the Capable Level. But the marginal improvements diminish gradually. 

The fact that these models are at Limited Level points to a clear physical meaning: if these models are to find out the answers to a vision-related task when they are wrong in the first place, the model would, in statistical terms, need an infinite number of guesses to get it right. In other words, the model not only makes incorrect predictions but also regards the correct answer as completely wrong. If it tries to solve the task, it will try many incorrect answers before finally outputting the correct one. 
Therefore, we should not place blind trust in visual models' results. Instead, we should provide supervision and guidance to ensure their reliability.

\subsection{Search}

Next, we evaluate the performance of text search models. Text search should be familiar to many people. It has widespread applications in search engines like Google, Bing, and Baidu. Given a query, the text search model ranks the candidate documents in order of relevance from highest to lowest. We regard this ranking list as its attempt sequence when applying \textit{Survival Game}.

We use a diverse range of datasets. 
We synthesize a basic dataset so that readers can have a better understanding of the task. We use Wikipedia as the raw data and construct a text search task with its titles and documents. Given a title, the search model ranks all the documents and should put the corresponding document at the top of the ranking list. The number of failure attempts is equal to how many incorrect documents are ranked higher than the correct ones. 
Besides this synthetic dataset, we also use many real-world search datasets.
We adopt two web search datasets, MS MARCO~\citep{Bajaj2016Msmarco} and T2Ranking~\citep{xie2023t2ranking}. The former is in English and the latter is in Chinese. Both were derived from real user queries on search engines. They are widely used to benchmark the effectiveness of text search models.
We also use datasets from finance domain and social platforms: FiQA~\citep{Maia2018FiQA}, CqadupStack~\citep{hoogeveen2015}, and Quora~\citep{Iyer2022first}. FiQA requires the model to find the relevant answers to financial questions. CqadupStack and Quora are released by StackExchange and Quora social platforms, respectively. Given a query, they require models to find duplicate queries.

We use three distinct search models for evaluation. The first is BM25~\citep{robertson1976relevance}, a popular model that was proposed decades ago. It is based on exact match and term frequency weighting. We implement it with Anserini toolkit~\citep{yang2017anserini}.
The second is dense retrieval~\citep{karpukhin2020dense, reimers2019sentence}, which represents both the query and the documents as semantic vectors and ranks them based on vector similarity. We use two open-sourced models from BGE~\citep{bge_embedding} since they are top performers on the related leaderboard. The two models vary in size, and we denote them as DR Small and DR Base.
The third is cross-encoder~\citep{nogueira2019passage}, which takes both the query and the document as input and uses attention mechanisms to model their interaction. In this way, it captures more nuanced matching signals and predicts relevance more accurately. We use two strong open-sourced models. On the English dataset, we use MiniLM cross-encoder~\citep{reimers2019sentence}. On the Chinese dataset, we use BGE cross-encoder~\citep{bge_embedding}.

\begin{figure*}[t]
    \subcapraggedrighttrue
    \subcaphangtrue
        \centering
        \subfigure{\includegraphics[width=0.23\textwidth]{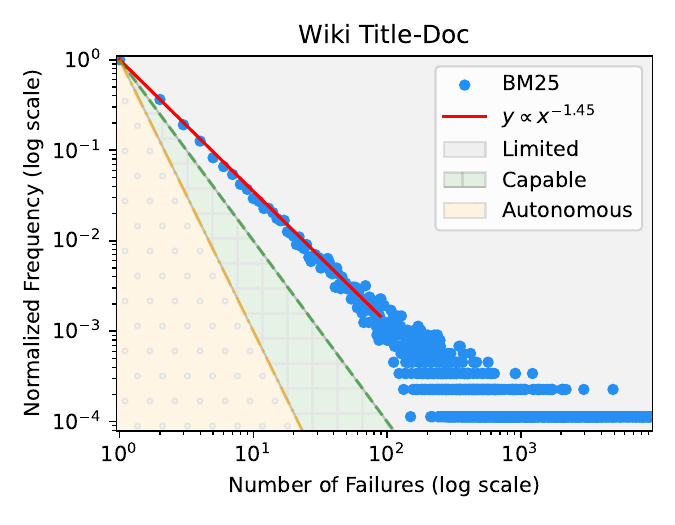}}
        \subfigure{\includegraphics[width=0.23\textwidth]{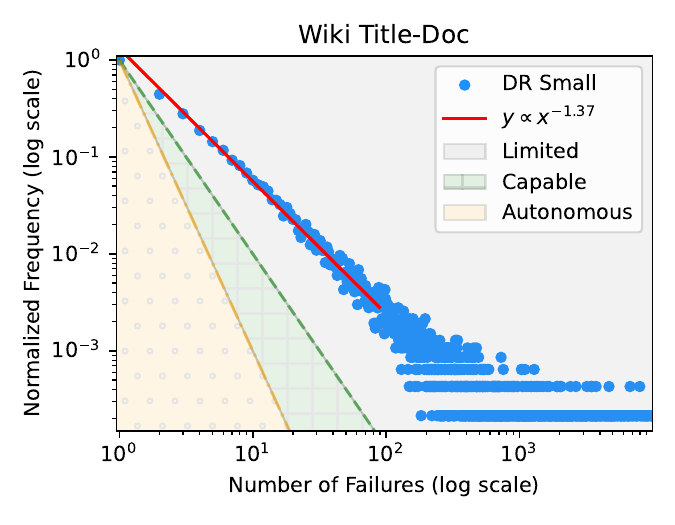}}
        \subfigure{\includegraphics[width=0.23\textwidth]{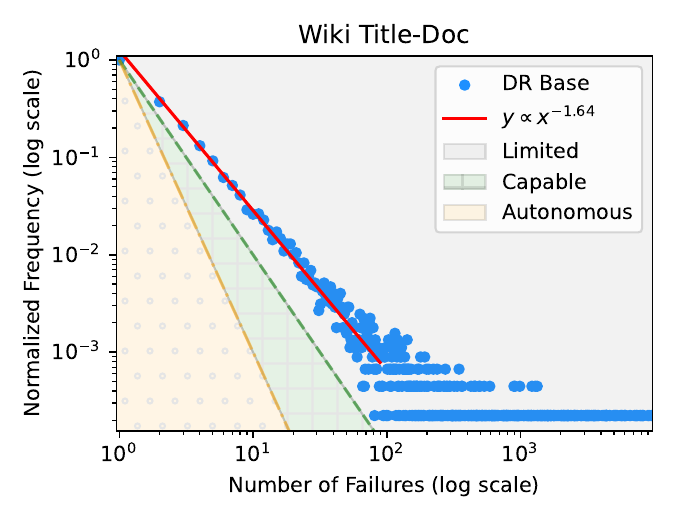}}
        \subfigure{\includegraphics[width=0.23\textwidth]{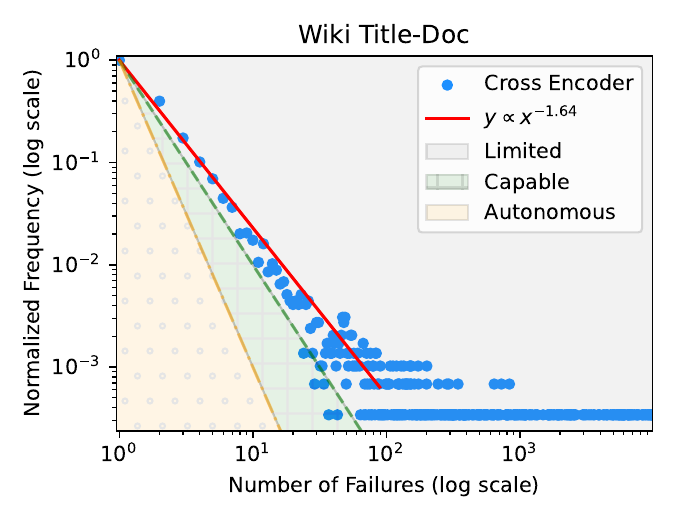}}
		
		\subfigure{\includegraphics[width=0.23\textwidth]{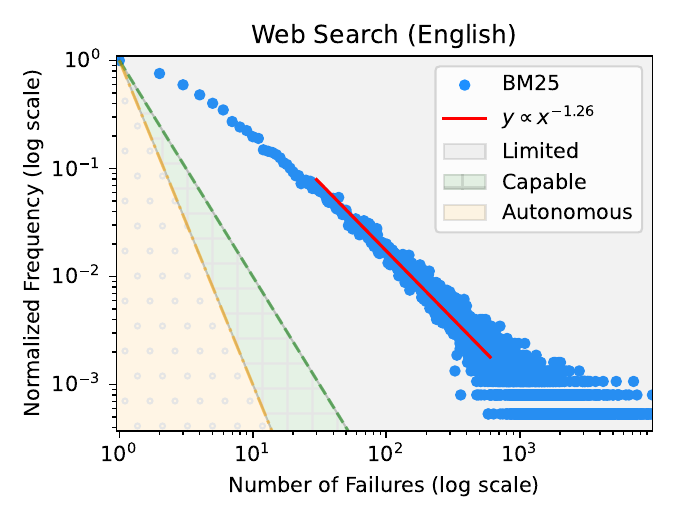}}
        \subfigure{\includegraphics[width=0.23\textwidth]{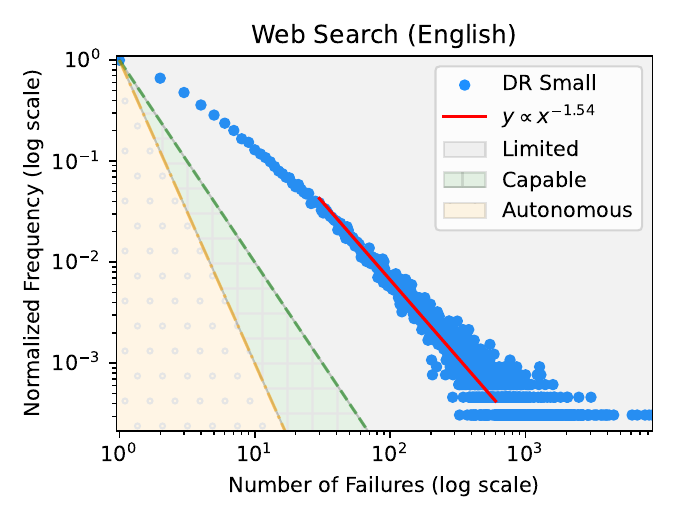}}
        \subfigure{\includegraphics[width=0.23\textwidth]{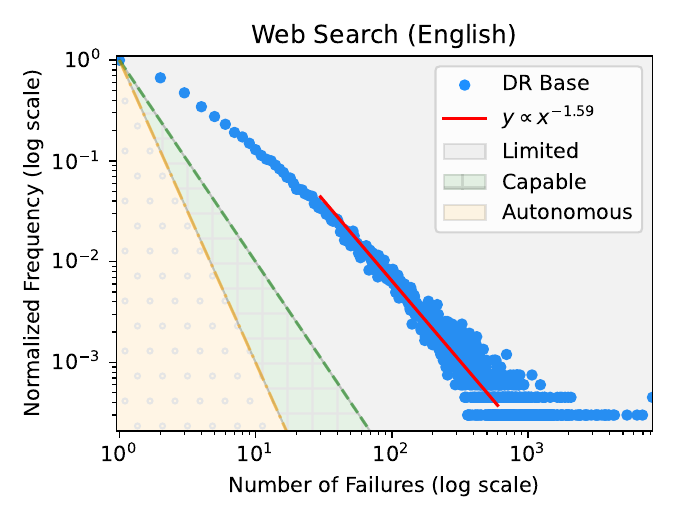}}
        \subfigure{\includegraphics[width=0.23\textwidth]{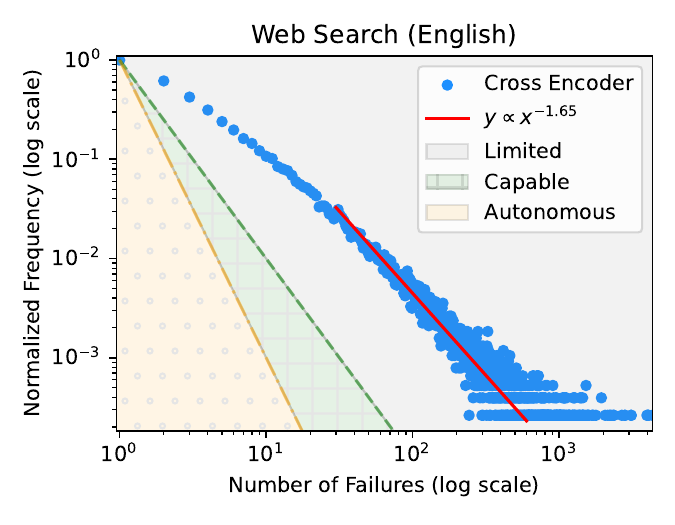}}

        \subfigure{\includegraphics[width=0.23\textwidth]{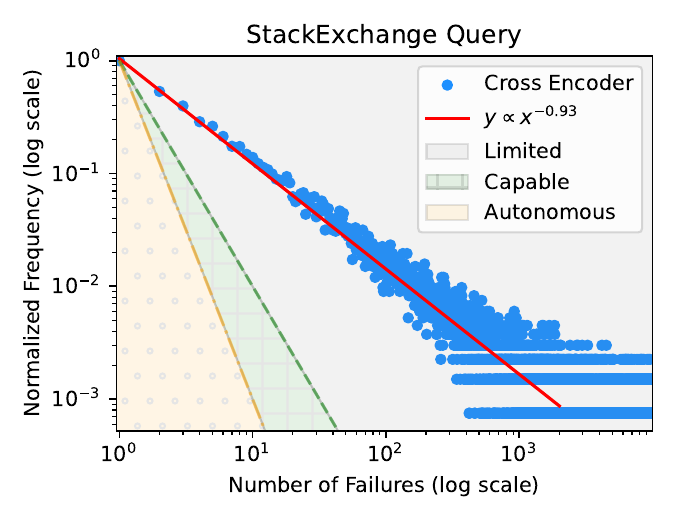}}
        \subfigure{\includegraphics[width=0.23\textwidth]{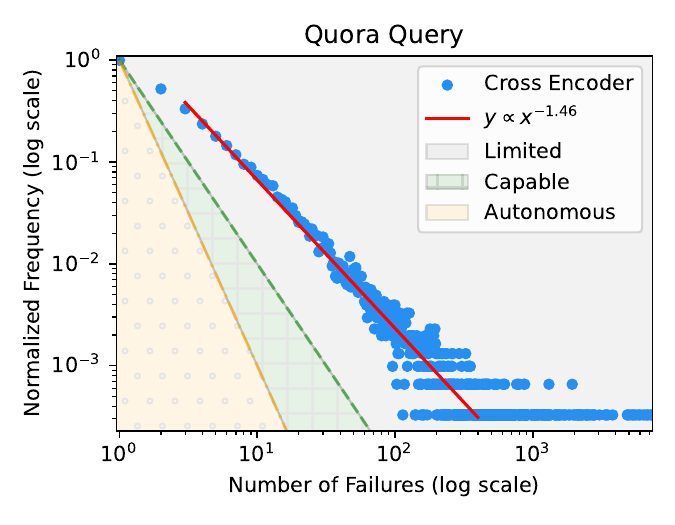}}
        \subfigure{\includegraphics[width=0.23\textwidth]{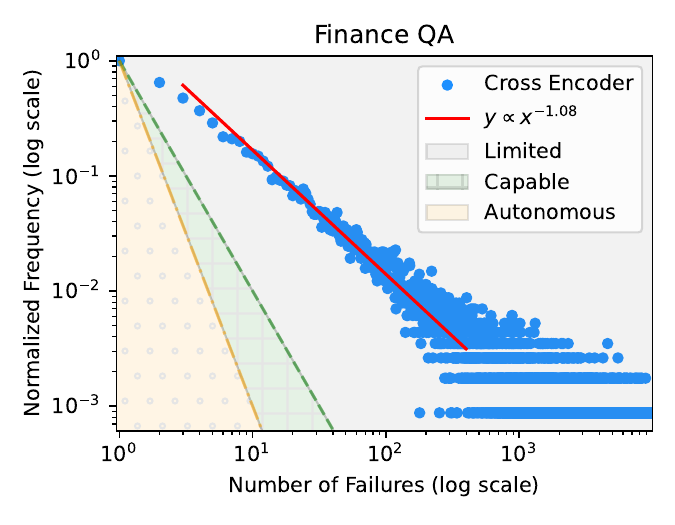}}
        \subfigure{\includegraphics[width=0.23\textwidth]{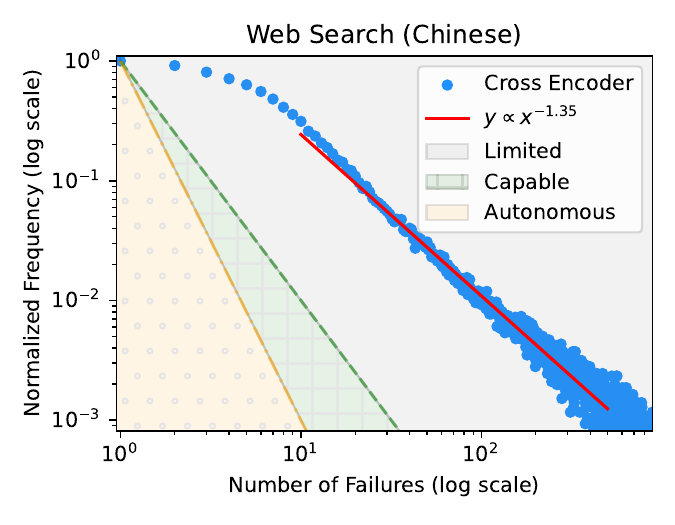}}
        \caption{Experimental Results of \textit{Survival Game} in Text Search. The first two rows show performance on the synthetic Wiki task and the web search task. The final row shows performance of the cross-encoder on another four tasks. The red line is a power law curve drawn based on the distribution of the model's data points. Its exponent roughly represents the model's failure decay rate. Results indicate that all models are at the Limited Level.}
        \label{fig:text_search_results}
    \end{figure*}

The experimental results are shown in Figure~\ref{fig:text_search_results}. The first two rows show the performance of the Wikipedia synthetic dataset and the English web search dataset, respectively. The third row shows the performance on other datasets. We can see that on all datasets and for all text search models, the performance remains at the Limited Level. On the synthetic Wikipedia dataset, the current models' performance is close to the Capable Level. On other real-world datasets, the models are far from the Capable Level. Besides, from the results in the first two rows, as the models become larger and more complex, their decay rate increases and data points move closer to the Capable Level.

Limited Level has a clear physical meaning in the text search scenario: when a user submits a query and the right document is not ranked at the top, the right document is likely to be ranked at the end of the list. The user needs to read, in statistical terms, infinite irrelevant documents before reaching the document they are looking for. In other words, when a search model makes a mistake, it is almost completely unable to correct itself.
This highlights the complexity of the text search task and the inadequacy of current search technologies. It inspires us that we cannot simply rely on search engines to seek information.

\subsection{Recommendation}

\begin{figure*}[t]
    \subcapraggedrighttrue
    \subcaphangtrue
        \centering
        \subfigure{\includegraphics[width=0.23\textwidth]{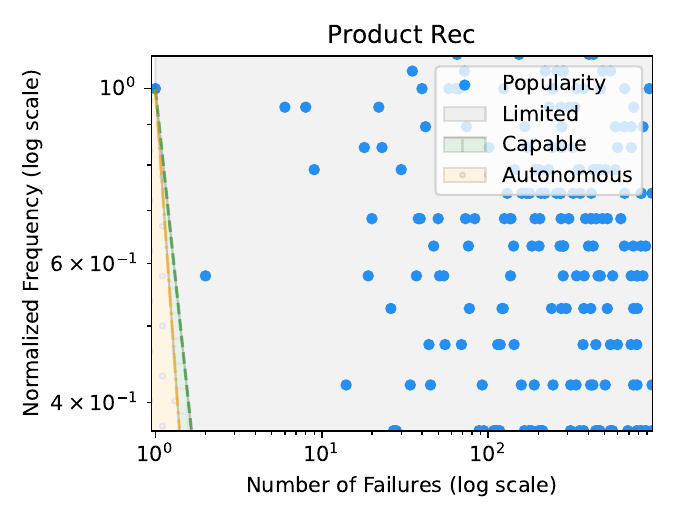}}
        \subfigure{\includegraphics[width=0.23\textwidth]{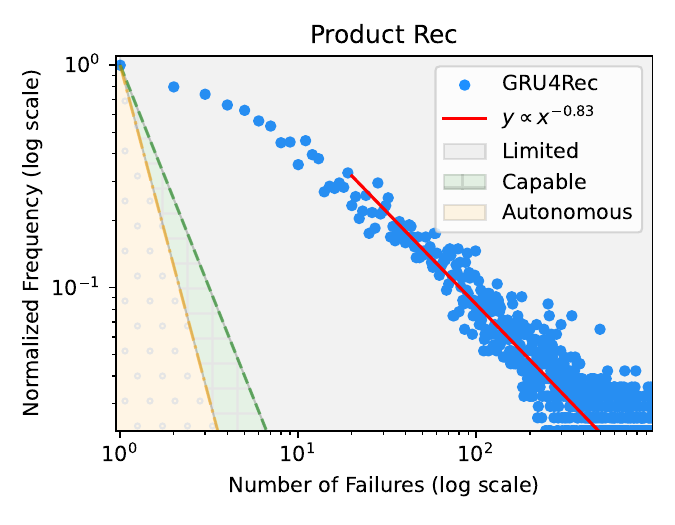}}
        \subfigure{\includegraphics[width=0.23\textwidth]{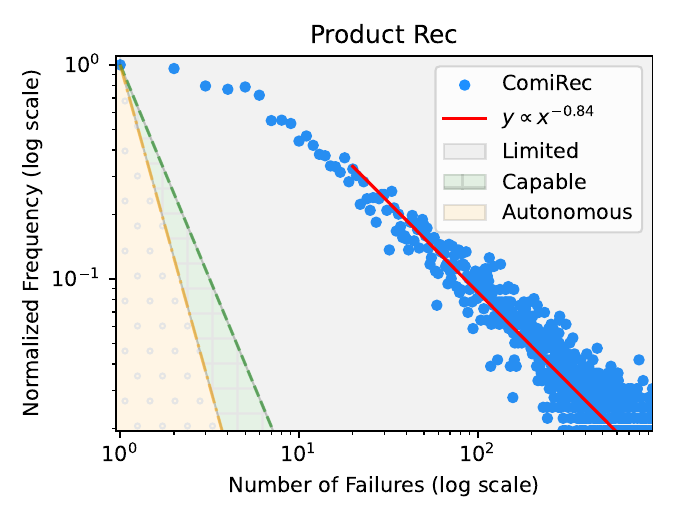}}
        \subfigure{\includegraphics[width=0.23\textwidth]{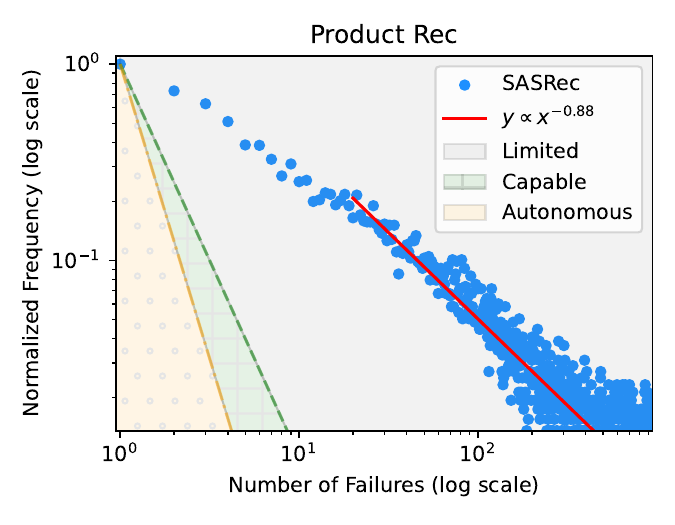}}
        
        \subfigure{\includegraphics[width=0.23\textwidth]{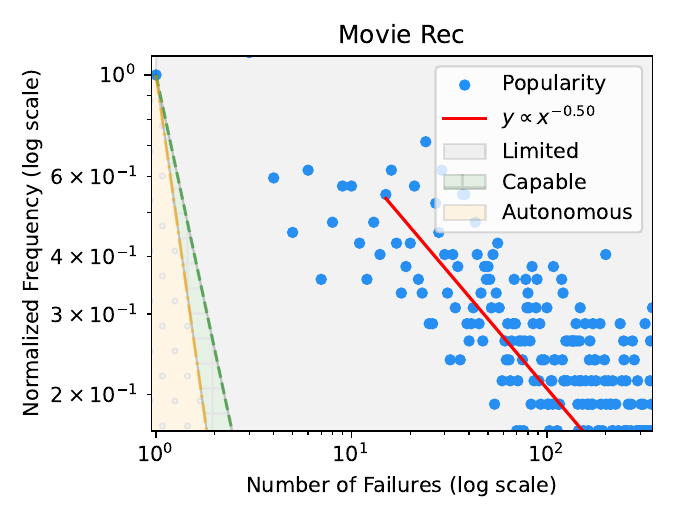}}
		\subfigure{\includegraphics[width=0.23\textwidth]{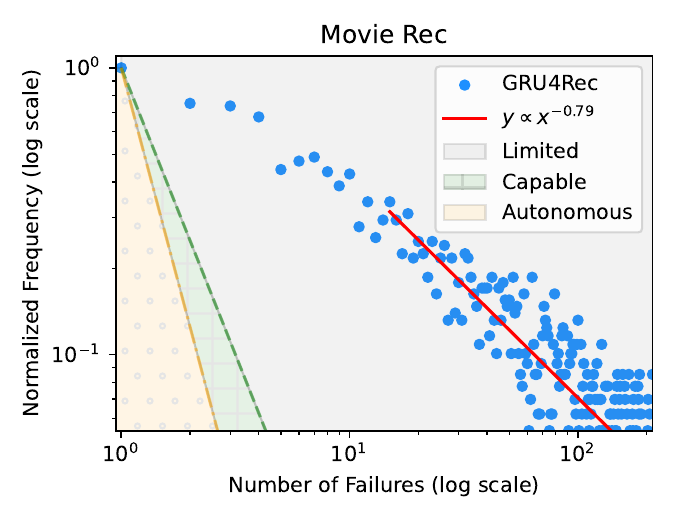}}
        \subfigure{\includegraphics[width=0.23\textwidth]{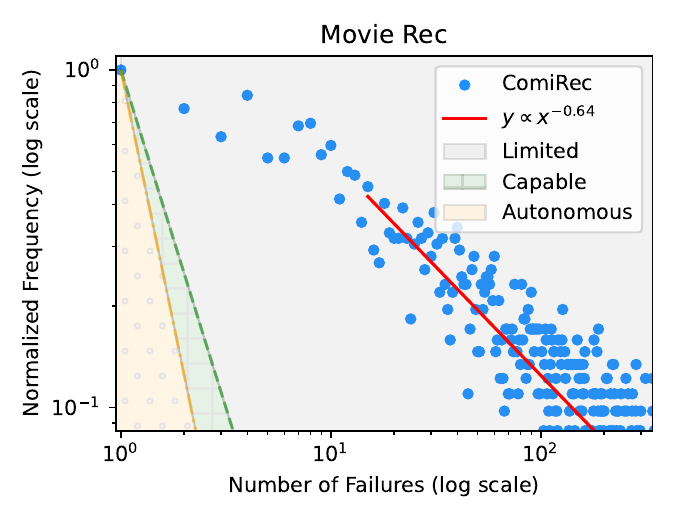}}
        \subfigure{\includegraphics[width=0.23\textwidth]{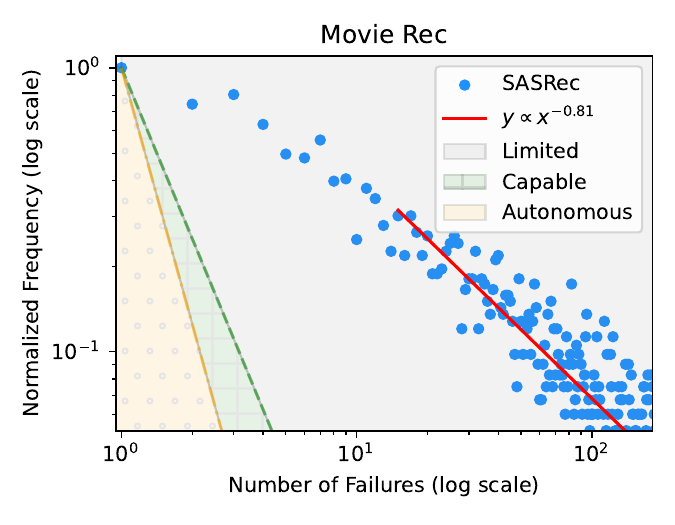}}

        \subfigure{\includegraphics[width=0.23\textwidth]{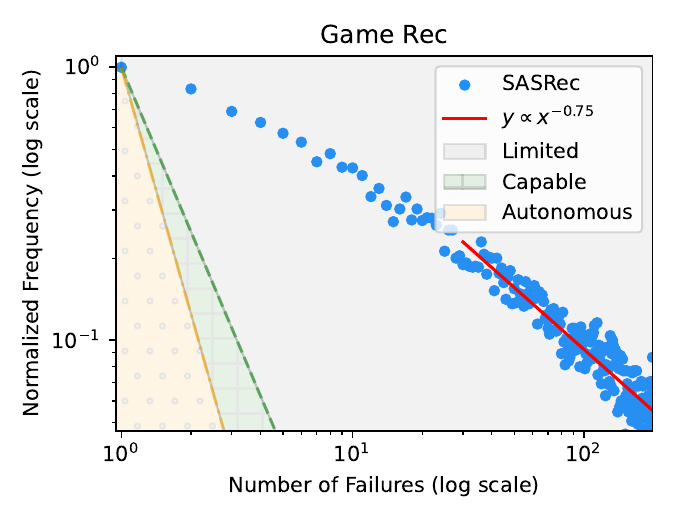}}
        \subfigure{\includegraphics[width=0.23\textwidth]{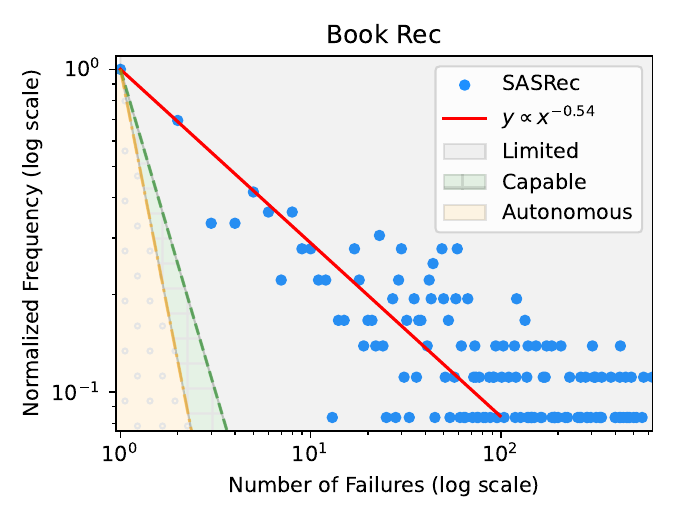}}
        \subfigure{\includegraphics[width=0.23\textwidth]{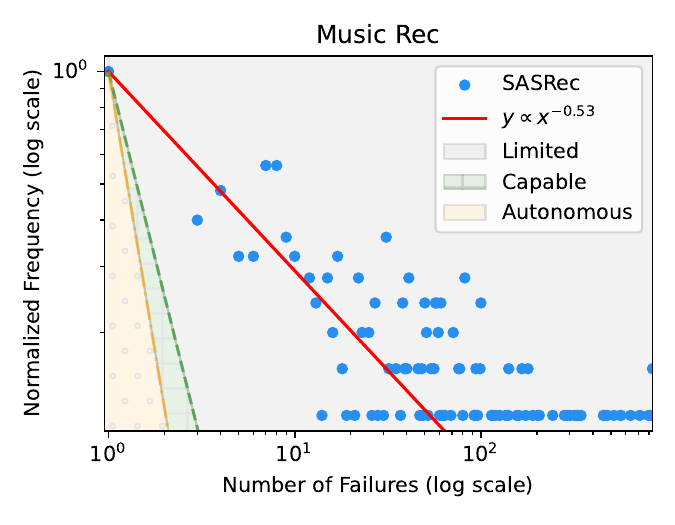}}
        \subfigure{\includegraphics[width=0.23\textwidth]{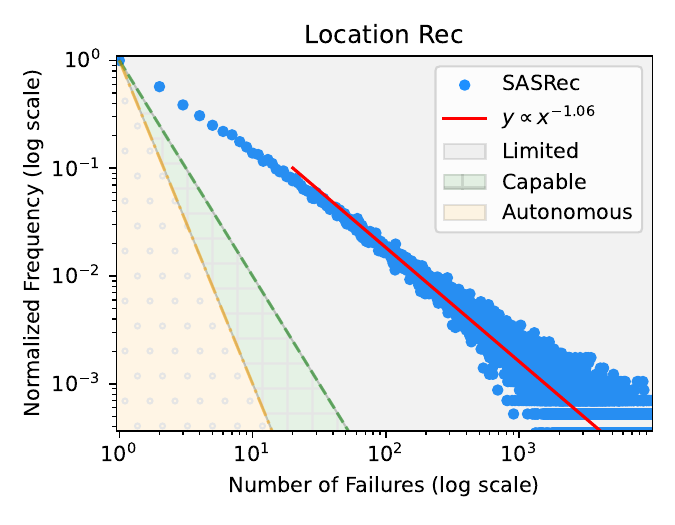}}
        \caption{Experimental Results of \textit{Survival Game} in Recommendation System. The first two rows show performance on Product Recommendation and Movie Recommendation. The final row shows the performance of SASRec on another four recommendation tasks. The red line is a power law curve drawn based on the distribution of the model's data points. Its exponent roughly represents the model's failure decay rate. Results indicate that models are at Limited Level.}
        \label{fig:recsys_results}
    \end{figure*}

After examining the results of search engines, let’s turn our attention to another widely used AI application: recommendation systems. Recommendation systems predict what a user likes based on past behavior and profile information. These systems have extensive applications in areas such as e-commerce, short videos, etc. 

We adopt many real-world datasets from a wide range of domains. 
We use the Amazon Beauty dataset~\citep{he2016ups} to represent users' preference in e-commerce recommendations. It focuses on skincare product recommendations on the Amazon platform. 
We use MovieLens~\citep{harper2015movielens} to represent movie recommendations. It is constructed based on user ratings of movies. 
We use Steam dataset~\citep{kang2018self} to represent users' preference in game recommendations. It recommends games to players on the Steam platform.
We use Douban Book~\citep{zhu2020graphical, zhu2019dtcdr} to represent book recommendations. Douban is a popular Chinese internet platform and this dataset is to recommend books to users.
We use Douban Music~\citep{zhu2020graphical, zhu2019dtcdr} to represent users' preference in music recommendations. It is also collected from the Douban platform and is to recommend music to users.
Finally, we use Gowalla dataset~\citep{cho2011friendship} to represent location recommendations. Gowalla is a location-based online social network application where users share their check-in location. The dataset is to recommend places users might like to visit. 

We test four widely recognized recommendation methods. 
The first is a popularity-based recommendation method. As the name suggests, it ranks items based on their popularity and recommends them accordingly. Although it is straightforward, it is effective and commonly used in real-world applications. 
The other three methods are sequential recommendation models: GRU4Rec~\citep{hidasi2015session, hidasi2018recurrent}, SASRec~\citep{kang2018self}, and ComiRec~\citep{cen2020controllable}. They differ in architecture. GRU4Rec employs recurrent neural networks to build user profiles based on the interaction history. SASRec uses attention mechanisms to model how past interactions influence future preferences. ComiRec captures users’ diverse interests with a dynamic modeling approach.

The experimental results are shown in Figure~\ref{fig:recsys_results}. The first row shows product recommendations, and the second row shows movie recommendations. The third row shows the performance of SASRec across different domains. According to the results, on all datasets and for all models, data points fall within the Limited Level region and are far from the Capable Level region. The estimated failure decay rate is even lower than $1$, meaning that the distribution of failure counts has a very heavy tail. In other words, the recommendation system has to try a lot of times before finding the item users like. We believe this poor performance originates from the nature of recommendations. Recommendations do not require explicit input from users and rely solely on historical interactions. Such a lack of explicit information input makes predictions very difficult.

This result has a clear physical meaning: When a recommendation model makes a mistake, it is almost impossible for it to find the correct product through continuous attempts. For users, it means that users will see, in statistical terms, infinite uninterested items before being presented with something they are truly interested in. If a user is disappointed every time they see an item they are not interested in, the current recommendation system will disappoint them countless times.

\subsection{Language}

We have assessed AI models in vision, search, and recommendation tasks. Now, we proceed to language tasks. 
Some studies claim that large language models have already achieved exceptionally high-level intelligence and passed the Turing Test~\citep{biever2023chatgpt, aharoni2024attributions, mei2024turing}.
With \textit{Survival Game}, we can examine their intelligence levels and re-think this conclusion. We will use four tasks for a comprehensive evaluation, including coding, mathematics, question answering, and writing.

\textbf{Experimental Setup}: 
We input the question to large language models and examine the models' correctness in predicting the answer. 
The answer written by humans is regarded as the correct one. 
If the answer contains more than one word, such as writing a math proof or a long passage, we concatenate the question and the first $n$ answer words as the models' input and evaluate the performance in predicting the $n+1$-th answer word.
The number of failure attempts equals the number of words that are scored higher than the correct one. 
For datasets where the answers need to follow a fixed format, such as multiple-choice questions or calculating a number, we provide several examples before the actual question to prompt the model about the required answer format. If we use $m$ examples, we will indicate that this is an $m$-shot result in the figure title. This approach helps the model respond in the specified format and improves accuracy~\citep{brown2020language}.

We evaluate state-of-the-art large language models, including Qwen2.5 series~\citep{qwen2, qwen2.5}, Deepseek V2 16B~\citep{deepseekv2}, and Llama3 72B~\citep{dubey2024llama}. They are state-of-the-art models at their scale and are even competitive compared to models with much larger scale~\citep{guo2025deepseek}. 
Although we cannot run models with more parameters due to limited resources, we will extrapolate our results to a larger scale in Section~\ref{sec:future_prediction}.

\subsubsection{Coding}
\label{sec:code_experiment}

\begin{figure*}[t]
    \subcapraggedrighttrue
    \subcaphangtrue
        \centering
        \subfigure{
            \includegraphics[width=0.23\textwidth]{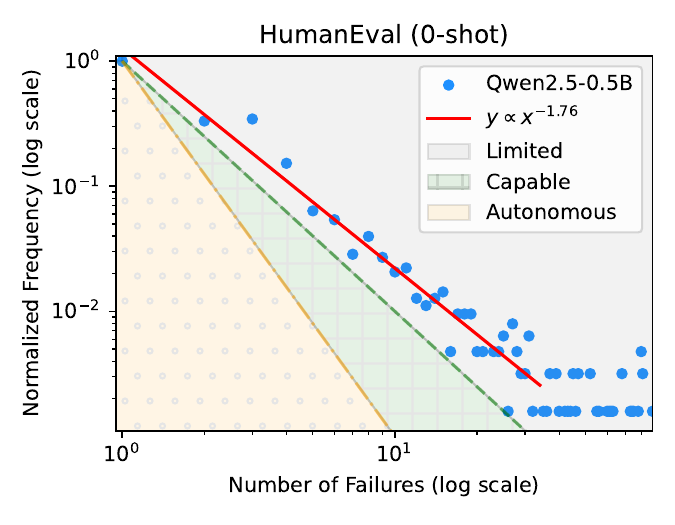}
        }
        \subfigure{
            \includegraphics[width=0.23\textwidth]{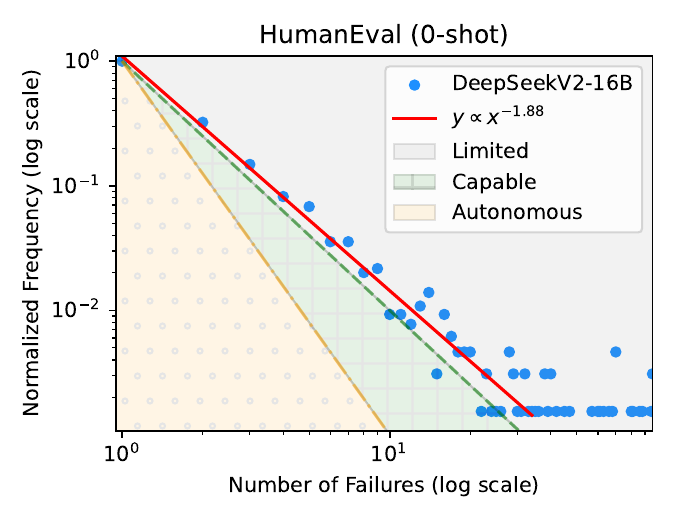}
        }
        \subfigure{
            \includegraphics[width=0.23\textwidth]{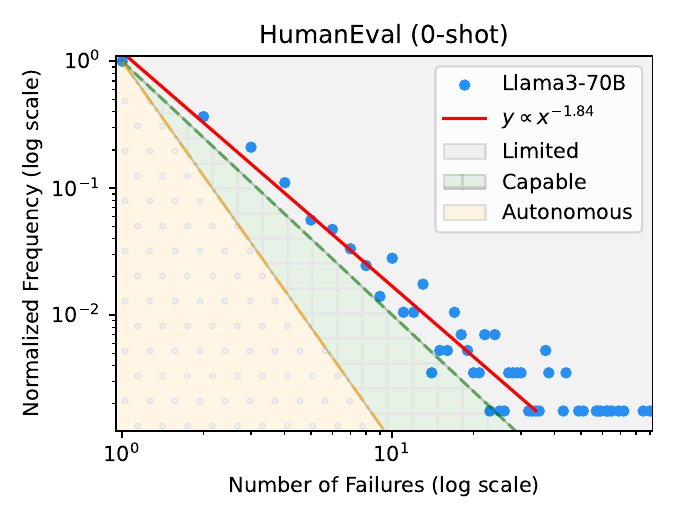}
        }
        \subfigure{
            \includegraphics[width=0.23\textwidth]{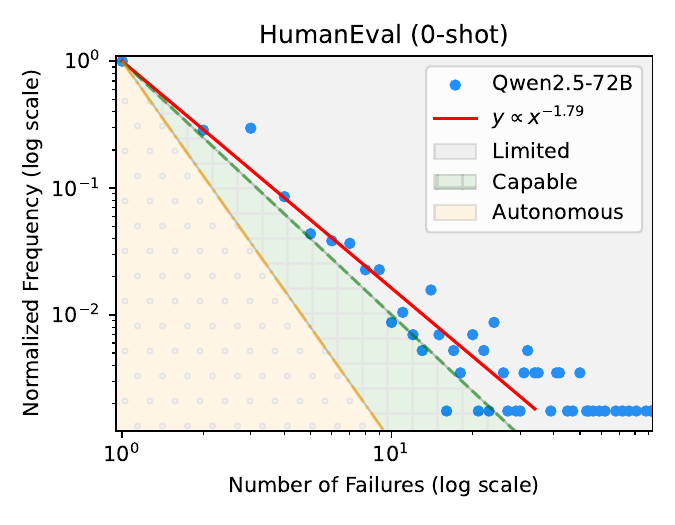}
        }

        \subfigure{
            \includegraphics[width=0.23\textwidth]{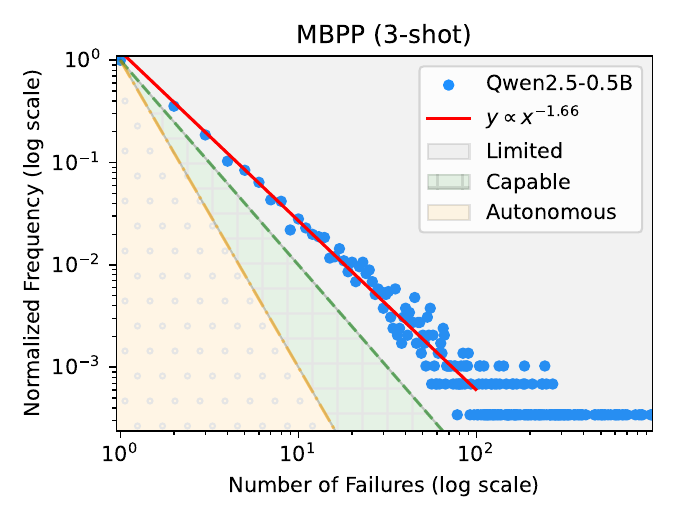}
        }
        \subfigure{
            \includegraphics[width=0.23\textwidth]{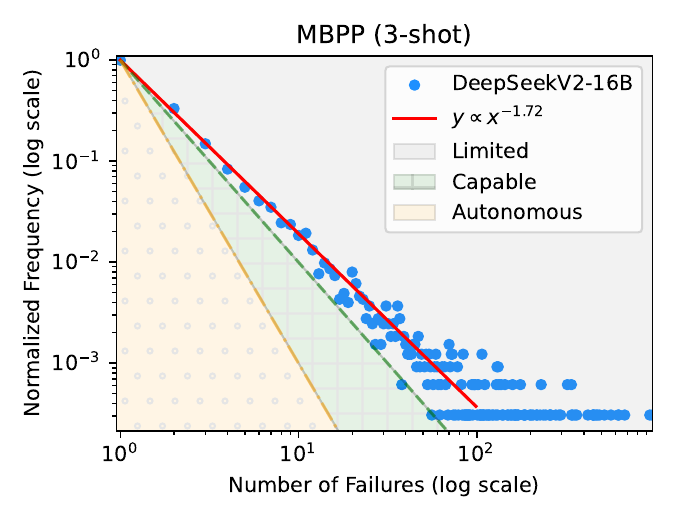}
        }
        \subfigure{
            \includegraphics[width=0.23\textwidth]{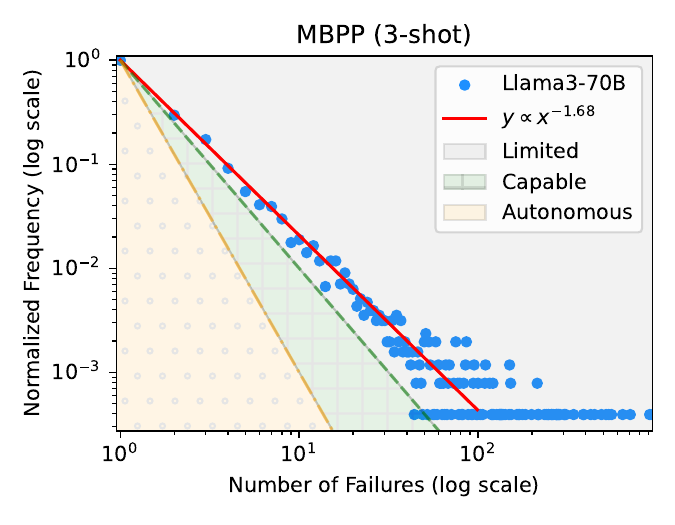}
        }
        \subfigure{
            \includegraphics[width=0.23\textwidth]{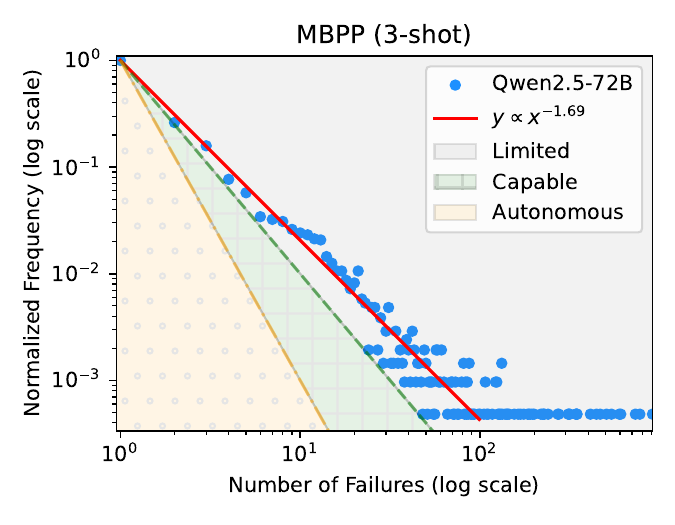}
        }

        \subfigure{
            \includegraphics[width=0.23\textwidth]{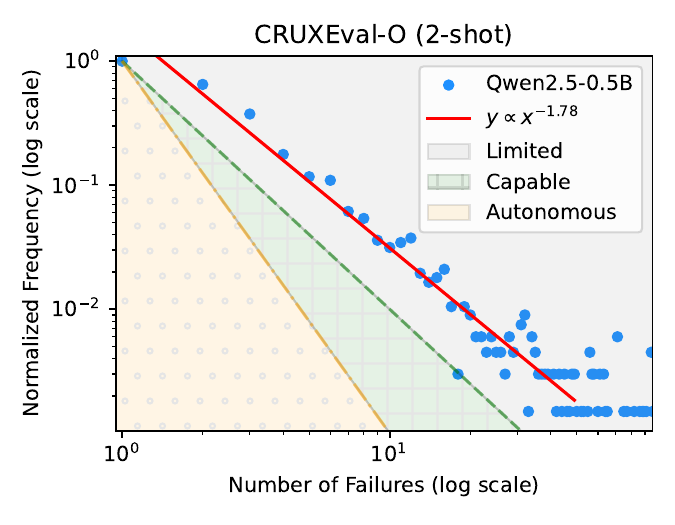}
        }
        \subfigure{
            \includegraphics[width=0.23\textwidth]{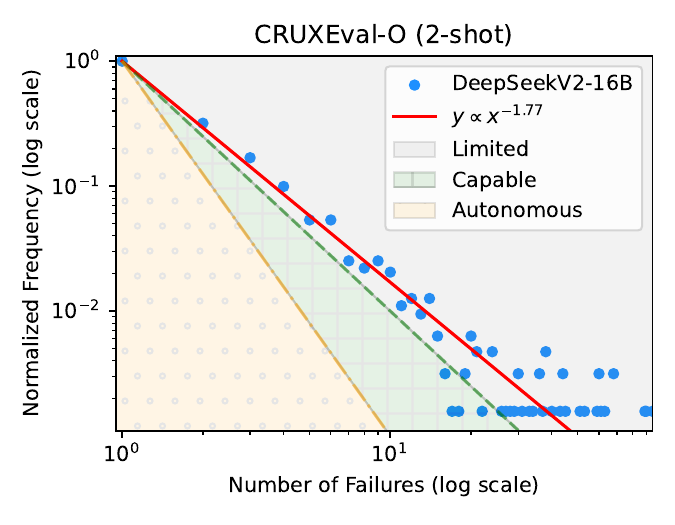}
        }
        \subfigure{
            \includegraphics[width=0.23\textwidth]{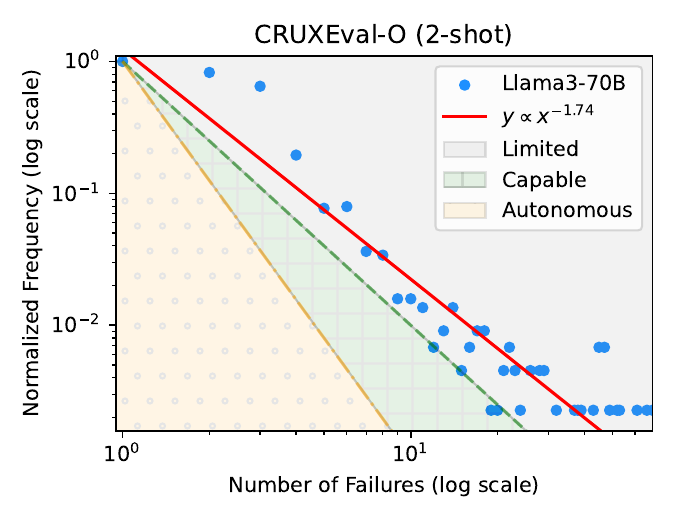}
        }
        \subfigure{
            \includegraphics[width=0.23\textwidth]{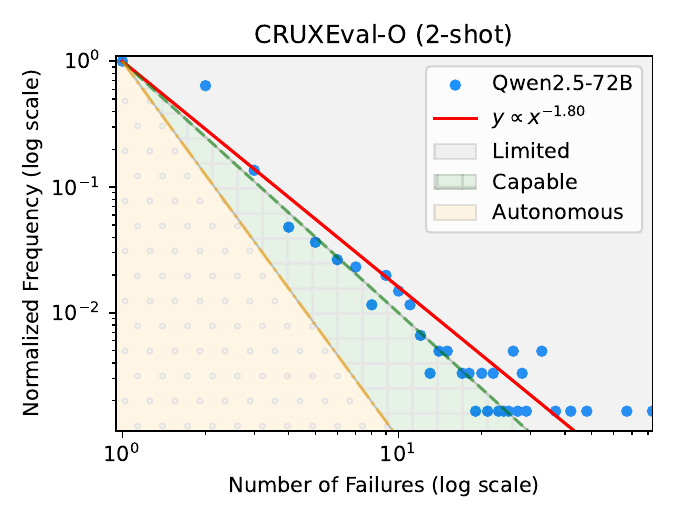}
        }
        \caption{Experimental Results of \textit{Survival Game} in Coding. The three rows correspond to three datasets, and figures in different columns correspond to different models. The red line is a power law curve drawn based on the distribution of the model's data points. Its exponent roughly represents the model's failure decay rate. The results indicate that models are approaching Capable Level.}
        \label{fig:code_results}
    \end{figure*}

We test models' ability to write code. Code has a clear structure, which makes it easier to predict compared to natural language. We use three widely recognized coding benchmarks. All three are designed for beginner-level programming tasks. The first is HumanEval~\citep{chen2021codex}. It provides the function signature as well as docstring and requires subjects to write the function body. The second is MBPP~\citep{austin2021program}. It requires subjects to write functions based on a natural language description. Answers for both HumanEval and MBPP are function definitions. The third is CRUXEval~\citep{gu2024cruxeval}. It requires subjects to understand a function and infer its output for a given input. The answer is usually a code object, such as a string or a list.

The experimental results are shown in Figure~\ref{fig:code_results}. The three rows present results on HumanEval, MBPP, and CRUXEval, respectively. We can see that models with more parameters are closer to the Capable Level. For 70B models, a few data points are already within the Capable Level region, yet a long tail of data points still falls at the Limited Level region. 
Therefore, although current models are relatively strong and approaching the Capable Level in coding,  they are mostly at the Limited Level. It means that they cannot reliably find correct solutions through trial and error for basic coding questions. Thus, human supervision is essential.

\subsubsection{Mathematics}
\label{sec:math_experiment}

\begin{figure*}[t]
    \subcapraggedrighttrue
    \subcaphangtrue
        \centering
        \subfigure{
            \includegraphics[width=0.23\textwidth]{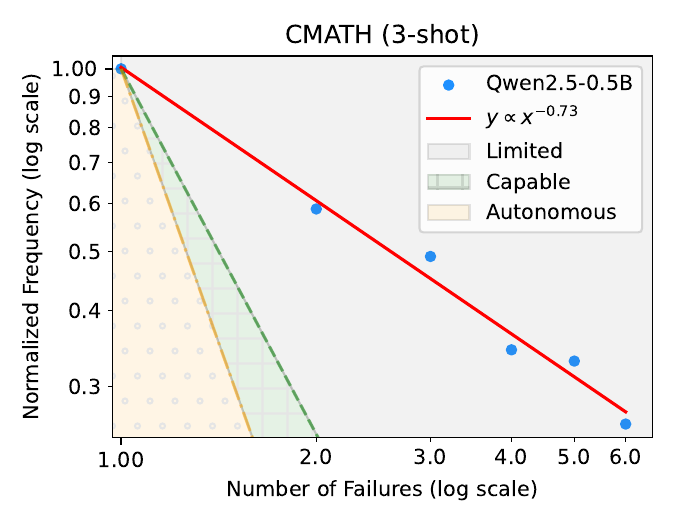}
        }
        \subfigure{
            \includegraphics[width=0.23\textwidth]{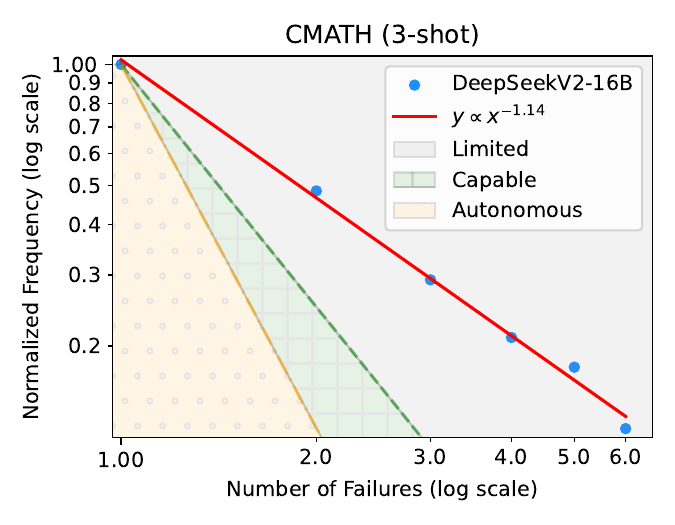}
        }
        \subfigure{
            \includegraphics[width=0.23\textwidth]{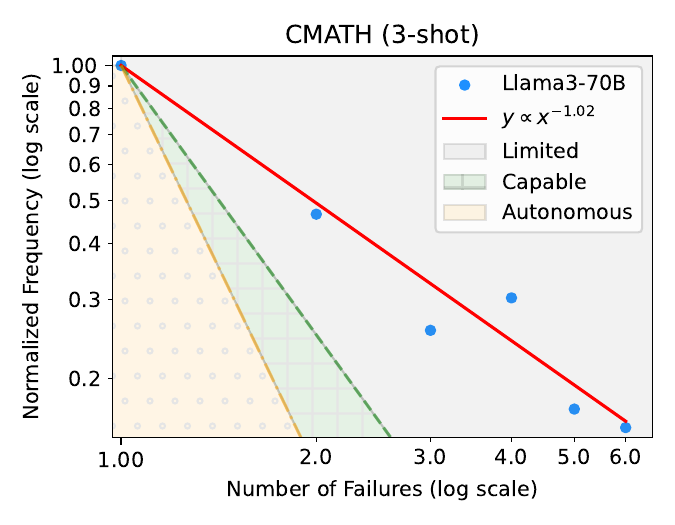}
        }
        \subfigure{
            \includegraphics[width=0.23\textwidth]{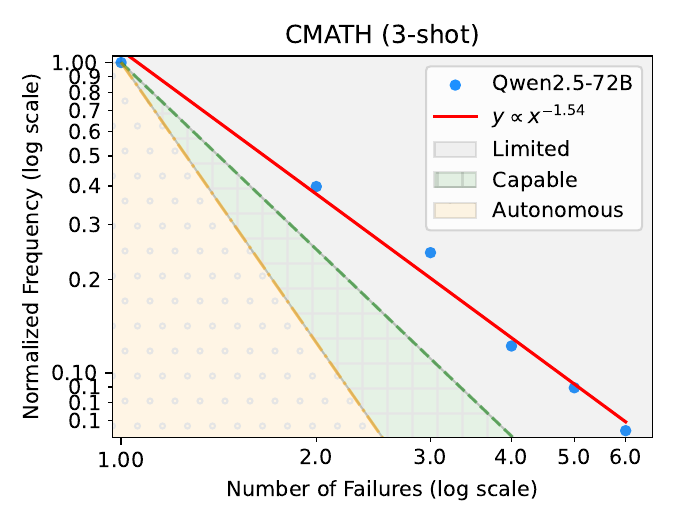}
        }

        \subfigure{
            \includegraphics[width=0.23\textwidth]{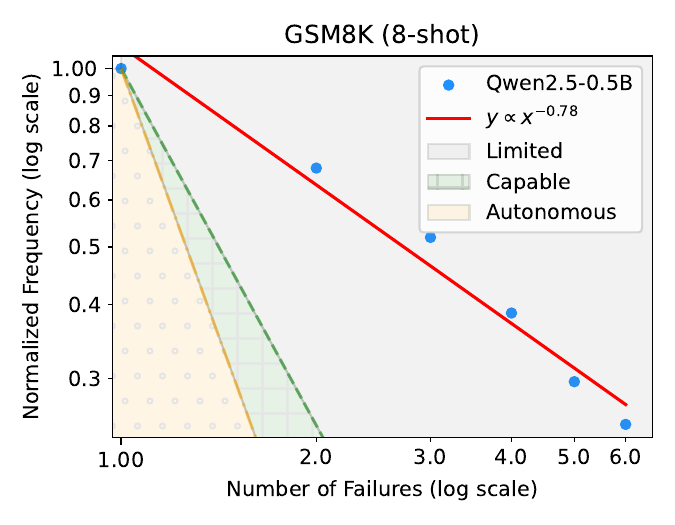}
        }
        \subfigure{
            \includegraphics[width=0.23\textwidth]{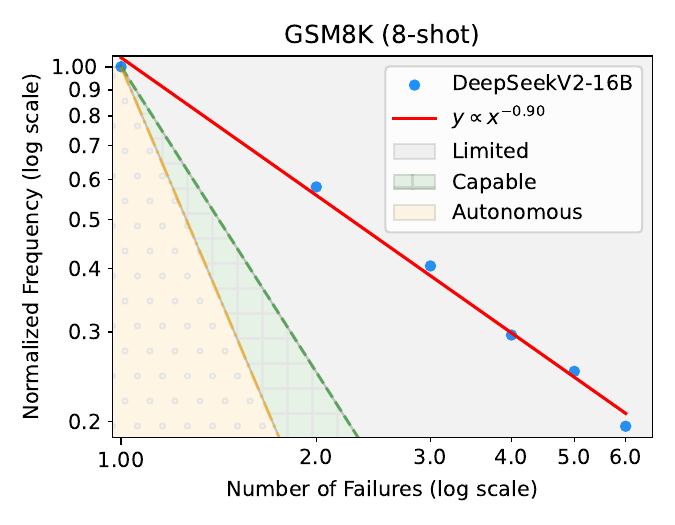}
        }
        \subfigure{
            \includegraphics[width=0.23\textwidth]{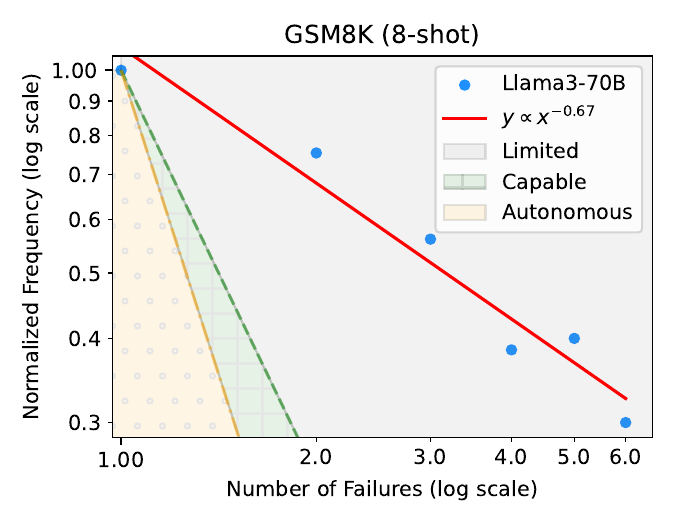}
        }
        \subfigure{
            \includegraphics[width=0.23\textwidth]{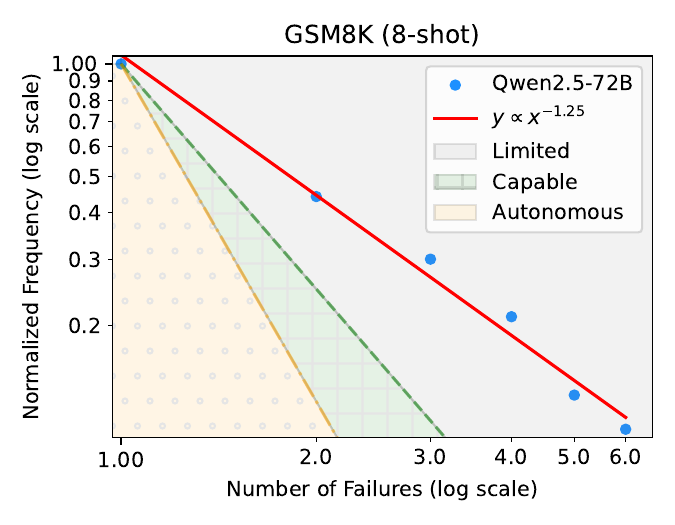}
        }

        \subfigure{
            \includegraphics[width=0.23\textwidth]{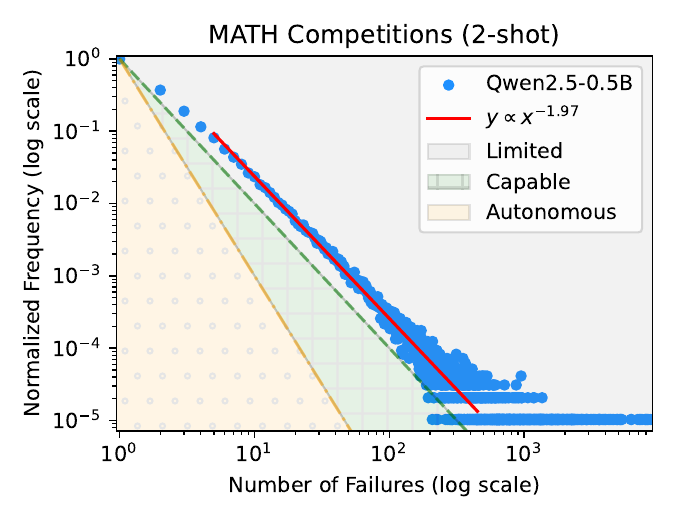}
        }
        \subfigure{
            \includegraphics[width=0.23\textwidth]{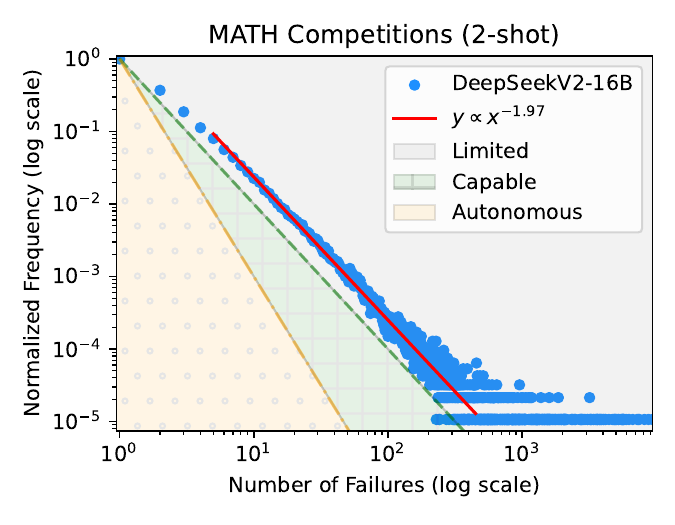}
        }
        \subfigure{
            \includegraphics[width=0.23\textwidth]{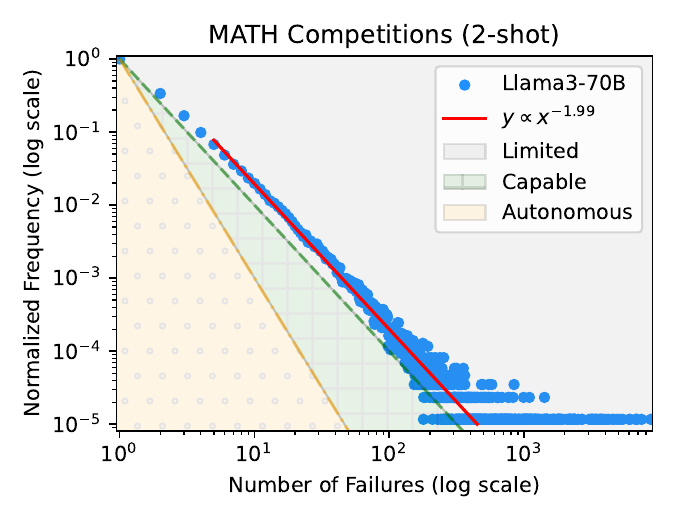}
        }
        \subfigure{
            \includegraphics[width=0.23\textwidth]{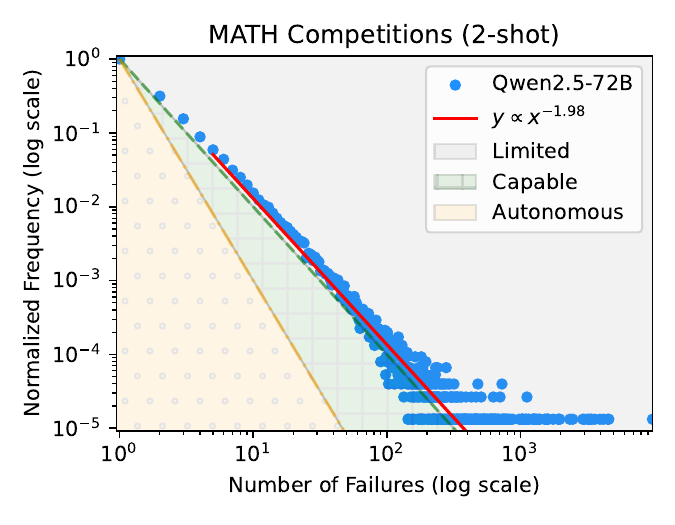}
        }
        \caption{Results of \textit{Survival Game} on Mathematics. The first two rows test the model on simple addition and subtraction of two-digit numbers, while the last row tests whether the model can provide the solution process for a math competition problem. The red line is a power law curve drawn based on the distribution of the model's data points. Its exponent roughly represents the model's failure decay rate. Results suggest that models are better at reasoning through complex problems than performing simple addition and subtraction.}
        \label{fig:math_results}
    \end{figure*}

Next, we test models in another structured domain, namely mathematics. We use three popular datasets. The first is CMath~\citep{wei2023cmath}. It is a Chinese dataset that focuses on elementary school-level math problems. It requires subjects with the ability of addition and subtraction. The second is GSM8K~\citep{cobbe2021gsm8k}. It is an English dataset with similar problems to CMath. For these two datasets, the correct answer that the model needs to output is a number, usually no more than two digits. The third is MATH competition dataset~\citep{hendrycksmath2021}. It contains complex math problems derived from math competitions. The answers to these math problems are usually a long text, such as a mathematical proof or the step-by-step process of solving the problem. 

The experimental results are shown in Figure~\ref{fig:math_results}. From top to bottom, the rows correspond to CMath, GSM8K, and MATH. We can see that on the first two datasets, models are far away from the Capable Level. Thus, current models can hardly perform basic addition and subtraction. In contrast, results in the third row suggest that models are relatively strong in Math Competitions. The data points are approaching the Capable Level. In summary, the models have difficulty solving simple elementary school math problems, yet they perform much better on complex competition-level math problems. This reflects a significant difference between AI and human intelligence. Besides, we also observe that as the model size increases, there is a clear trend of moving closer to the Capable Level.

Therefore, we should exercise caution when using large language models to solve mathematical problems. Although they might solve some complex math questions, they still make significant errors on basic math problems that are easy for humans. In general, current models are at Limited Level. This means that they require a large amount of trials before finding the correct solutions. Thus, it is always necessary to validate their outputs.

\subsubsection{Question-Answering}

\begin{figure*}[t]
    \subcapraggedrighttrue
    \subcaphangtrue
        \centering
        \subfigure{\includegraphics[width=0.23\textwidth]{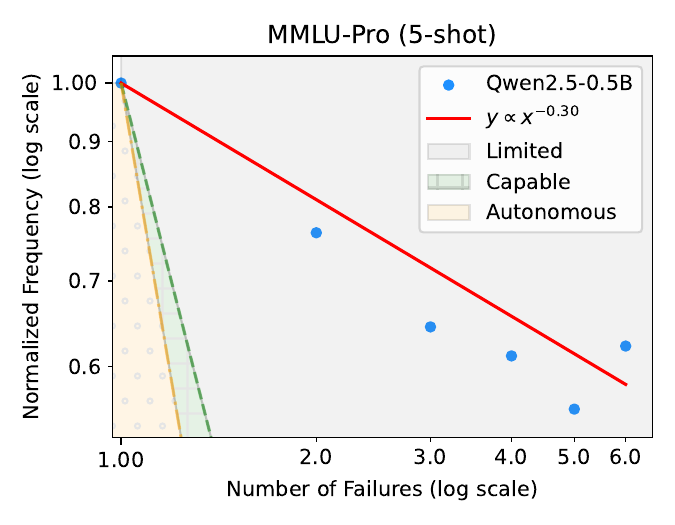}}
        \subfigure{\includegraphics[width=0.23\textwidth]{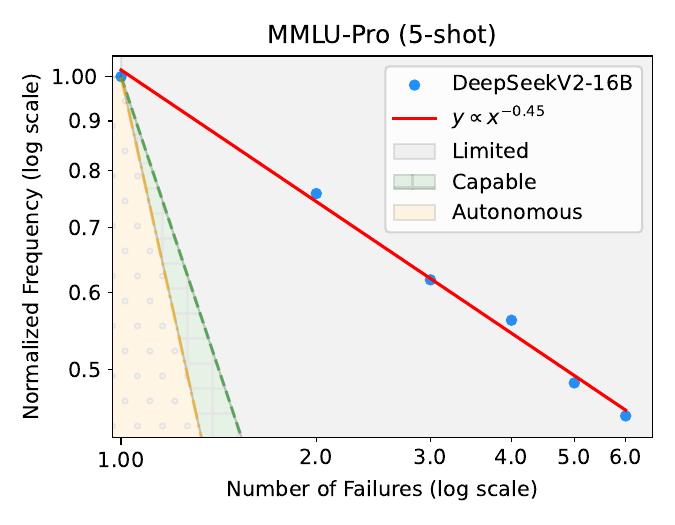}}
        \subfigure{\includegraphics[width=0.23\textwidth]{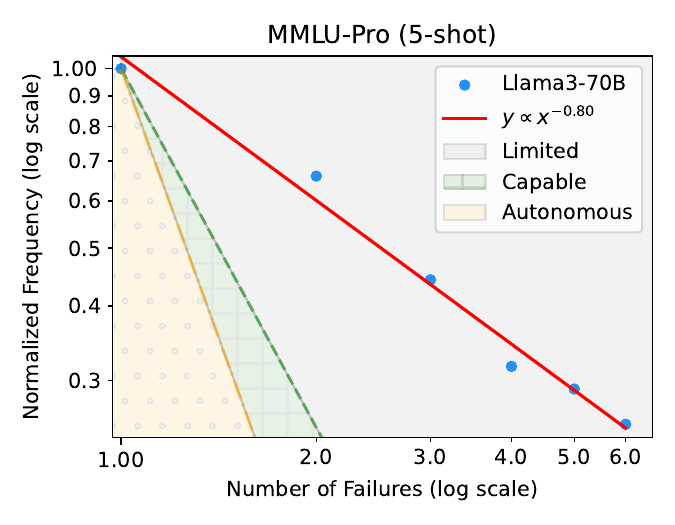}}
        \subfigure{\includegraphics[width=0.23\textwidth]{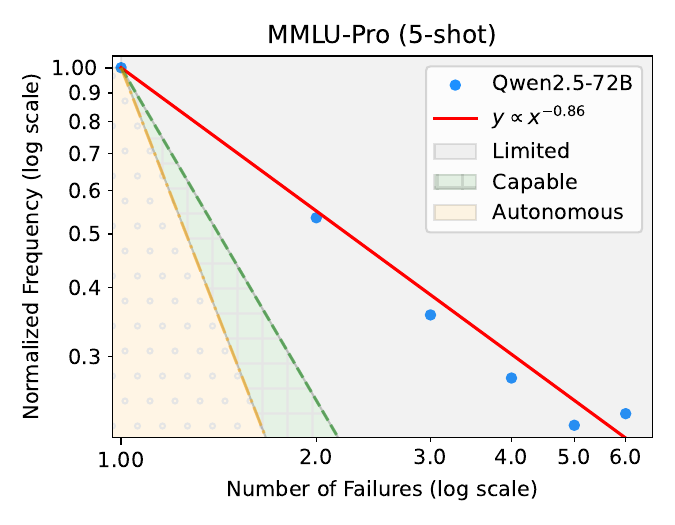}}

        \subfigure{\includegraphics[width=0.23\textwidth]{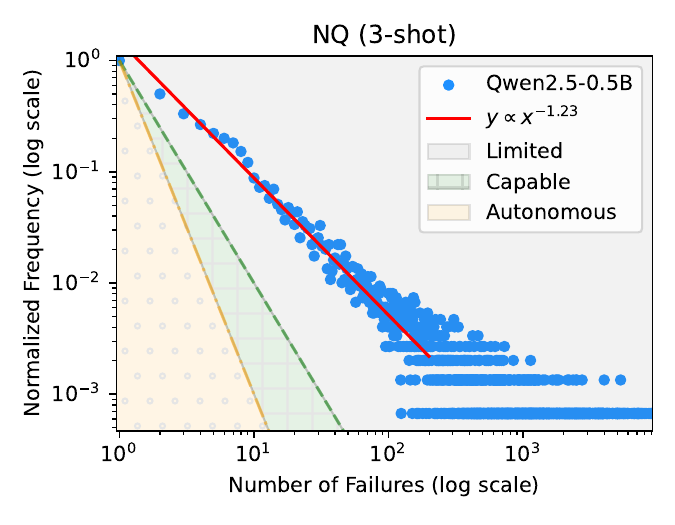}}
        \subfigure{\includegraphics[width=0.23\textwidth]{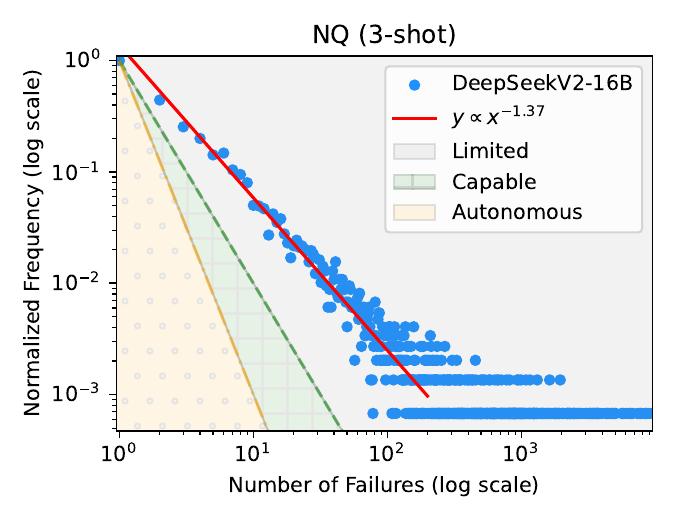}}
        \subfigure{\includegraphics[width=0.23\textwidth]{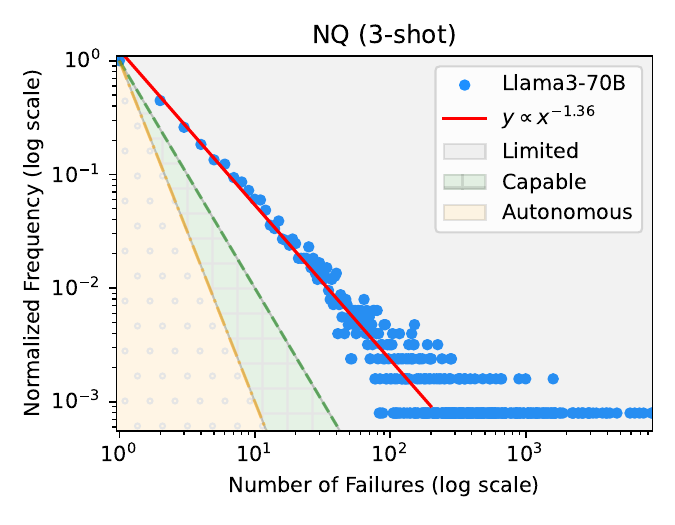}}
        \subfigure{\includegraphics[width=0.23\textwidth]{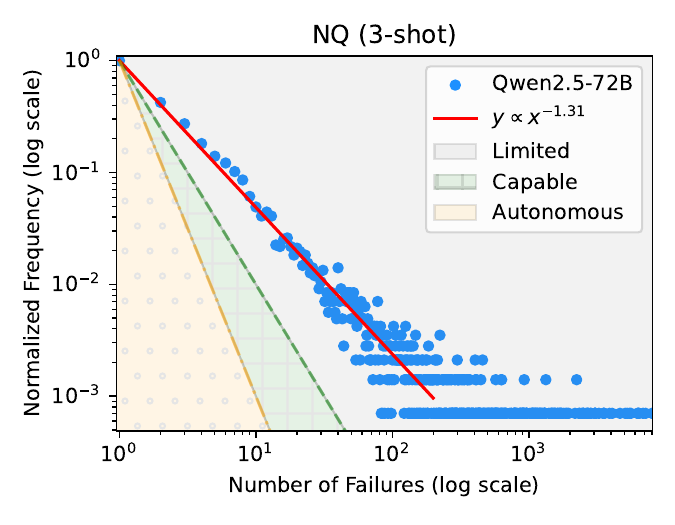}}
        
        \subfigure{\includegraphics[width=0.23\textwidth]{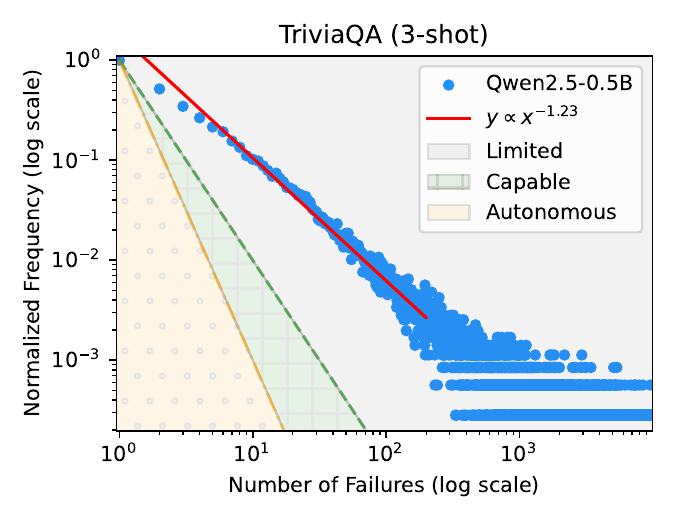}}
        \subfigure{\includegraphics[width=0.23\textwidth]{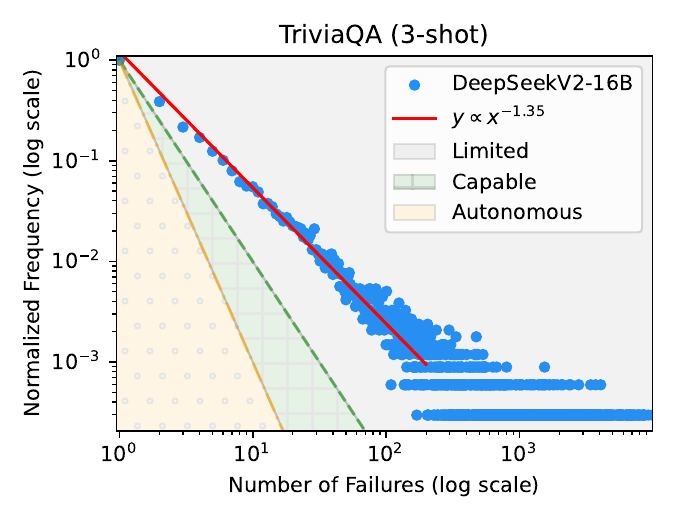}}
        \subfigure{\includegraphics[width=0.23\textwidth]{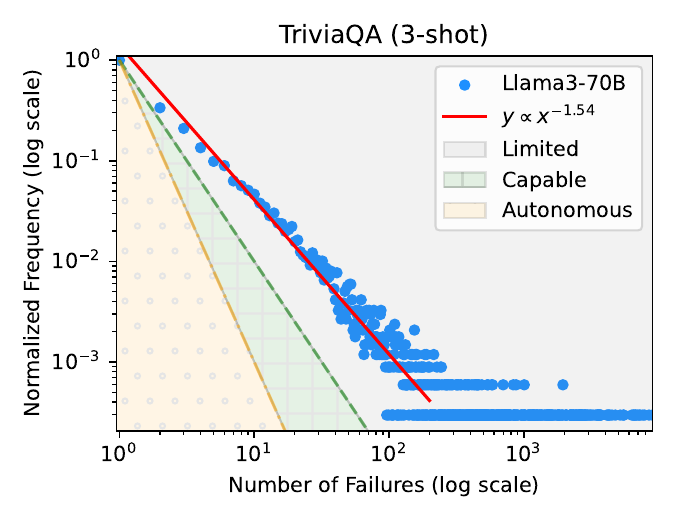}}
        \subfigure{\includegraphics[width=0.23\textwidth]{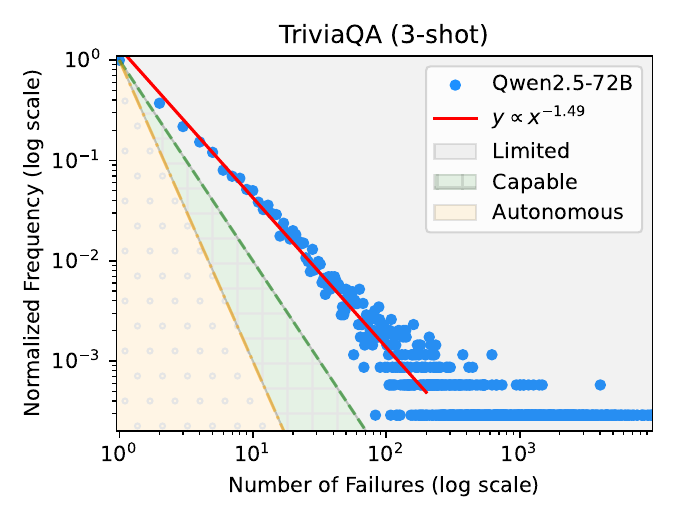}}
        \caption{Experimental Results of \textit{Survival Game} in Question-Answering. The three rows correspond to three datasets, and figures in different columns correspond to different models. The red line is a power law curve drawn based on the distribution of the model's data points. Its exponent roughly represents the model's failure decay rate. Results indicate that all models remain at Limited Level.}
        \label{fig:qa_result}
    \end{figure*}

Next, we examine the models' ability in the Question Answering (QA) task. We select three widely used datasets: MMLU-Pro~\citep{wang2024mmlu}, Natural Questions (NQ)~\citep{lee2019latent, Kwiatkowski2019NaturalQ}, and Trivia QA~\citep{2017arXivtriviaqa}. MMLU-Pro consists of multiple-choice questions across various fields such as mathematics, chemistry, law, etc. The model needs to choose one answer from ten options. NQ is a dataset of real-world questions about factual information. TriviaQA is similar to NQ. Answers in both datasets are only several words long. 

The experimental results are shown in Figure~\ref{fig:qa_result}. The three rows represent MMLU-Pro, NQ, and TriviaQA, respectively. From the results, we observe that all four models are at Limited Level. In MMLU-Pro, the models' failure decay rate is less than $1$. In NQ and TriviaQA, the performance is slightly better than in MMLU-Pro, but the models are still far from reaching the Capable Level. Furthermore, we can see that as the model size increases, the decay rate also increases, gradually moving toward the Capable Level. However, the marginal gains diminish: there is a significant improvement when going from 0.5B to 16B, but then the progress slows down. This suggests that the improvement is sublinear with respect to model size.

Results reflect that question-answering systems can make serious mistakes. In some cases, the systems regard the correct answer as completely incorrect. As a result, we cannot fully trust current question-answering systems, and it is crucial to verify the accuracy of their outputs.

\subsubsection{Writing}

\begin{figure*}[t]
    \subcapraggedrighttrue
    \subcaphangtrue
        \centering
        \subfigure{
            \includegraphics[width=0.23\textwidth]{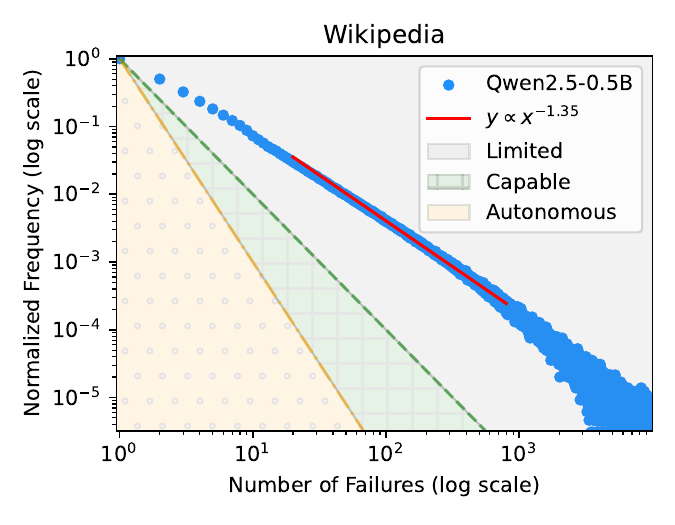}
        }
        \subfigure{
            \includegraphics[width=0.23\textwidth]{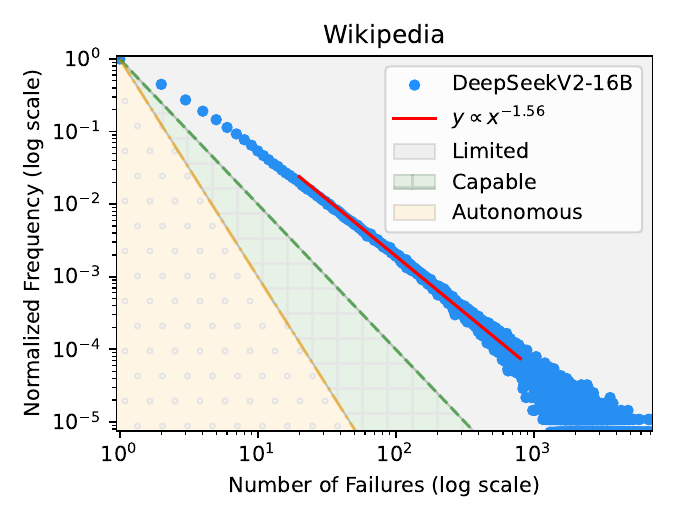}
        }
        \subfigure{
            \includegraphics[width=0.23\textwidth]{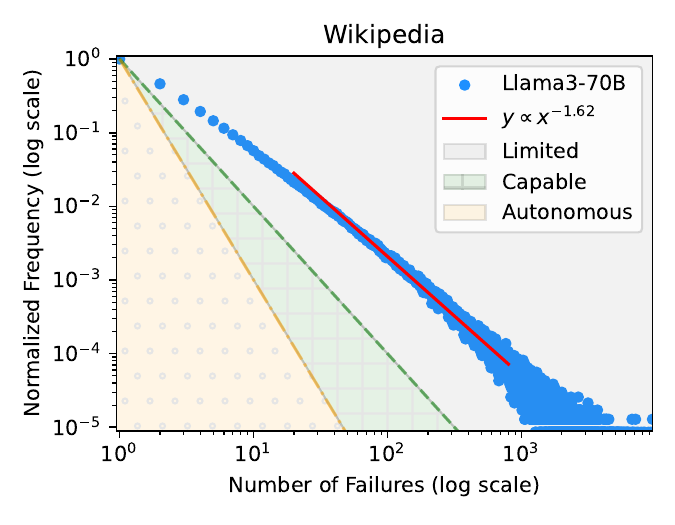}
        }
        \subfigure{
            \includegraphics[width=0.23\textwidth]{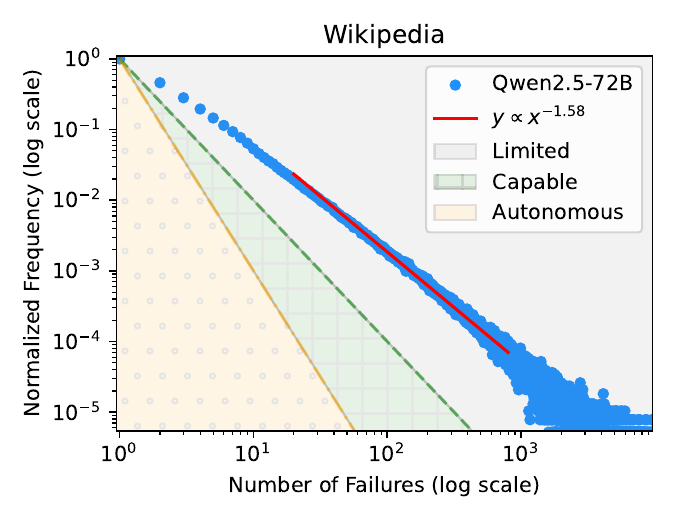}
        }

        \subfigure{
            \includegraphics[width=0.23\textwidth]{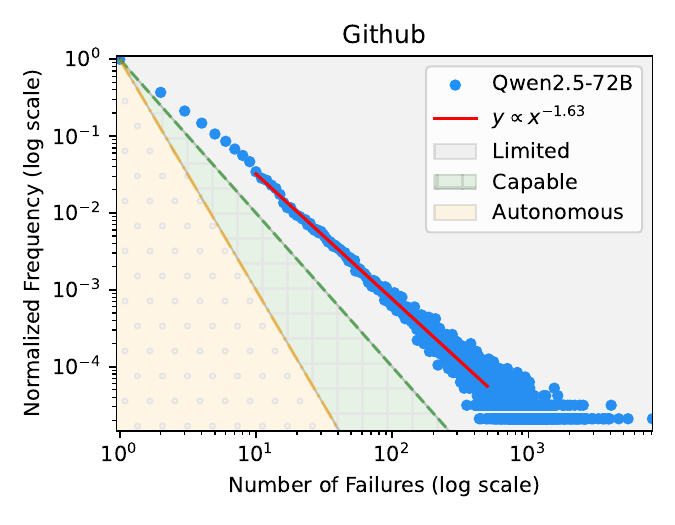}
        }
        \subfigure{
            \includegraphics[width=0.23\textwidth]{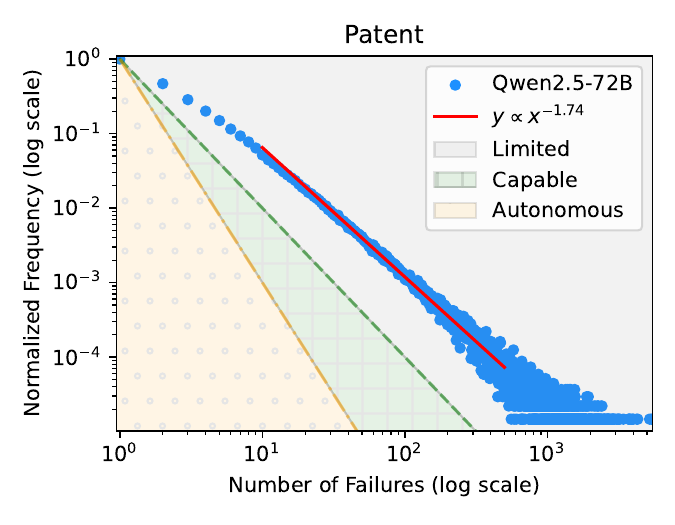}
        }
        \subfigure{
            \includegraphics[width=0.23\textwidth]{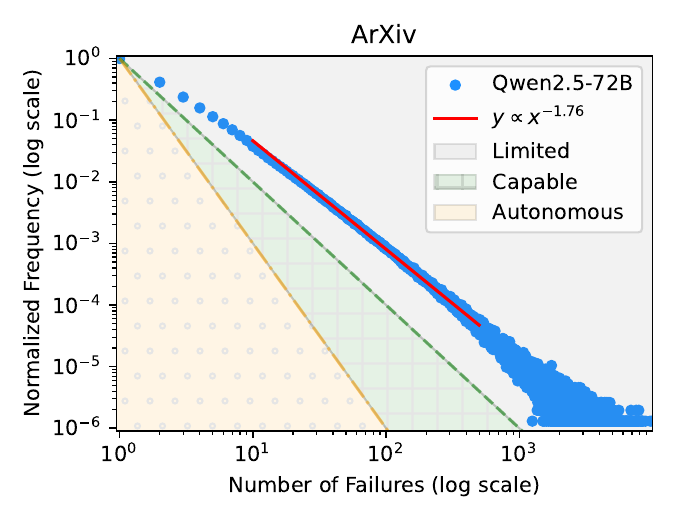}
        }
        \subfigure{
            \includegraphics[width=0.23\textwidth]{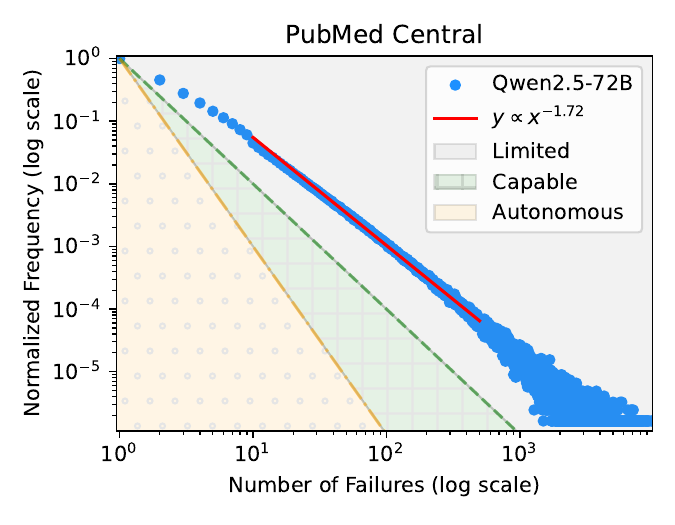}
        }

        \subfigure{
            \includegraphics[width=0.23\textwidth]{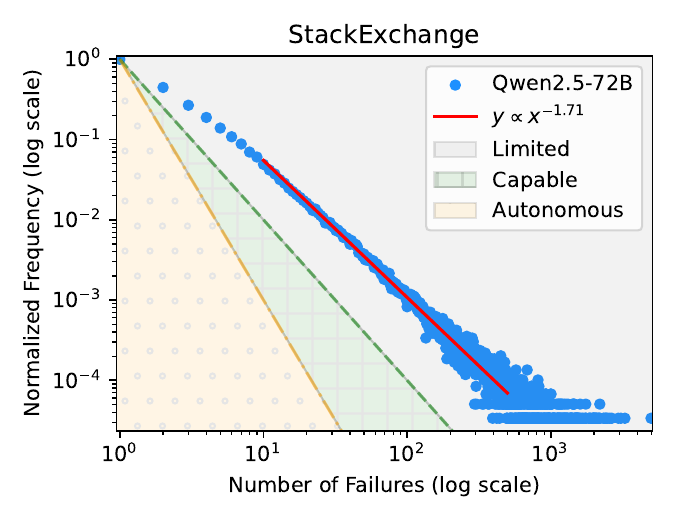}
        }
        \subfigure{
            \includegraphics[width=0.23\textwidth]{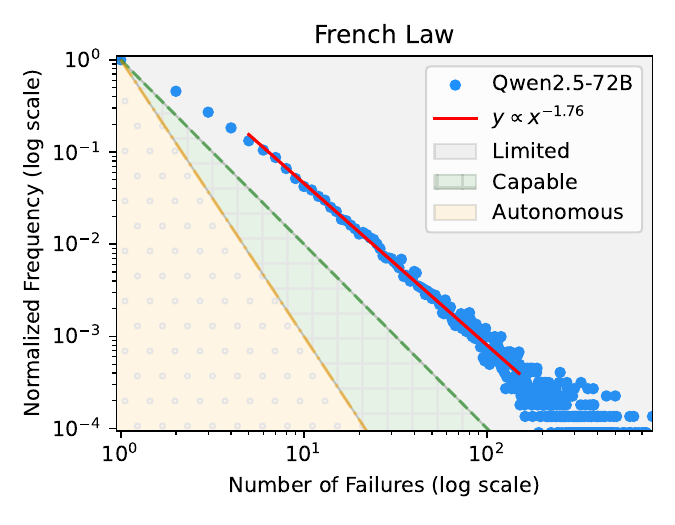}
        }
        \subfigure{
            \includegraphics[width=0.23\textwidth]{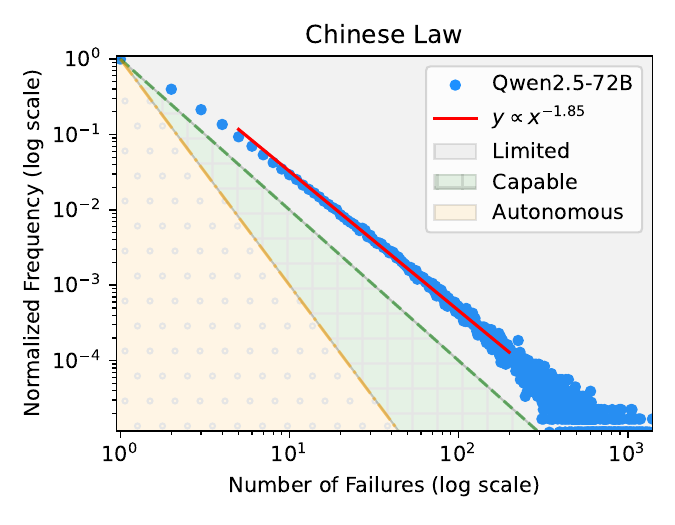}
        }
        \subfigure{
            \includegraphics[width=0.23\textwidth]{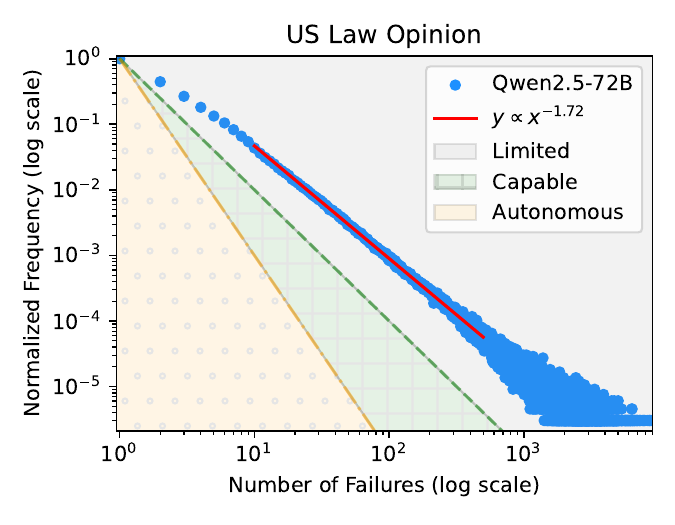}
        }
        \caption{Results of \textit{Survival Game} for Writing in Different Domains. The first row shows language modeling results on Wikipedia. The other two rows present QWen2.5 72B’s writing performance in other domains. The red line is a power law curve drawn based on the distribution of the model's data points. Its exponent roughly represents the model's failure decay rate. Results indicate that all models are at Limited Level.}
        \label{fig:language_domain_results}
    \end{figure*}

Now, we evaluate general writing ability. Based on many human-written articles, we examine whether current systems can also write like humans. During the evaluation, we use the first $n$ words as the input and examine whether subjects can accurately predict the $n+1$-th word. The number of failure attempts equals the number of words scored higher than the $n+1$-th word written by humans.
To ensure the model has sufficient context to make its prediction, we only consider cases where the input prefix is long enough, such as when $n \geq 1,000$. Since those AI systems are trained on large amounts of human data to mimic humans' writing, we believe it is appropriate to adopt human's next token as a reference.

First, we test model performance in different domains. Domains include Wikipedia, code (from Github), patent backgrounds, scientific papers (from ArXiv), medical articles (from PubMed), community QA, and legal texts. Most of the domain data is from Pile~\citep{gao2020pile}. For the legal domain, we use legal texts from France, China, and the US. China and France follow civil law systems with codified legal texts, and we examine whether models accurately memorize them. The data is from \citet{HFforLegal2024} and \citet{wang2023chinese}.
The US follows a common law system, and the test examines whether models can write legal opinions by federal and state courts. The data is from the FreeLaw subset in Pile~\citep{gao2020pile}.

The experimental results are shown in Figure~\ref{fig:language_domain_results}. The first row compares the performance of different models on Wikipedia, while the second and third rows show results on other domains. We can see that all models are at Limited Level. The first row illustrates that as model size increases, the slope of the performance curve becomes steeper and the data points move closer to the Capable Level region. Besides, we notice that on the French and Chinese law datasets, models are also at the Limited Level even though the two datasets simply test their memorization ability of regulations. This suggests that memorizing legal texts is not as simple as it sounds. Overall, current models are still in the early stages in terms of writing. It is important to carefully validate their outputs.

\begin{figure*}[t]
    \subcapraggedrighttrue
    \subcaphangtrue
        \centering
        \subfigure{
            \includegraphics[width=0.23\textwidth]{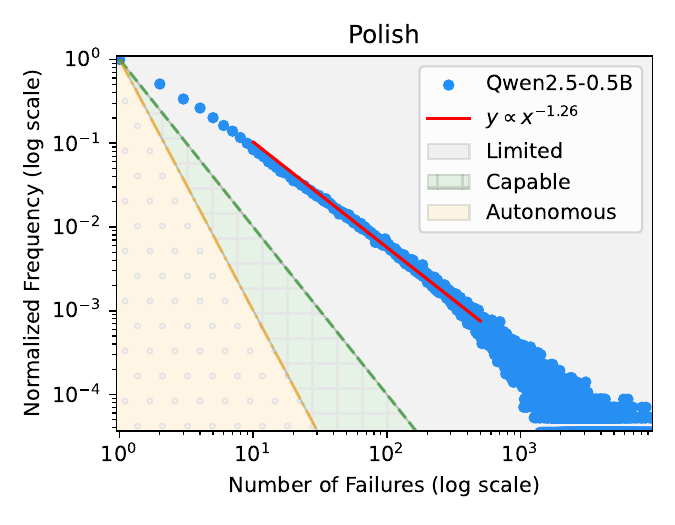}
        }
        \subfigure{
            \includegraphics[width=0.23\textwidth]{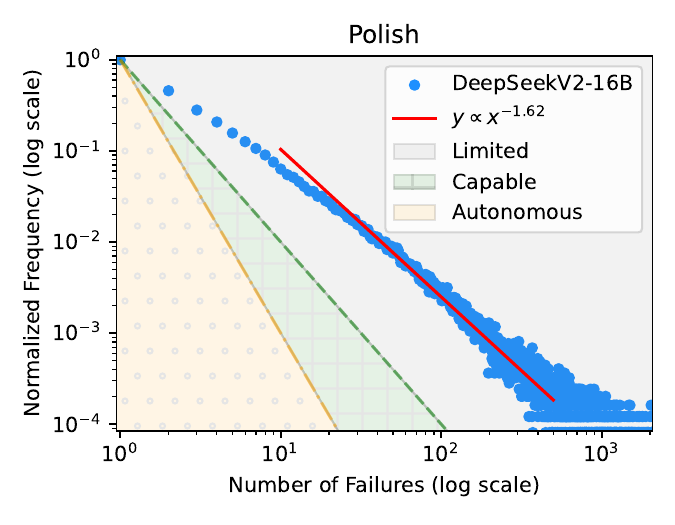}
        }
        \subfigure{
            \includegraphics[width=0.23\textwidth]{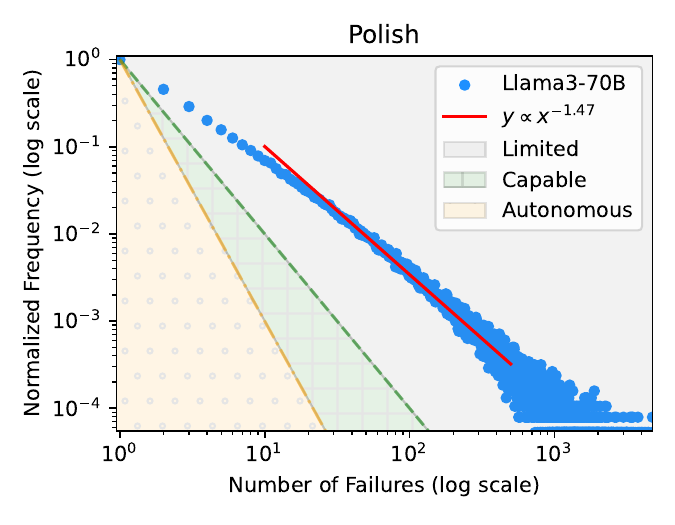}
        }
        \subfigure{
            \includegraphics[width=0.23\textwidth]{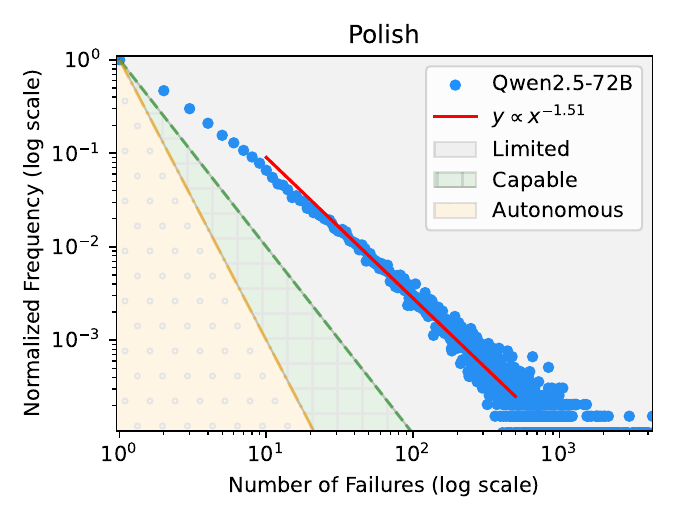}
        }

        \subfigure{
            \includegraphics[width=0.23\textwidth]{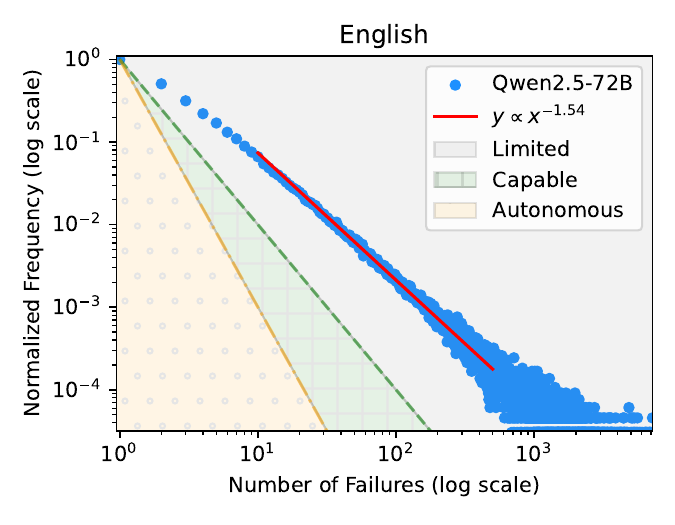}
        }
        \subfigure{
            \includegraphics[width=0.23\textwidth]{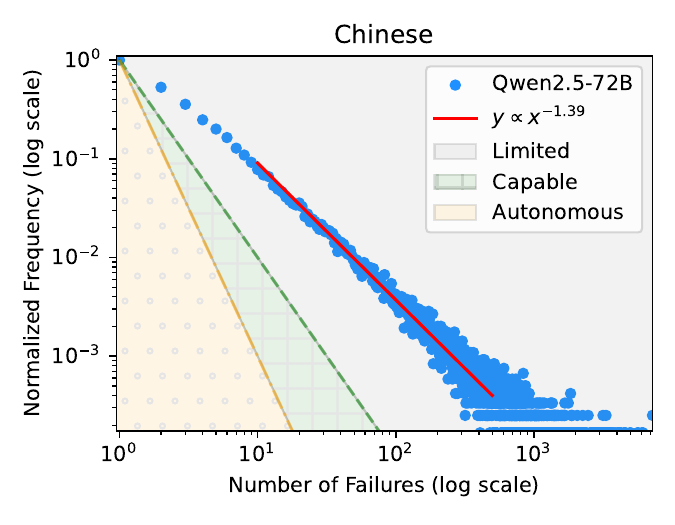}
        }
        \subfigure{
            \includegraphics[width=0.23\textwidth]{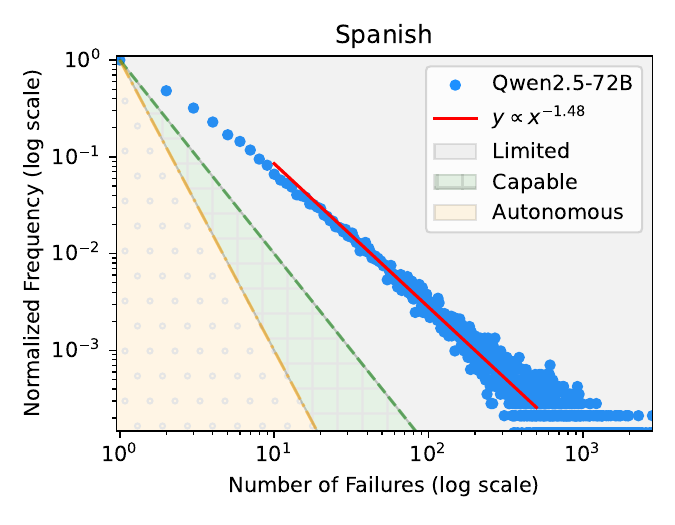}
        }   
        \subfigure{
            \includegraphics[width=0.23\textwidth]{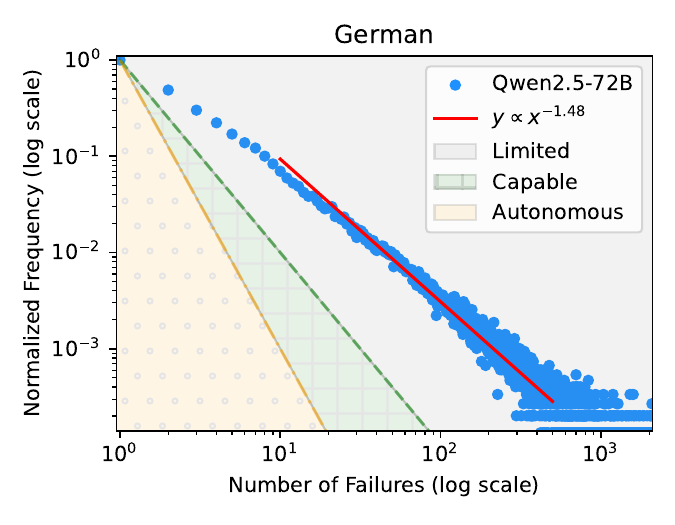}
        }

        \subfigure{
            \includegraphics[width=0.23\textwidth]{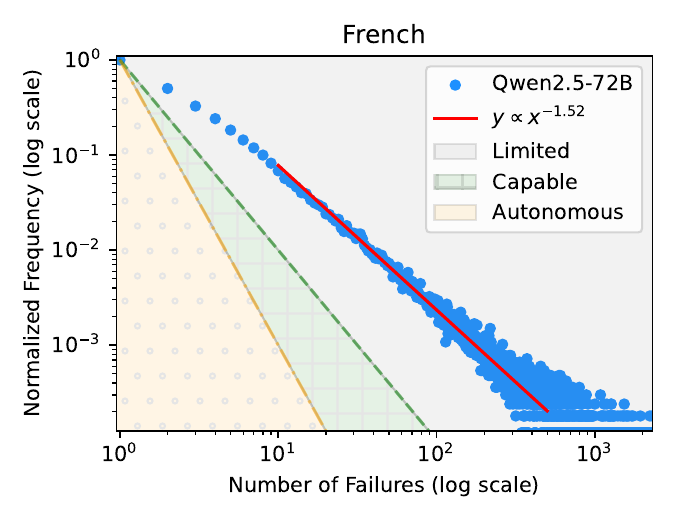}
        }
        \subfigure{
            \includegraphics[width=0.23\textwidth]{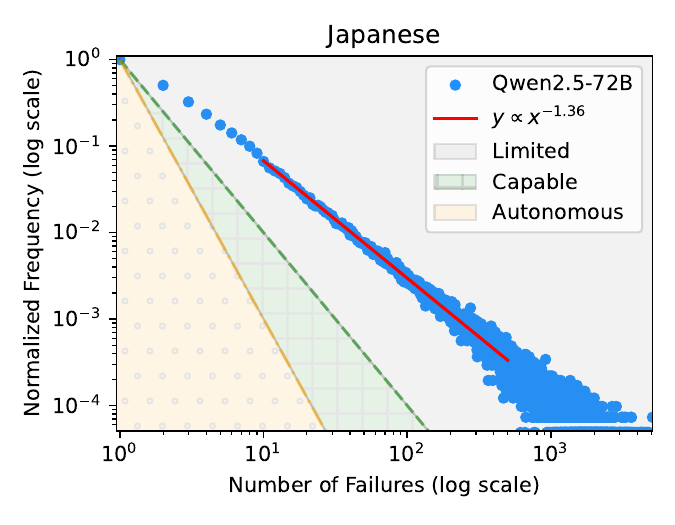}
        }
        \subfigure{
            \includegraphics[width=0.23\textwidth]{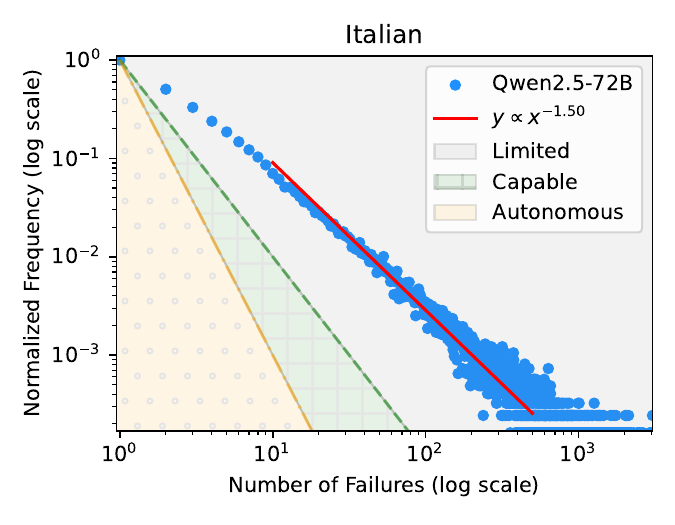}
        }
        \subfigure{
            \includegraphics[width=0.23\textwidth]{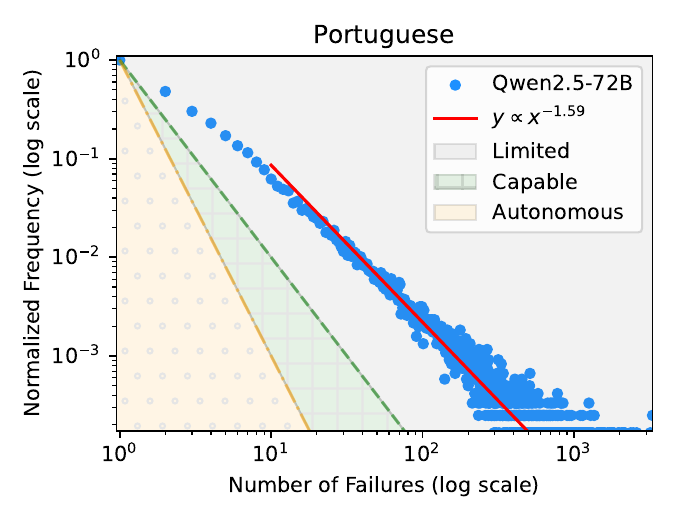}
        }
        \caption{Results of \textit{Survival Game} for Writing in Different Languages. The first row shows language modeling results in Polish. The other two rows present QWen2.5 72B’s writing performance in different languages. The red line is a power law curve drawn based on the distribution of the model's data points. Its exponent roughly represents the model's failure decay rate. Results show that all models lie withat Limited Level.}
        \label{fig:multilingual_results}
    \end{figure*}

Next, we examine whether this result holds across different languages. We test the language modeling capabilities in English, Chinese, Spanish, German, French, Japanese, Italian, Portuguese, and Polish. We use the C4 dataset~\citep{raffel2020exploring}, which consists of a large number of articles from the internet and already categorizes them into different subsets based on the languages. This dataset is commonly used to train and test large language models.

The experimental results are shown in Figure~\ref{fig:multilingual_results}. The first row shows the performance of different models in Polish. The second and third rows show the results for other languages. We can see that across all languages, models are at the Limited Level. It further demonstrates that current language models are at the Limited Level regardless of the language they use.

\subsection{Revisiting Current AI Techniques}

In previous subsections, we evaluate current AI systems in areas such as vision, search, recommendation, and language. We can see that AI remains at the Limited Level. Although this insight was not widely recognized before this study, we find that it has already profoundly impacted existing AI technologies. In other words, current AI technologies are exactly developed in the context of Limited-Level intelligence.

We begin by establishing a connection between AI technology and \textit{Survival Game} through the concept of \textit{loss}. Loss plays a crucial role in AI, especially in deep learning, as it quantifies the degree of error made by an AI system. The smaller the loss, the more advanced the AI system is considered. 
We can see that the concept of loss is very similar to the concept of failure count in \textit{Survival Game}, where failure count quantifies the extent of subjects' errors. The smaller the failure count, the more intelligent the subject is considered. Therefore, we can treat failure count as a form of loss, which we will refer to as ``Survival Game Loss'' in this subsection. 

Survival Game Loss has a strong physical meaning and naturally reflects the performance of AI systems. Yet, most advanced AI systems are stuck at the Limited Level and Survival Game Loss diverges, making directly adopting this loss infeasible. In the following, we demonstrate that many current AI technologies are profoundly related to the divergence of Survival Game Loss, even though these technologies were not explicitly designed or used with this awareness in mind.

\textbf{Hard Negative Sampling}:
Hard negative sampling is a widely used optimization technique in many AI fields, including vision~\citep{shrivastava2016training}, search~\citep{zhan2021optimizing}, recommendation~\citep{ding2020simplify}, and language~\citep{kalantidis2020hard} tasks. It penalizes the model's top-k most incorrect predictions (i.e., hard negatives), rather than punishing all of its wrong predictions (i.e., random negatives). Researchers explain its effectiveness with various hypotheses, such as increasing gradient magnitudes~\citep{xiong2021approximate}, bootstrapping the training data~\citep{shrivastava2016training}, simulating an easy-to-hard curriculum learning process~\citep{chen2021curriculum}, etc. However, from the perspective of Survival Game Loss, its effectiveness becomes easy to understand. Since Hard Negative Sampling focuses on top-$k$ errors, its loss can be seen as $\min (\text{Survival Game Loss}, k)$. By truncating Survival Game Loss with $k$, this approach ensures convergence. This truncation operation gives up on the difficult cases where models fail more than $k$ times before finding the correct solutions. It only optimizes performance in easy cases where the failure count can be smaller than $k$. Ignoring poor performance in difficult cases enables the model to focus on improving accuracy in simple cases. This approach is effective for Limited-Level intelligence, but if the model could reach the Autonomous Level, there would be no need to ignore difficult cases, and this method would not be so effective.

An interesting story about hard negative sampling is that we studied its effectiveness years ago and demonstrated its effectiveness in ignoring difficult cases~\citep{zhan2021optimizing}. Yet, it is only now that we realize how deeply it relates to the essence of intelligence.

\textbf{Cross-Entropy Loss}:
Cross-entropy loss is a commonly used loss function, widely applied across tasks such as vision~\citep{oord2018representation, radford2021learning} and language~\citep{radford2019language, izacard2021unsupervised}. Earlier researchers provided heuristic explanations for its effectiveness, such as making the neural network’s embedding distribution more uniform~\citep{wang2020understanding} and automatically weighting different negatives~\citep{chen2020simple}. Yet, from the perspective of Survival Game Loss, its role becomes clearer. It can be seen as using $\log (\text{Survival Game Loss})$ as the loss. $\log$ transformation leads to better convergence. For instance, if Survival Game Loss follows a power-law distribution with an exponent between 1 and 2, its expectation does not converge, but the $\log$ transformation does. In this way, Cross-Entropy loss helps address the divergence of Survival Game Loss and thus makes the training process more effective. Nevertheless, if models were at the Autonomous Level, it would not be so effective since the loss would already be convergent.

\textbf{Reinforcement Learning} (RL):
RL~\citep{kaelbling1996reinforcement} lets AI systems explore solutions themselves and rewards them when they succeed. It is similar to how animals learn. However, its application is limited because the training cost is prohibitively high~\citep{dulac2021challenges}. We can explain this high training cost based on the convergence of Survival Game Loss at Limited Level. The cost of RL is closely related to the number of failed attempts, which is indeed Survival Game Loss. Therefore, the cost is infinite at Limited Level, making RL infeasible. Recently, DeepSeek-R1~\citep{guo2025deepseek} shows that RL can be applied in mathematical and coding tasks. This is because current advanced models are approaching the Capable Level in the two areas, as reflected by our previous experiments in Section~\ref{sec:code_experiment} and \ref{sec:math_experiment}. 
Models at Capable Level are more likely to find correct answers and their cost in the RL process is much lower. Yet for many other tasks, such as writing, current models are still far from the Capable Level, making RL difficult to apply.

We can see that \textit{Survival Game} provides deep insights for understanding AI techniques. Although earlier researchers mainly designed algorithms based on heuristics without the knowledge of this test, \textit{Survival Game} effectively reveals the fundamental reasons behind their success. With the guidance of \textit{Survival Game}, we believe researchers will easily design more advanced AI techniques in the future.

\section{Scaling in Survival Game}
\label{sec:future_prediction}

In this section, we make predictions about the model size needed to achieve the Autonomous Level in language tasks. 
We begin by introducing the empirical relationship between model size and decay rate in \textit{Survival Game}. Using this relationship, we will extrapolate to even larger model sizes, making predictions about the future trajectory of AI development.

\subsection{Fitting Failure Count Distribution}
\label{sec:fit_failure_count_distribution}

\begin{figure*}[t]
    \subcapraggedrighttrue
    \subcaphangtrue
        \centering
        \subfigure{\includegraphics[width=0.23\textwidth]{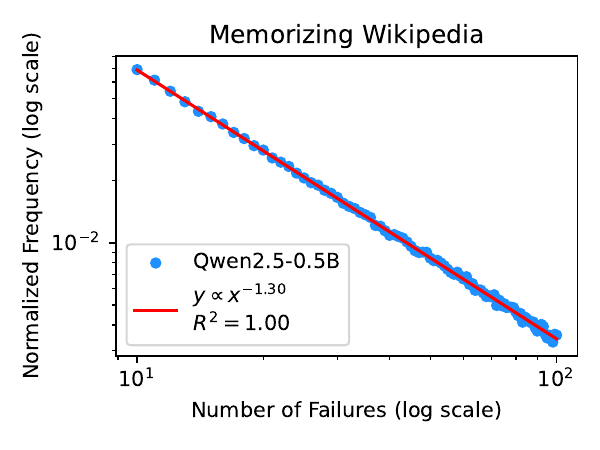}}
        \subfigure{\includegraphics[width=0.23\textwidth]{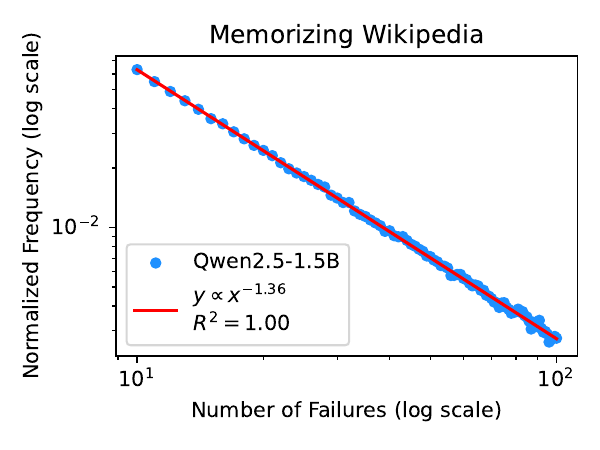}}
        \subfigure{\includegraphics[width=0.23\textwidth]{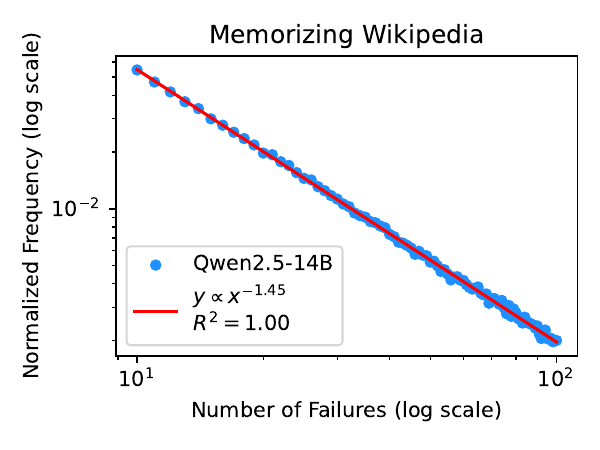}}
        \subfigure{\includegraphics[width=0.23\textwidth]{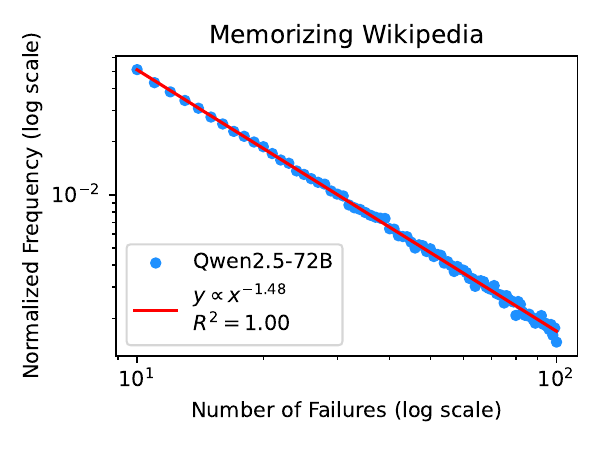}}
        
        \subfigure{\includegraphics[width=0.23\textwidth]{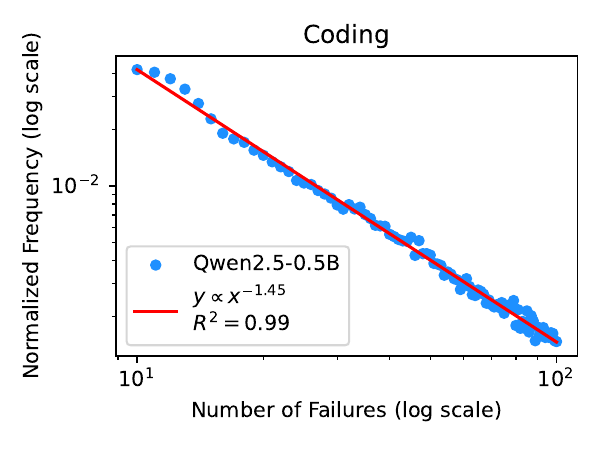}}
        \subfigure{\includegraphics[width=0.23\textwidth]{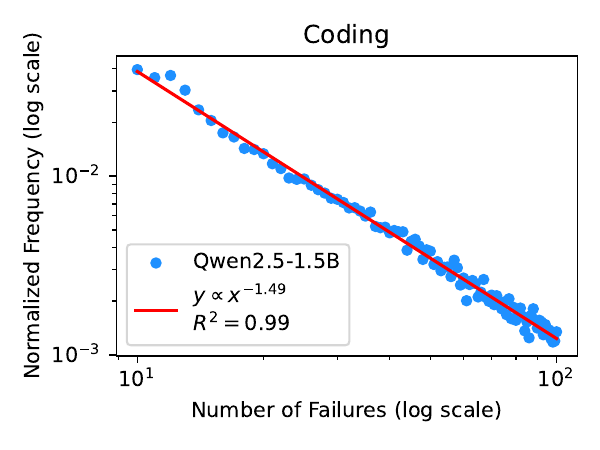}}
        \subfigure{\includegraphics[width=0.23\textwidth]{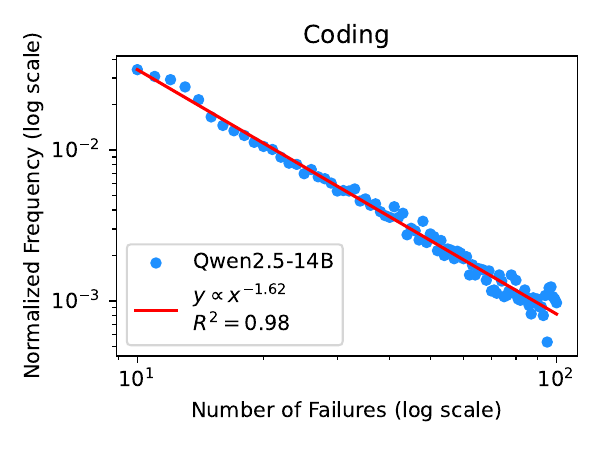}}
        \subfigure{\includegraphics[width=0.23\textwidth]{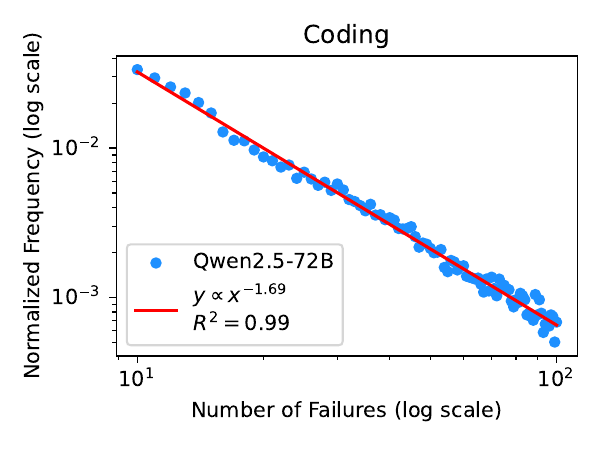}}
        
        \subfigure{\includegraphics[width=0.23\textwidth]{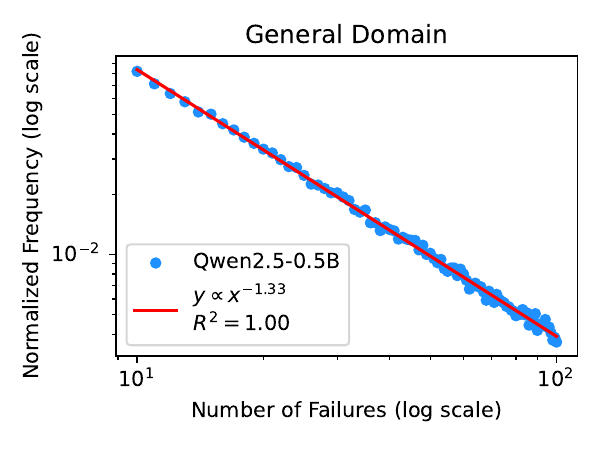}}
        \subfigure{\includegraphics[width=0.23\textwidth]{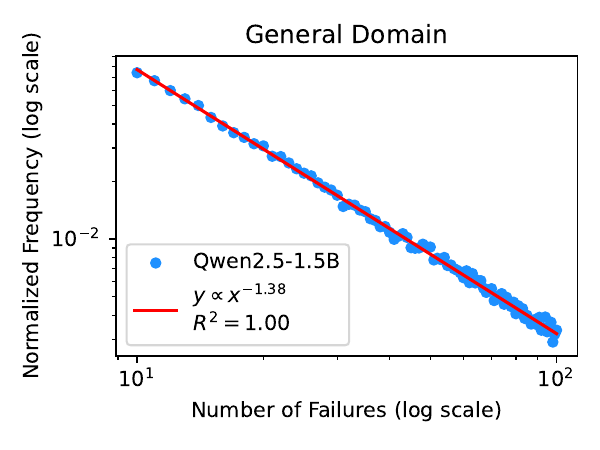}}
        \subfigure{\includegraphics[width=0.23\textwidth]{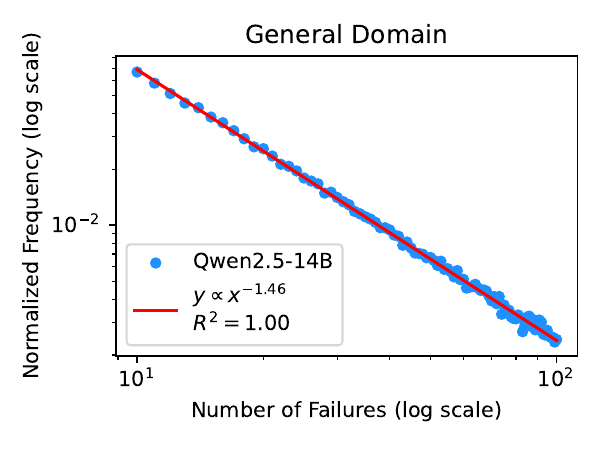}}
        \subfigure{\includegraphics[width=0.23\textwidth]{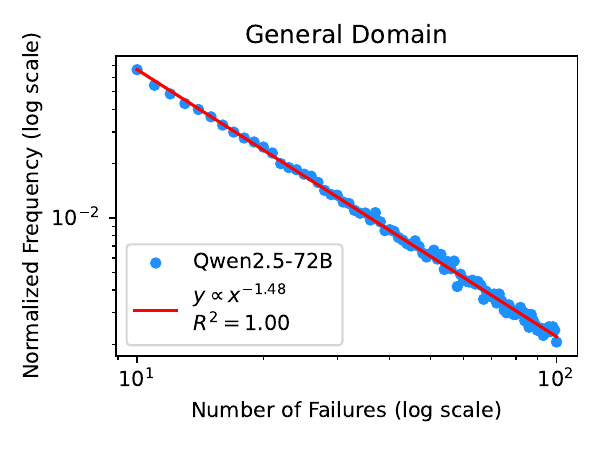}}
        \caption{Distribution of Failure Count in Language Tasks. Failure count ranges from $10$ to $100$. Results show that data closely follows a power law across different domains and model sizes.}
        \label{fig:fit_decay_with_powerlaw}
    \end{figure*}

In this subsection, we aim to quantitatively characterize the performance of different models in \textit{Survival Game}. Take a look at the experimental results from the previous section, such as Figure~\ref{fig:language_domain_results} and \ref{fig:multilingual_results}. The distribution of failure count is close to a straight line in a log-log plot, especially for failure count between $10$ and $100$. When the failure count exceeds $100$, the data points become scattered. When the failure count is less than $10$, the distribution is a curve that bends downward with an increasingly steeper slope. Since a straight line in a log-log plot suggests a power law distribution, this observation suggests that the failure count distribution can be approximated by a power law, especially when the failure count is neither too small nor too large.

Therefore, we use the power law to fit the distribution of the failure count and directly use the exponent obtained from the power law fit as the subject's decay rate. In this way, we further quantify the subject's intelligence from the distribution of failure count to a fitted number of its decay rate. Thus, we no longer use a log-log plot to see where the distribution falls as suggested in Section~\ref{sec:decay_rate_classification}, but instead, directly compare the fitted decay rate with $2$ and $3$. If the decay rate is less than $2$, the subject is classified as Limited Level. If the decay rate is between $2$ and $3$, the subject is classified as Capable Level. If the decay rate is greater than $3$, the subject is classified as Autonomous Level.

We validate the fitting quality of this approach in Figure~\ref{fig:fit_decay_with_powerlaw}. The task is language writing tasks, and the three rows correspond to Wikipedia, code, and general domains, respectively. We evaluate models of different sizes, ranging from 0.5B to 72B. The x-axis represents the failure count, and the y-axis represents the frequency. Both axes are on a logarithmic scale. We fit the distribution within the failure count range of $10$ to $100$. We can see that data points closely follow a power law distribution across domains and model sizes. The $R^2$ values for the fits are marked on the figure and are very close to $1$, suggesting that the fitting quality is near perfect. Thus, we use this setup to obtain the decay rate in this section.

\subsection{Fitting Effect of Scaling}
\label{sec:fit_model_scale_to_decay}

Now, we empirically examine how model sizes relate to failure decay rate. We conduct experiments with models from Qwen 2.5 series~\citep{qwen2, qwen2.5}. The model sizes range from 0.5B to 72B. We use these models because they are state-of-the-art in terms of their respective parameter sizes and can even approach the performance of models that are ten times larger~\citep{guo2025deepseek}. Using such advanced models allows us to draw conclusions that represent the current cutting-edge technology.

\begin{figure*}[t]
    \subcapraggedrighttrue
    \subcaphangtrue
        \centering
        \subfigure{\includegraphics[width=0.23\textwidth]{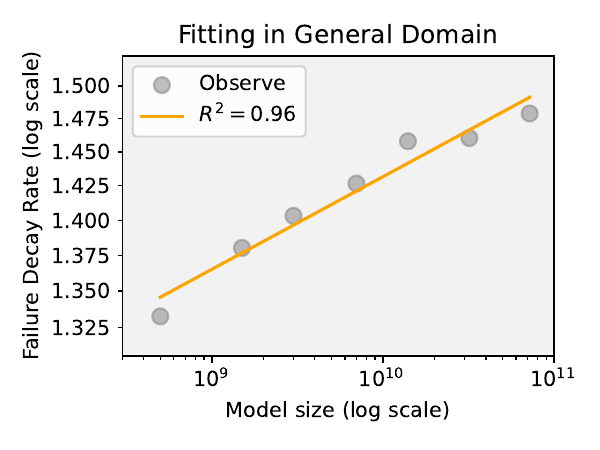}}
        \subfigure{\includegraphics[width=0.23\textwidth]{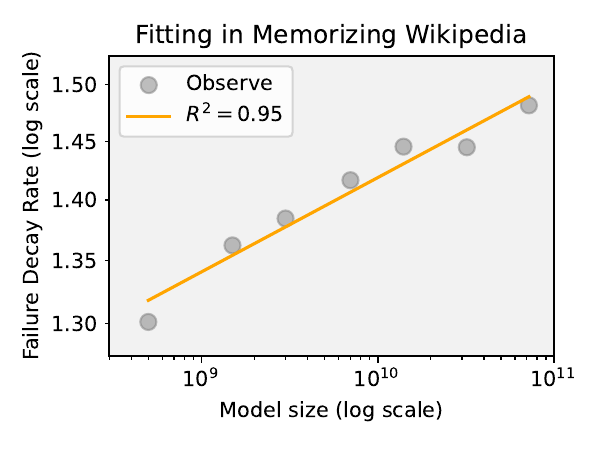}}
        \subfigure{\includegraphics[width=0.23\textwidth]{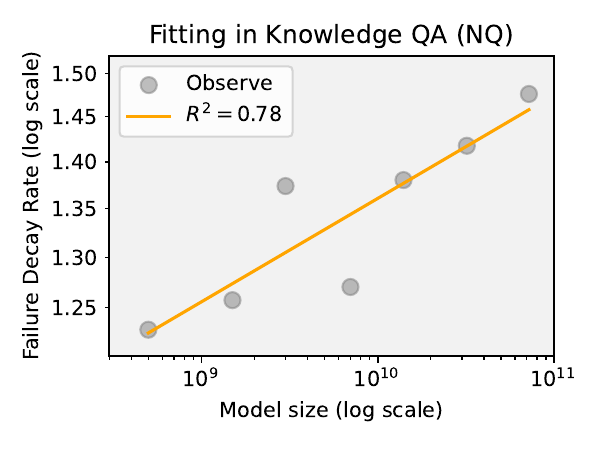}}
		\subfigure{\includegraphics[width=0.23\textwidth]{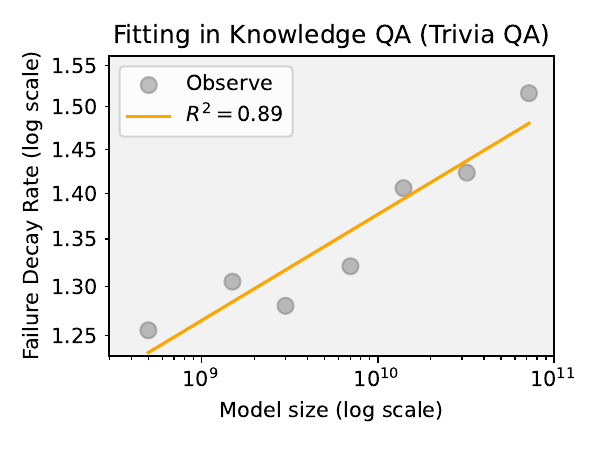}}
		
		\subfigure{\includegraphics[width=0.23\textwidth]{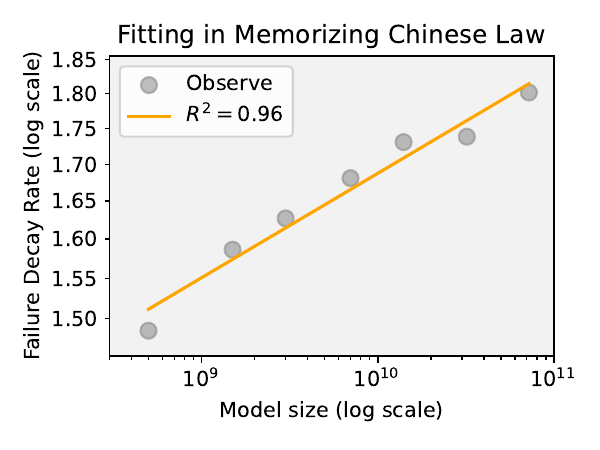}}
		\subfigure{\includegraphics[width=0.23\textwidth]{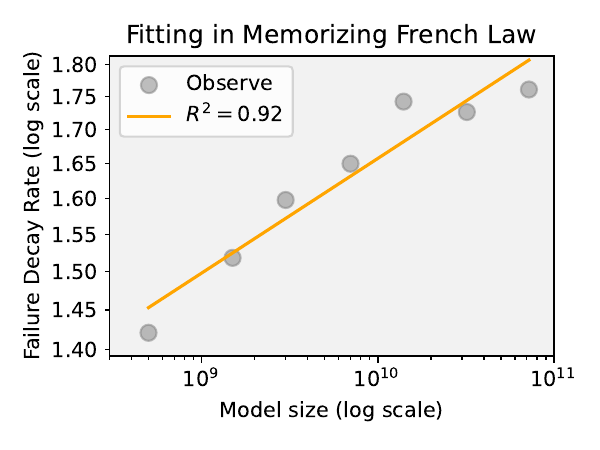}}
		\subfigure{\includegraphics[width=0.23\textwidth]{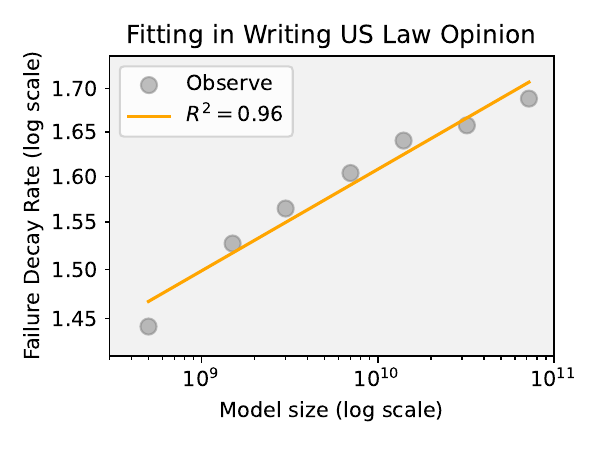}}
        \subfigure{\includegraphics[width=0.23\textwidth]{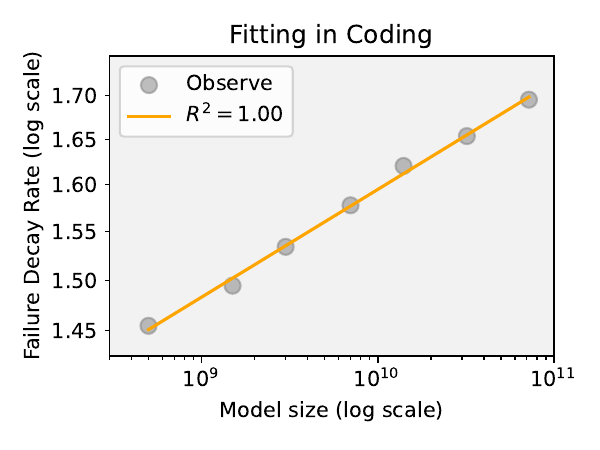}}
        
        \subfigure{\includegraphics[width=0.23\textwidth]{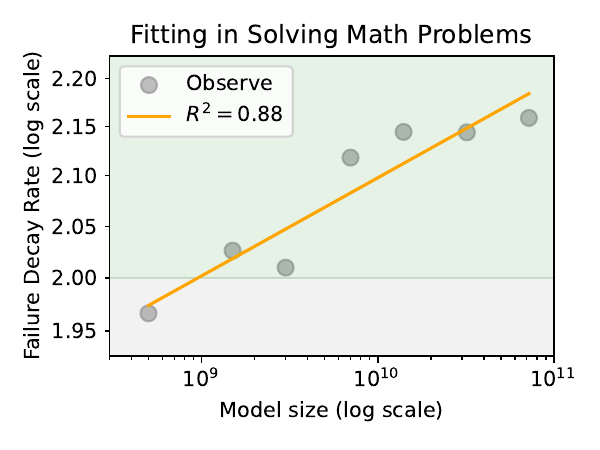}}
        \subfigure{\includegraphics[width=0.23\textwidth]{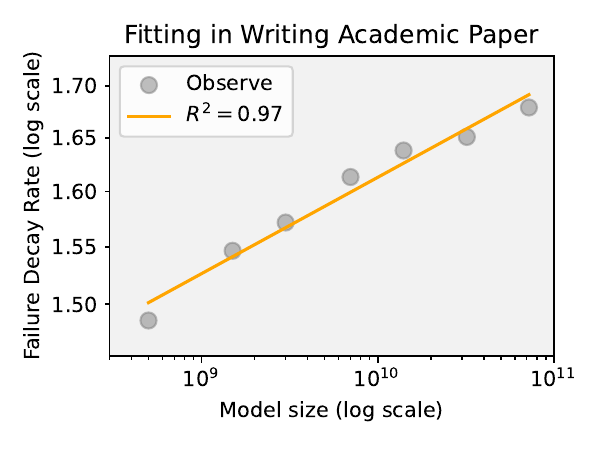}}
        \subfigure{\includegraphics[width=0.23\textwidth]{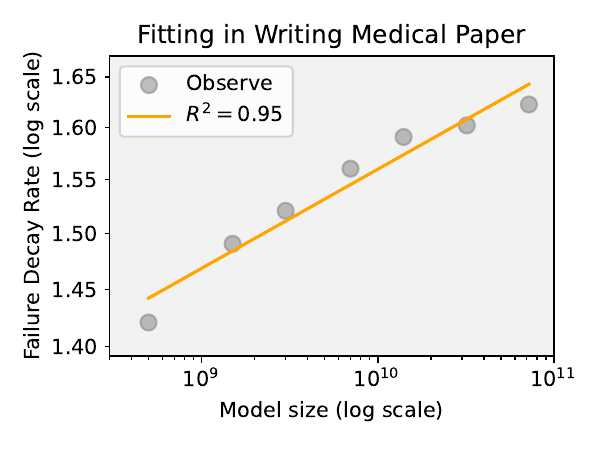}}
        \subfigure{\includegraphics[width=0.23\textwidth]{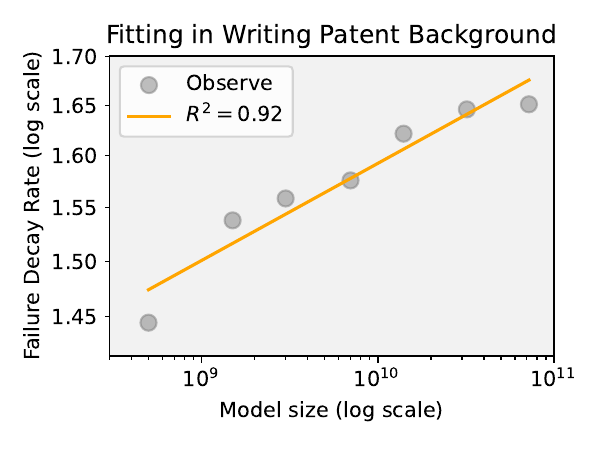}}
        \caption{Impact of Model Size on Failure Decay Rate in Language Tasks. The x-axis represents model size and the y-axis represents failure decay rate, both on a log scale. The results show that the relationship between model size and failure decay rate approximates a straight line on a log-log plot.}
        \label{fig:fit_model_size}
    \end{figure*}

Figure~\ref{fig:fit_model_size} illustrates the impact of model size on failure decay rate. The x-axis represents the model size, and the y-axis represents the fitted decay rate, both on a logarithmic scale. The circle markers represent the performance of the Qwen models. In general, these points approximately follow a straight line. Among these tasks, the points are the closest to a straight line in Knowledge QA (NQ and Trivia QA) and Coding tasks. 
In other tasks, we observe that the curve formed by the data points begins to bend downward as the model size becomes larger. This suggests that increasing the model size with the current training techniques will result in a sub-log-linear improvement rate. If this phenomenon holds, we would seriously overestimate the performance when we use a straight line on the log-log plot to predict the decay rate of larger models. Yet, from an optimistic perspective, we can attribute this slowing growth trend to the limitations of current training methods. Specifically, it is enough to use current training techniques and data size for training small models. As a result, the improvements when model sizes are small fall along the straight line on the log-log plot. However, current data and techniques are limited when the model size reaches 32B or 72B. Consequently, the improvements fall short of expectations. Optimistically, if future researchers make breakthroughs in training techniques, the performance of these large models could still return to the straight line on the log-log plot.

Besides the results of QWen 2.5, we also show the performance across different model architectures in Figure~\ref{fig:full_perf}. The models include GPT2~\citep{radford2019language}, OPT~\citep{zhang2022opt}, GPT-Neo~\citep{gpt-neo}, Llama-1~\citep{touvron2023llama1}, Llama-2~\citep{touvron2023llama2}, Llama-3~\citep{dubey2024llama}, Phi-2~\citep{javaheripi2023phi}, GLM4~\citep{glm2024chatglm}, DeepSeek-V2~\citep{deepseekv2}, BaichuanM1~\citep{baichuan_m1}, Mistral~\citep{jiang2023mistral7b}, QWen2.5~\citep{qwen2.5}.
We can see that across different model architectures, larger models tend to perform better, forming an approximately linear trend. This is evident in models such as the OPT series, Llama 3 \& 3.2 series. This observation aligns with our idea of fitting a straight line. Additionally, we observe that more recently released models tend to achieve better performance over time. Therefore, optimistically speaking, although the fitted line slightly overestimates the performance of large models, their performance are expected to improve over time. Moreover, we can see that among these models, Qwen2.5 demonstrates state-of-the-art performance. As a result, we use Qwen2.5 as a benchmark to represent the current cutting-edge level.

In this paper, we adopt this optimistic perspective. We model the effect of model size on failure decay rate as a straight line on the log-log plot and assume that this linear approximation is still valid when researchers train models of larger scales. We proceed with this optimistic fitting approach into the next subsection, where we predict the development of scaling AI in the future.

\subsection{Predicting Future}

\begin{figure*}[t]
    \subcapraggedrighttrue
    \subcaphangtrue
        \centering
        \subfigure{\includegraphics[width=0.23\textwidth]{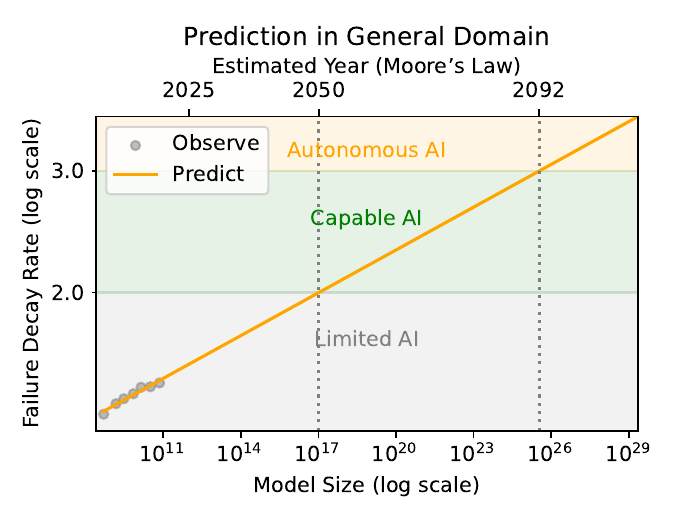}}
        \subfigure{\includegraphics[width=0.23\textwidth]{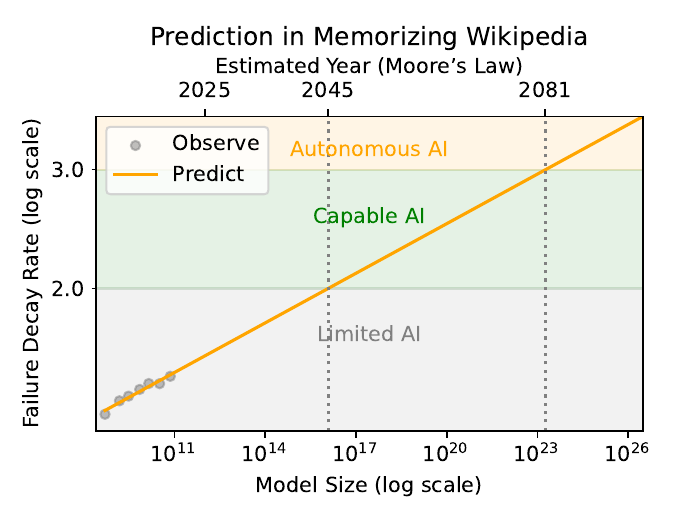}}
        \subfigure{\includegraphics[width=0.23\textwidth]{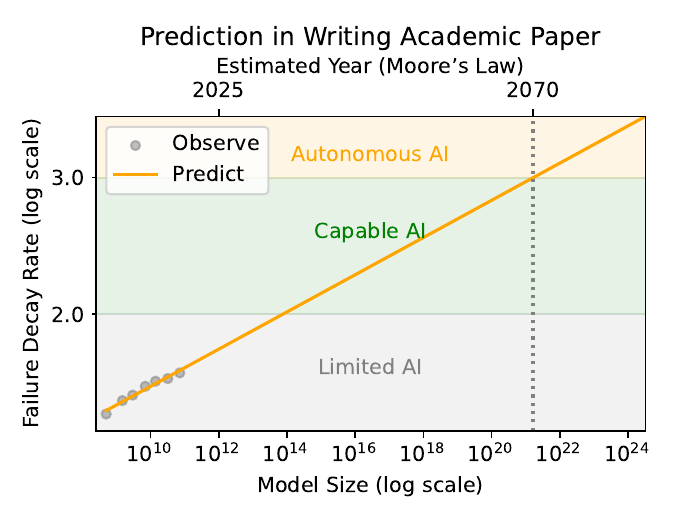}}
        \subfigure{\includegraphics[width=0.23\textwidth]{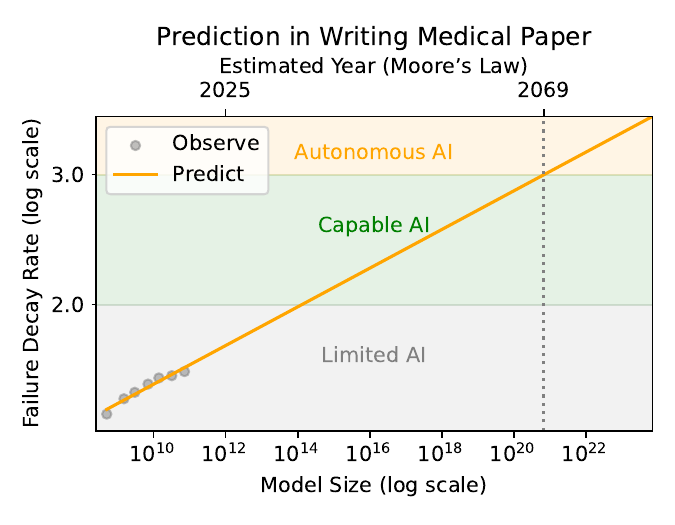}}
        
        \subfigure{\includegraphics[width=0.23\textwidth]{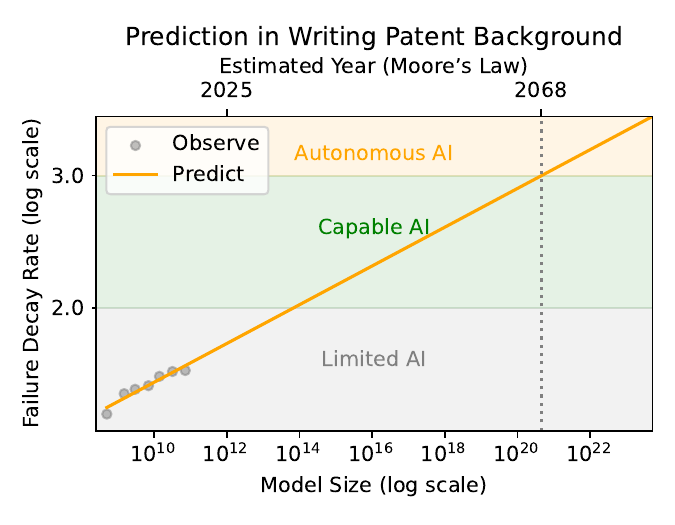}}
        \subfigure{\includegraphics[width=0.23\textwidth]{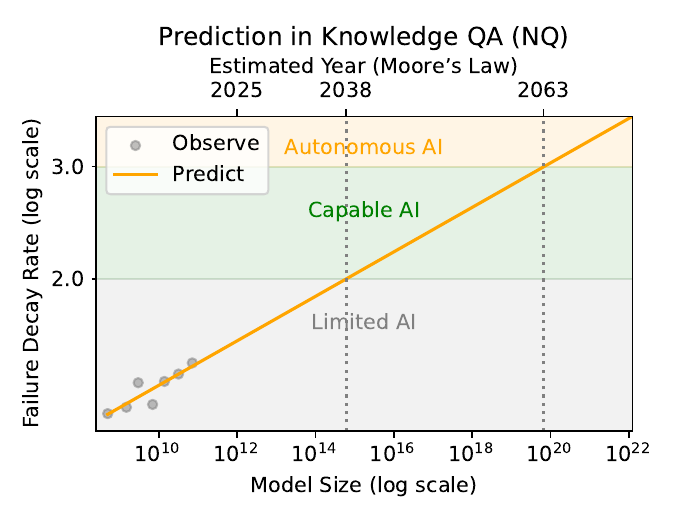}}
        \subfigure{\includegraphics[width=0.23\textwidth]{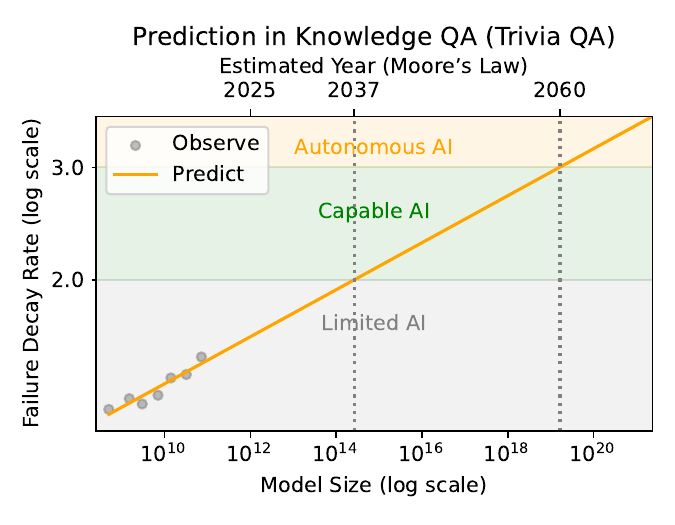}}
        \subfigure{\includegraphics[width=0.23\textwidth]{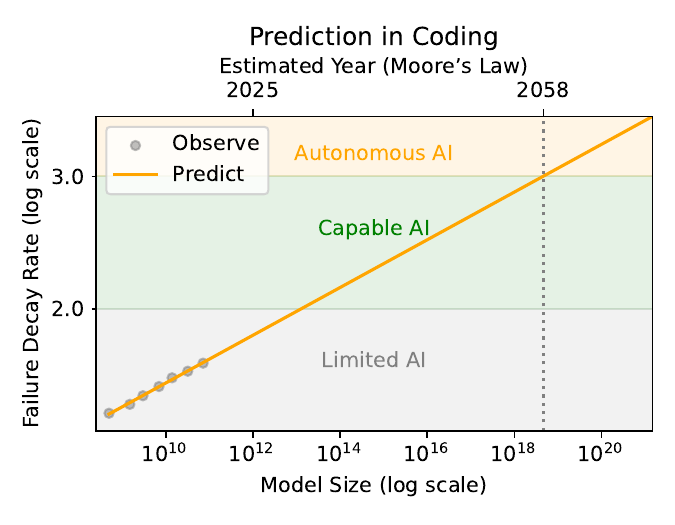}}
        
        \subfigure{\includegraphics[width=0.23\textwidth]{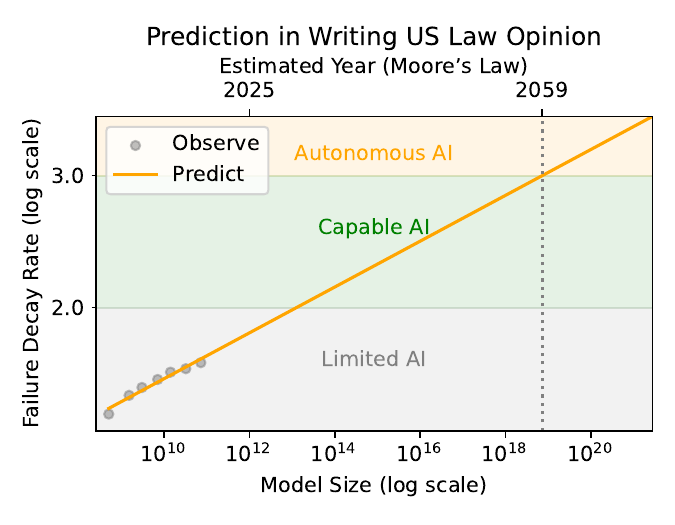}}
        \subfigure{\includegraphics[width=0.23\textwidth]{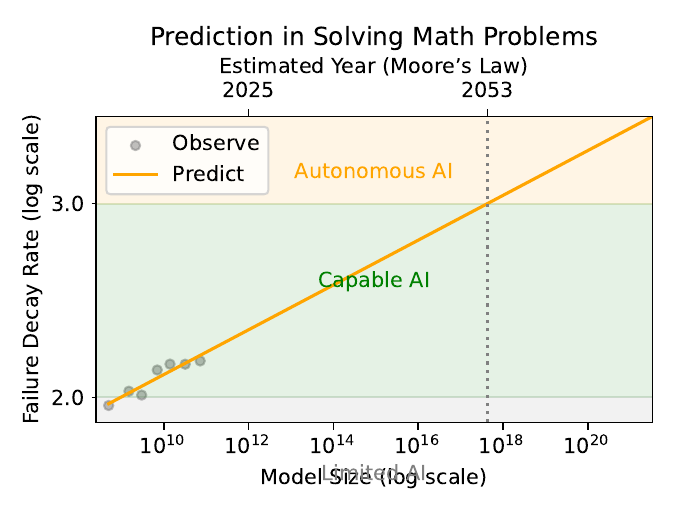}}
        \subfigure{\includegraphics[width=0.23\textwidth]{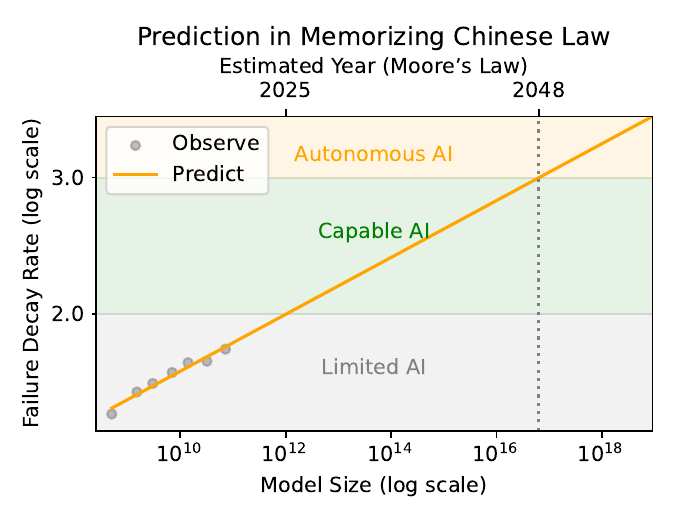}}
        \subfigure{\includegraphics[width=0.23\textwidth]{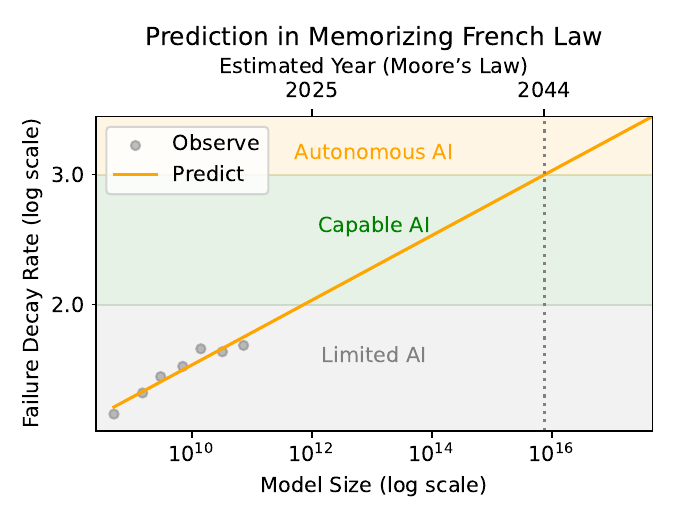}}
        \caption{Prediction for Higher Intelligence Levels in Language Tasks. The x-axis represents model size, and the top axis shows the estimated time based on Moore's Law. The y-axis represents the failure decay rate. Different colors indicate different intelligence levels. The results suggest that it will take several decades to achieve Autonomous Level in language tasks.}
        \label{fig:future_pred}
    \end{figure*}

We optimistically assume that, as the model size increases, the failure decay rate improves along a straight line on a log-log plot. We fit the straight line based on current models and extrapolate the straight line to predict models with larger scales in the future. Figure~\ref{fig:future_pred} shows the results of this extrapolation. The x-axis represents the model size, and the y-axis represents the failure decay rate, both on a logarithmic scale. We calculate the model size required to enable the decay rate to $3$, which corresponds to the Autonomous Level.
The results indicate that for structured tasks, such as mathematics, law, and coding, the required parameter size is around $10^{18}$. These tasks are governed by clear rules and formats, which facilitate learning for AI systems and require fewer parameters. On the other hand, more complex tasks, such as knowledge-based question answering, patent applications, and writing medical or academic papers, require a parameter size around $10^{21}$. These tasks are more intricate, demand specialized knowledge, and require sophisticated reasoning abilities. Thus, AI systems need to be very large to grasp them.
Finally, we adopt a general task that requires the model to comprehend all the information on the Internet. Specifically, we use the C4 English dataset~\citep{raffel2020exploring}, which contains high-quality English corpus from the Internet. We test whether models can memorize the information by asking models to predict the next word given all previous words. According to the prediction results, the required parameter size for Autonomous Level is astonishing, around $10^{26}$.

Achieving such a scale with current hardware is virtually impossible. In the case of general language tasks, the required scale is on the order of $10^{26}$. This number is even $5$ orders of magnitude higher than the total number of neurons in all of humanity's brains combined. Specifically, the number of neurons in a single human brain is around $10^{11}$, and considering the global population is approximately $10^{10}$, the total number of neurons in all human brains is about $10^{21}$. This is only $10^{-5}$ of the scale needed for the AI model. From this perspective, building such a large AI model would be like creating a machine with computational complexity far greater than the total computational capacity of the human species. 
If we ignore any computational costs, such as training and inference, and just focus on loading this massive model onto H100 GPUs, here's the calculation: Since the memory of an H100 GPU is 80GB, we would need $5 \times 10^{15}$ GPUs. Based on the cost of H100 GPU (\$30,000) and the market value of Apple Inc. (\$3.7 trillion) in February 2025, the total value of these GPUs would be equivalent to $4 \times 10^7$ times the market value of Apple Inc. As we can see, without breakthroughs in hardware and AI technology, it is infeasible to afford scaling for Autonomous-Level intelligence.

If hardwares like GPU and CPU continue to improve with the rate suggested by Moore's Law, we can predict the time when sufficiently large models can be developed. Moore's Law states that the performance of chips doubles approximately every 18 months. As chip performance doubles, we can also double the size of AI systems without too much cost.
Assuming Moore's Law continues to hold, and taking the current maximum trainable model size as 1 trillion parameters, we can forecast the maximum trainable model size at each time in the future. In Figure~\ref{fig:future_pred}, the top x-axis shows the predicted timeline. The results suggest that for structured tasks such as mathematics, law, and coding, it will take approximately 30 more years before sufficiently large models can be trained to achieve Autonomous Level. For more complex tasks, such as question answering, patent applications, and writing medical or academic papers, we project that it will take about 40 years. Finally, for general tasks, which require models to handle the full breadth of knowledge across various domains, we anticipate that it will take 70 years to train sufficiently large models.
This projection provides a timeline for the future of AI development. Nevertheless, a model with larger scale require more training data and corresopnding training techniques. Even with such hardware improvement, it is still important to achieve breakthrough in stably training large models with only limited data.

\section{Theoretical Analysis of Survival Game}
\label{sec:theory}

In previous sections, we present the experimental results of current AI systems. Results demonstrate that many AI systems are stuck at the Limited Level on complex human tasks and that achieving the Autonomous Level requires extremely high parameter size. This raises an intriguing question: \textit{Why is the Autonomous Level so difficult to achieve for current AI systems?}

To address this, we turn to the framework of Self-Organized Criticality (SOC)~\citep{PhysRevLett.59.381}, a complexity theory in Physics. SOC describes systems in a criticality state where small perturbations can trigger large-scale changes. We propose that \textit{Survival Game} is deeply related to SOC and that drawing this connection provides insights into the nature of human tasks and AI. In the following, we will first discuss how \textit{Survival Game} relates to SOC, then model \textit{Survival Game} with a SOC framework.

\subsection{Human Tasks exhibit Criticality}

SOC refers to a phenomenon within complex systems where the system tends to settle into a stable state and yet is inherently sensitive to perturbations. On one hand, the system exhibits a self-organizing property: when a disturbance occurs, the system adapts and re-establishes a new equilibrium. On the other hand, the system is always in a criticality state where even the slightest perturbation in any single part can trigger cascading changes throughout the entire system. In this way, the system remains in a delicate balance: it is in a stable state but a slight perturbation can result in a massive reorganization throughout the entire system.
This property has been observed in many natural phenomena, such as earthquakes~\citep{turcotte1985collapse}, forest fires~\citep{malamud1998forest}, proteins~\citep{phillips2014fractals}, and neuronal avalanches~\citep{chialvo2010emergent}. 

We propose that human tasks exhibit criticality, which leads to \textit{Survival Game} as a SOC system.  In \textit{Survival Game}, the questions posed and the corresponding correct answers form a system. The self-organizing nature of this system is evident in the way that different questions correspond to different correct answers. We can imagine the correspondence between a question and its answer as a dynamic process. When a question and the correct answer are given, the system is in a stable state. When certain parts of the question are modified, the answer should undergo a self-organizing modification process and eventually evolve into a new correct answer, thus bringing the entire system back to stability. The system’s criticality arises from the nature of human tasks. For example, a minor alteration in a mathematical question can drastically change the approach required to solve it. This criticality property requires the test subject to address subtle differences in the question that can lead to significantly distinct answers. Merely memorizing answers for several specific cases does not help because a small change results in entirely different answers. We believe this is the cause of why so many AI systems are at Limited Level. AI systems might simply memorize some answers and yet human tasks are not friendly to memorization because of criticality property.

In the following, we will show several examples to help illustrate how human tasks exhibit criticality.

\textbf{Physics}:
Physics exhibits criticality. In physics, even seemingly similar problems can lead to vastly different solutions depending on the initial conditions. For instance, consider a question about a matter's state or superconductivity. The answer relies crucially on whether the temperature is above or below a threshold. Similarly, if we ask about physical laws, the appropriate theory, quantum mechanics or classical physics, depends on the scale. This phenomenon is ubiquitous in physics: small variations in initial conditions can lead to fundamentally different results.
By incorporating such questions into an \textit{Survival Game}, we naturally create a system with SOC property. The test subjects must adapt to subtle changes in initial conditions. Otherwise, their responses would be entirely wrong. If a participant relies solely on memorizing answers for some specific conditions, they will struggle to apply their knowledge to new, subtly altered questions. In such cases, the failure attempts are likely to approach infinity, as the answers will deviate drastically from the memorized solutions.

\textbf{Mathematics}:
Mathematics exhibits criticality. Mathematics embodies the very essence of SOC. Take Fermat's Last Theorem as an example. It asserts that there are no integer solutions to the equation $x^n + y^n = z^n$ for any integer $n>2$. At first glance, the change of $n$ might seem like a minor adjustment and does not affect the essence of the problem. However, the theorem's sensitivity to the value of n is profound. As soon as $n$ grows larger, the problem becomes much more complex. The theorem with small $n$ values was proven soon, but it took more than two centuries before the general case was finally proven. Many math problems show such criticality property that seemingly small modifications can alter the entire landscape of solutions. Thus, when mathematical problems are used in \textit{Survival Game}, subjects must be acutely aware of the details and select the appropriate mathematical tools. Subjects that do not possess this ability will take infinite attempts to arrive at the correct answer.

\textbf{Law}:
Legal issues exhibit criticality. In law, seemingly minor differences in behavior can lead to vastly divergent legal consequences. For instance, a suspect’s actions might determine whether the charge is premeditated murder or voluntary manslaughter, whether it was excessive self-defense or justifiable defense, or even whether they are guilty or innocent. Such significant shifts in legal outcomes can arise from differences in behavior that, on the surface, might appear negligible. When legal questions are incorporated into an \textit{Survival Game}, they require the subjects to discern these nuances and reason the outcomes. If a subject has only memorized conclusions for specific cases, they will struggle in new scenarios and will inevitably make infinite attempts before arriving at the right answers.

\textbf{Medicine}:
Medicine exhibits criticality. In medicine, small, seemingly insignificant changes in physiological parameters can result in dramatic shifts in treatment plans. For example, in cancer treatment, subtle genetic differences between patients can lead to vastly different responses to immunotherapy~\citep{hwang2020immune, martinez2023genetic}. In diabetes management, minor fluctuations in blood sugar levels may cause significant organ damage, necessitating precise adjustments in treatment~\citep{zhang2019molecular}. 
Although the differences in physiological parameters may appear minimal, the underlying medical phenomena and the corresponding treatments can vary greatly. Therefore, answering medical questions requires grasping the underlying principles. If the subjects simply rely on memorizing specific examples, they will make catastrophic errors in new cases.

These human tasks exhibit criticality and make \textit{Survival Game} exhibit SOC. Criticality is a typical symbol of complexity: Since a slight change in the question can result in an entirely new answer, successfully operating the task requires a full understanding of the underlying mechanism. Otherwise, subjects will be completely wrong about the correct answer. It will be difficult for them to arrive at the right one through trial and error. This is indeed the case for current AI systems as shown in our experiments. They remain at the Limited Level where the number of trials is infinite. This suggests that most AI systems do not fully understand the mechanism. 
Even if they are trained on enormous data and have memorized all of it, it is not enough for tasks with criticality properties.

\subsection{Modeling Survival Game as a Self-Organized Criticality System}

We have qualitatively explored how \textit{Survival Game} exhibits SOC property on human tasks. Now, we will adopt a quantitative perspective to validate this hypothesis. We will see that it closely resembles a typical SOC model and exhibits very similar experimental phenomena.

\begin{figure}[t]
  \centering
  \includegraphics[width=0.3\textwidth]{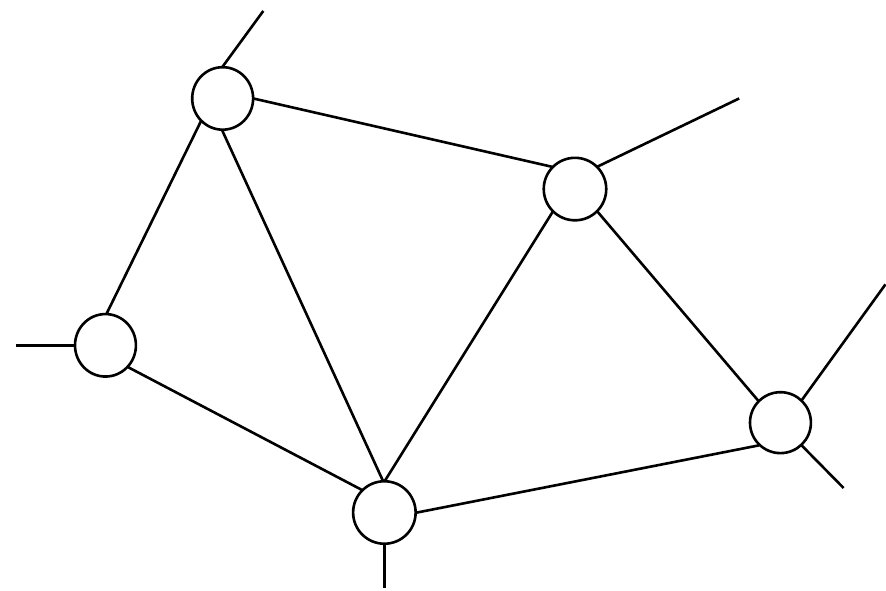}
  \caption{Modeling \textit{Survival Game} as a complex network. Nodes represent concepts and edges represent interdependence. The network is also a sandpile model~\citep{PhysRevLett.59.381}.}
  \label{fig:soc_nodes}
\end{figure}

In \textit{Survival Game}, both questions and answers inherently consist of many basic concepts. Since answers change when questions change, these concepts are intricately coupled. Because of the criticality property of human tasks, even a small change in one concept can trigger significant changes in many other concepts. We can imagine this interconnection as a network graph, as shown in Figure~\ref{fig:soc_nodes}. Each node represents a concept, and the edges between the nodes represent the couplings between these concepts. Slightly changing the question is akin to disturbing a node. When a node becomes unstable, it is activated and is likely to make its neighbors also unstable. This might eventually result in cascading activation throughout the network. Such changes throughout the entire network imitate how seemingly small changes in the question can lead to very different answers in \textit{Survival Game}.

This conceptual network is very similar to a typical SOC system, namely the sandpile model~\citep{PhysRevLett.59.381}. It has been used to model complex systems in the natural world. In the sandpile model, each node accumulates sand. To disturb the network is to add a little more sand on a node. When the amount of sand exceeds a certain threshold, the sand topples from this node. The toppled sand is distributed to its neighbors, and thus the neighbors' sand may also topple. This cascade of toppling sometimes results in an avalanche throughout the entire network. 
We believe that this sand avalanche effect is close to the cascading activation of concepts in our conceptual network for \textit{Survival Game}. Therefore, we further implement our conceptual network as a sandpile model. Specifically, a slight change in a question is equivalent to adding a small amount of sand to a node. The degree to which the answer changes after altering the question is analogous to the avalanche size triggered by adding sand. Through this correspondence, we model \textit{Survival Game} as a sandpile model. The validity of this modeling can be confirmed by testing whether the sandpile model accurately reflects the characteristics of \textit{Survival Game} as observed in real-world applications.

\begin{figure*}[t]
    \subcapraggedrighttrue
    \subcaphangtrue
        \centering
        \subfigure{\includegraphics[width=0.3\textwidth]{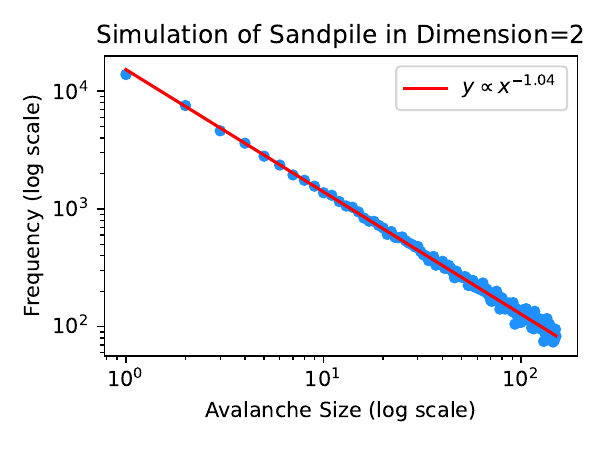}}
        \hfill
        \subfigure{\includegraphics[width=0.3\textwidth]{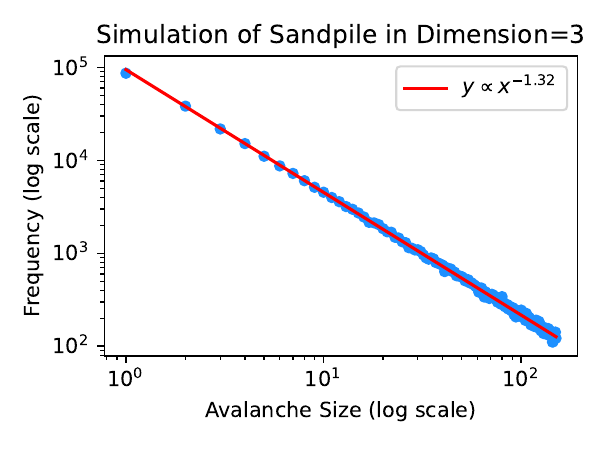}}
        \hfill
        \subfigure{\includegraphics[width=0.3\textwidth]{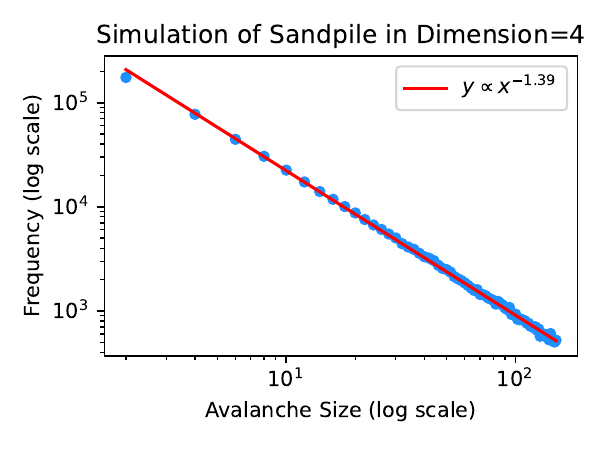}}
        \caption{Simulation Results of the sandpile model with a grid topology in 2, 3, and 4 dimensions. We examine the distribution of sand avalanche size after a disturbance. The x-axis represents the avalanche size, and the y-axis shows the frequency. The distribution of avalanche size approximates a power law.}
        \label{fig:simulate_sandpile}
    \end{figure*}

Modeling \textit{Survival Game} as a sandpile model can explain why the failure count distribution of current AI systems resembles a power law.
In real-world experiments, we empirically show that failure count distribution is very close to a power law, as described in Section~\ref{sec:fit_failure_count_distribution}. However, the underlying reason is a mystery.
Now, we can investigate this phenomenon based on our abstraction of \textit{Survival Game}, namely the sandpile model.
Figure~\ref{fig:simulate_sandpile} shows the simulated distribution of avalanche size in the sandpile model. Based on our analogy, this also corresponds to the distribution of how much answers change after altering the questions in \textit{Survival Game}. The distribution follows a power law, which closely matches the failure count distribution of AI systems.
This actually reflects that current AI systems rely on memorization and exploration to solve new problems. Here is the reason. If AI relies on memorization and exploration, it means that when faced with a new question in \textit{Survival Game}, it recalls similar questions it has memorized and, based on the remembered answers, explores potential solutions. As a result, the number of explorations correlates with the distance between the answers. In contrast, if the AI system genuinely understood the test questions, the distance between answers would not correlate with the number of failure attempts. In fact, the mechanics of the sandpile model are clear, and an AI system that truly grasps these mechanics could directly compute the stable state without any failure attempt. Therefore, the phenomena indicate that AI systems do not fully understand the mechanics of the task and instead rely on memorization and exploration to find answers.

Modeling \textit{Survival Game} as a sandpile model also helps to illuminate how scaling works. From our experiments in Section~\ref{sec:fit_model_scale_to_decay}, we empirically observe that scaling model size can improve the decay rate and thereby improve intelligence. Yet, the underlying reason is a mystery.
However, we find that the dimensionality of the sandpile topology has a similar effect and can explain this phenomenon. 
More precisely, both theoretical analyses~\citep{dhar2006theoretical, zachariou2015generalised} and our simulations in Figure~\ref{fig:simulate_sandpile} show that the power law exponent of the sandpile model increases as the topology becomes high-dimensional. This is because a higher-dimensional topology creates more connections between nodes, enabling the network to stabilize more quickly when disturbed. We can draw the analogy between the sandpile's stabilization through these connections and intelligence systems' solving new problems through exploration. When we scale an intelligent system, we expand the problem-solving space to a higher-dimensional level and construct new paths between concept nodes. These new paths enable current intelligence systems to explore new solutions from memorized ones more quickly. This is akin to how a sandpile model stabilizes itself through new connections in a higher dimensionality. If we regard these new paths for exploration as the new connections in SOC systems, the effects of scaling can be explained: scaling makes exploration more effective.

In summary, \textit{Survival Game} can be modeled as a sandpile model. It helps us gain a deep understanding of the nature of human tasks and current AI. The sandpile model is a complex system where a little more sand can result in an avalanche throughout the entire landscape. Similarly, \textit{Survival Game} with human tasks are also complex where a small change in environments requires completely different responses. 
No matter whether it is to predict the state of the sandpile model or find the correct solution in \textit{Survival Game}, it is necessary to fully understand the underlying mechanisms. 
Yet, most AI systems do not understand these mechanisms and rely on superficial imitation, such as memorization and exploration. 
They exhibit a typical power law with a small exponent. 
Scaling is able to transform the exploration space to a higher dimensionality and makes the exploration process more effective. 
Nevertheless, if AI lacks a full understanding of underlying mechanisms, SOC property will make it extremely difficult to achieve the Autonomous Level of intelligence.

\section{Conclusions and Future Work}
\label{sec:conclusion}

In this paper, we introduce \textit{Survival Game}. It is inspired by Natural Selection and quantifies the intelligence of any subject in any task. It demonstrates three advantages.
\begin{itemize}
	\item Firstly, the test offers a clear physical meaning and a well-defined mathematical framework. It categorizes intelligence into three levels and enables a deep understanding of the subject's intelligence. 
	\item Secondly, \textit{Survival Game} provides a roadmap for the future development of AI. It shows a clear relationship to the scale of AI systems, which enables us to project the time to reach high-level intelligence.
	\item Finally, \textit{Survival Game} helps reveal the nature of human tasks and AI. It suggests that human tasks exhibit criticality properties and AI's superficial manner to solve human tasks. 
\end{itemize}

We believe that future efforts should focus on three key areas.
Firstly, since current AI technologies are mostly at the Limited Level, we need to identify appropriate application scenarios and design an effective human supervision framework.
Secondly, since \textit{Survival Game} connects the model size to the intelligence levels, we can use \textit{Survival Game} to plan the roadmap of future AI development. 
Finally, linking \textit{Survival Game} to SOC theory shows promising results and more efforts shall be made in this direction. This will not only help design more effective AI models but also deepen our understanding of humans ourselves.

\addcontentsline{toc}{section}{Reproducibility}
\section*{Reproducibility}
Code is open-sourced at \url{https://github.com/jingtaozhan/IntelligenceTest}.

\addcontentsline{toc}{section}{Contributions}
\section*{Contributions}
  
Jingtao Zhan initiated the project, proposed \textit{Survival Game}, conducted experiments, and used SOC for theoretical analysis.
Jiahao Zhao contributed to the experiments by conducting the text search task. 
Jiayu Li contributed to the experiments by conducting the recommendation system task.
Yiqun Liu supervised the project, engaged in multiple discussions, and guided the researchers in exploring the fundamental principles underlying the observed phenomena. 
Bo Zhang participated in multiple discussions, provided important advice, and pointed out the similarity between the experimental results and phenomenon in SOC systems.
Qingyao Ai, Jiaxin Mao, Hongning Wang, Min Zhang, and Shaoping Ma were involved in the early-stage discussions and provided invaluable feedback.
Jingtao Zhan drafted the manuscript, and all other authors contributed important insights and suggestions to improve the writing.

\bibliographystyle{my}
\bibliography{custom}

\begin{thebibliography}{96}
\providecommand{\natexlab}[1]{#1}
\providecommand{\url}[1]{\texttt{#1}}
\expandafter\ifx\csname urlstyle\endcsname\relax
  \providecommand{\doi}[1]{doi: #1}\else
  \providecommand{\doi}{doi: \begingroup \urlstyle{rm}\Url}\fi

\bibitem[Aharoni et~al.(2024)Aharoni, Fernandes, Brady, Alexander, Criner,
  Queen, Rando, Nahmias, and Crespo]{aharoni2024attributions}
Eyal Aharoni, Sharlene Fernandes, Daniel~J Brady, Caelan Alexander, Michael
  Criner, Kara Queen, Javier Rando, Eddy Nahmias, and Victor Crespo.
\newblock Attributions toward artificial agents in a modified moral turing
  test.
\newblock \emph{Scientific reports}, 14\penalty0 (1):\penalty0 8458, 2024.

\bibitem[Austin et~al.(2021)Austin, Odena, Nye, Bosma, Michalewski, Dohan,
  Jiang, Cai, Terry, Le, et~al.]{austin2021program}
Jacob Austin, Augustus Odena, Maxwell Nye, Maarten Bosma, Henryk Michalewski,
  David Dohan, Ellen Jiang, Carrie Cai, Michael Terry, Quoc Le, et~al.
\newblock Program synthesis with large language models.
\newblock \emph{arXiv preprint arXiv:2108.07732}, 2021.

\bibitem[Baichuan(2024)]{baichuan_m1}
Baichuan.
\newblock Baichuan-m1-14b.
\newblock \url{https://github.com/baichuan-inc/Baichuan-M1-14B}, 2024.

\bibitem[Baird et~al.(2003)Baird, Coates, and Fateman]{baird2003pessimalprint}
Henry~S Baird, Allison~L Coates, and Richard~J Fateman.
\newblock Pessimalprint: a reverse turing test.
\newblock \emph{International Journal on Document Analysis and Recognition},
  5:\penalty0 158--163, 2003.

\bibitem[Bajaj et~al.(2016)Bajaj, Campos, Craswell, Deng, Gao, Liu, Majumder,
  McNamara, Mitra, Nguyen, et~al.]{Bajaj2016Msmarco}
Payal Bajaj, Daniel Campos, Nick Craswell, Li~Deng, Jianfeng Gao, Xiaodong Liu,
  Rangan Majumder, Andrew McNamara, Bhaskar Mitra, Tri Nguyen, et~al.
\newblock Ms marco: A human generated machine reading comprehension dataset.
\newblock \emph{arXiv preprint arXiv:1611.09268}, 2016.

\bibitem[Bak(2013)]{bak2013nature}
Per Bak.
\newblock \emph{How nature works: the science of self-organized criticality}.
\newblock Springer Science \& Business Media, 2013.

\bibitem[Bak et~al.(1987)Bak, Tang, and Wiesenfeld]{PhysRevLett.59.381}
Per Bak, Chao Tang, and Kurt Wiesenfeld.
\newblock Self-organized criticality: An explanation of the 1/f noise.
\newblock \emph{Phys. Rev. Lett.}, 59:\penalty0 381--384, Jul 1987.
\newblock \doi{10.1103/PhysRevLett.59.381}.
\newblock URL \url{https://link.aps.org/doi/10.1103/PhysRevLett.59.381}.

\bibitem[Biever(2023)]{biever2023chatgpt}
Celeste Biever.
\newblock Chatgpt broke the turing test-the race is on for new ways to assess
  ai.
\newblock \emph{Nature}, 619\penalty0 (7971):\penalty0 686--689, 2023.

\bibitem[Black et~al.(2021)Black, Leo, Wang, Leahy, and Biderman]{gpt-neo}
Sid Black, Gao Leo, Phil Wang, Connor Leahy, and Stella Biderman.
\newblock {GPT-Neo: Large Scale Autoregressive Language Modeling with
  Mesh-Tensorflow}, March 2021.
\newblock URL \url{https://doi.org/10.5281/zenodo.5297715}.
\newblock {If you use this software, please cite it using these metadata.}

\bibitem[Bringsjord et~al.(2003)Bringsjord, Bello, and
  Ferrucci]{bringsjord2003creativity}
Selmer Bringsjord, Paul Bello, and David Ferrucci.
\newblock Creativity, the turing test, and the (better) lovelace test.
\newblock \emph{The Turing test: the elusive standard of artificial
  intelligence}, pp.\  215--239, 2003.

\bibitem[Brown et~al.(2020)Brown, Mann, Ryder, Subbiah, Kaplan, Dhariwal,
  Neelakantan, Shyam, Sastry, Askell, et~al.]{brown2020language}
Tom Brown, Benjamin Mann, Nick Ryder, Melanie Subbiah, Jared~D Kaplan, Prafulla
  Dhariwal, Arvind Neelakantan, Pranav Shyam, Girish Sastry, Amanda Askell,
  et~al.
\newblock Language models are few-shot learners.
\newblock \emph{Advances in neural information processing systems},
  33:\penalty0 1877--1901, 2020.

\bibitem[Cen et~al.(2020)Cen, Zhang, Zou, Zhou, Yang, and
  Tang]{cen2020controllable}
Yukuo Cen, Jianwei Zhang, Xu~Zou, Chang Zhou, Hongxia Yang, and Jie Tang.
\newblock Controllable multi-interest framework for recommendation.
\newblock In \emph{Proceedings of the 26th ACM SIGKDD International Conference
  on Knowledge Discovery and Data Mining}, pp.\  2942–2951. ACM, 2020.

\bibitem[Chen et~al.(2021{\natexlab{a}})Chen, Tworek, Jun, Yuan,
  de~Oliveira~Pinto, Kaplan, Edwards, Burda, Joseph, Brockman, Ray, Puri,
  Krueger, Petrov, Khlaaf, Sastry, Mishkin, Chan, Gray, Ryder, Pavlov, Power,
  Kaiser, Bavarian, Winter, Tillet, Such, Cummings, Plappert, Chantzis, Barnes,
  Herbert-Voss, Guss, Nichol, Paino, Tezak, Tang, Babuschkin, Balaji, Jain,
  Saunders, Hesse, Carr, Leike, Achiam, Misra, Morikawa, Radford, Knight,
  Brundage, Murati, Mayer, Welinder, McGrew, Amodei, McCandlish, Sutskever, and
  Zaremba]{chen2021codex}
Mark Chen, Jerry Tworek, Heewoo Jun, Qiming Yuan, Henrique~Ponde
  de~Oliveira~Pinto, Jared Kaplan, Harri Edwards, Yuri Burda, Nicholas Joseph,
  Greg Brockman, Alex Ray, Raul Puri, Gretchen Krueger, Michael Petrov, Heidy
  Khlaaf, Girish Sastry, Pamela Mishkin, Brooke Chan, Scott Gray, Nick Ryder,
  Mikhail Pavlov, Alethea Power, Lukasz Kaiser, Mohammad Bavarian, Clemens
  Winter, Philippe Tillet, Felipe~Petroski Such, Dave Cummings, Matthias
  Plappert, Fotios Chantzis, Elizabeth Barnes, Ariel Herbert-Voss,
  William~Hebgen Guss, Alex Nichol, Alex Paino, Nikolas Tezak, Jie Tang, Igor
  Babuschkin, Suchir Balaji, Shantanu Jain, William Saunders, Christopher
  Hesse, Andrew~N. Carr, Jan Leike, Josh Achiam, Vedant Misra, Evan Morikawa,
  Alec Radford, Matthew Knight, Miles Brundage, Mira Murati, Katie Mayer, Peter
  Welinder, Bob McGrew, Dario Amodei, Sam McCandlish, Ilya Sutskever, and
  Wojciech Zaremba.
\newblock Evaluating large language models trained on code.
\newblock 2021{\natexlab{a}}.

\bibitem[Chen et~al.(2020)Chen, Kornblith, Norouzi, and Hinton]{chen2020simple}
Ting Chen, Simon Kornblith, Mohammad Norouzi, and Geoffrey Hinton.
\newblock A simple framework for contrastive learning of visual
  representations.
\newblock In \emph{International conference on machine learning}, pp.\
  1597--1607. PmLR, 2020.

\bibitem[Chen et~al.(2021{\natexlab{b}})Chen, Wang, Fan, Huang, Yang, and
  Zhu]{chen2021curriculum}
Yudong Chen, Xin Wang, Miao Fan, Jizhou Huang, Shengwen Yang, and Wenwu Zhu.
\newblock Curriculum meta-learning for next poi recommendation.
\newblock In \emph{Proceedings of the 27th ACM SIGKDD Conference on Knowledge
  Discovery \& Data Mining}, pp.\  2692--2702, 2021{\natexlab{b}}.

\bibitem[Chialvo(2010)]{chialvo2010emergent}
Dante~R Chialvo.
\newblock Emergent complex neural dynamics.
\newblock \emph{Nature physics}, 6\penalty0 (10):\penalty0 744--750, 2010.

\bibitem[Cho et~al.(2011)Cho, Myers, and Leskovec]{cho2011friendship}
Eunjoon Cho, Seth~A Myers, and Jure Leskovec.
\newblock Friendship and mobility: user movement in location-based social
  networks.
\newblock In \emph{Proceedings of the 17th ACM SIGKDD international conference
  on Knowledge discovery and data mining}, pp.\  1082--1090, 2011.

\bibitem[Cobbe et~al.(2021)Cobbe, Kosaraju, Bavarian, Chen, Jun, Kaiser,
  Plappert, Tworek, Hilton, Nakano, Hesse, and Schulman]{cobbe2021gsm8k}
Karl Cobbe, Vineet Kosaraju, Mohammad Bavarian, Mark Chen, Heewoo Jun, Lukasz
  Kaiser, Matthias Plappert, Jerry Tworek, Jacob Hilton, Reiichiro Nakano,
  Christopher Hesse, and John Schulman.
\newblock Training verifiers to solve math word problems.
\newblock \emph{arXiv preprint arXiv:2110.14168}, 2021.

\bibitem[DeepSeek-AI(2024)]{deepseekv2}
DeepSeek-AI.
\newblock Deepseek-v2: A strong, economical, and efficient mixture-of-experts
  language model, 2024.

\bibitem[Deng et~al.(2009)Deng, Dong, Socher, Li, Li, and
  Fei-Fei]{deng2009imagenet}
Jia Deng, Wei Dong, Richard Socher, Li-Jia Li, Kai Li, and Li~Fei-Fei.
\newblock Imagenet: A large-scale hierarchical image database.
\newblock In \emph{2009 IEEE conference on computer vision and pattern
  recognition}, pp.\  248--255. Ieee, 2009.

\bibitem[Deng(2012)]{deng2012mnist}
Li~Deng.
\newblock The mnist database of handwritten digit images for machine learning
  research.
\newblock \emph{IEEE Signal Processing Magazine}, 29\penalty0 (6):\penalty0
  141--142, 2012.

\bibitem[Dhar(2006)]{dhar2006theoretical}
Deepak Dhar.
\newblock Theoretical studies of self-organized criticality.
\newblock \emph{Physica A: Statistical Mechanics and its Applications},
  369\penalty0 (1):\penalty0 29--70, 2006.

\bibitem[Ding et~al.(2020)Ding, Quan, Yao, Li, and Jin]{ding2020simplify}
Jingtao Ding, Yuhan Quan, Quanming Yao, Yong Li, and Depeng Jin.
\newblock Simplify and robustify negative sampling for implicit collaborative
  filtering.
\newblock \emph{Advances in Neural Information Processing Systems},
  33:\penalty0 1094--1105, 2020.

\bibitem[Dubey et~al.(2024)Dubey, Jauhri, Pandey, Kadian, Al-Dahle, Letman,
  Mathur, Schelten, Yang, Fan, et~al.]{dubey2024llama}
Abhimanyu Dubey, Abhinav Jauhri, Abhinav Pandey, Abhishek Kadian, Ahmad
  Al-Dahle, Aiesha Letman, Akhil Mathur, Alan Schelten, Amy Yang, Angela Fan,
  et~al.
\newblock The llama 3 herd of models.
\newblock \emph{arXiv preprint arXiv:2407.21783}, 2024.

\bibitem[Dulac-Arnold et~al.(2021)Dulac-Arnold, Levine, Mankowitz, Li,
  Paduraru, Gowal, and Hester]{dulac2021challenges}
Gabriel Dulac-Arnold, Nir Levine, Daniel~J Mankowitz, Jerry Li, Cosmin
  Paduraru, Sven Gowal, and Todd Hester.
\newblock Challenges of real-world reinforcement learning: definitions,
  benchmarks and analysis.
\newblock \emph{Machine Learning}, 110\penalty0 (9):\penalty0 2419--2468, 2021.

\bibitem[Gao et~al.(2020)Gao, Biderman, Black, Golding, Hoppe, Foster, Phang,
  He, Thite, Nabeshima, et~al.]{gao2020pile}
Leo Gao, Stella Biderman, Sid Black, Laurence Golding, Travis Hoppe, Charles
  Foster, Jason Phang, Horace He, Anish Thite, Noa Nabeshima, et~al.
\newblock The pile: An 800gb dataset of diverse text for language modeling.
\newblock \emph{arXiv preprint arXiv:2101.00027}, 2020.

\bibitem[Geman et~al.(2015)Geman, Geman, Hallonquist, and
  Younes]{geman2015visual}
Donald Geman, Stuart Geman, Neil Hallonquist, and Laurent Younes.
\newblock Visual turing test for computer vision systems.
\newblock \emph{Proceedings of the National Academy of Sciences}, 112\penalty0
  (12):\penalty0 3618--3623, 2015.

\bibitem[GLM et~al.(2024)GLM, Zeng, Xu, Wang, Zhang, Yin, Rojas, Feng, Zhao,
  Lai, Yu, Wang, Sun, Zhang, Cheng, Gui, Tang, Zhang, Li, Zhao, Wu, Zhong, Liu,
  Huang, Zhang, Zheng, Lu, Duan, Zhang, Cao, Yang, Tam, Zhao, Liu, Xia, Zhang,
  Gu, Lv, Liu, Liu, Yang, Song, Zhang, An, Xu, Niu, Yang, Li, Bai, Dong, Qi,
  Wang, Yang, Du, Hou, and Wang]{glm2024chatglm}
Team GLM, Aohan Zeng, Bin Xu, Bowen Wang, Chenhui Zhang, Da~Yin, Diego Rojas,
  Guanyu Feng, Hanlin Zhao, Hanyu Lai, Hao Yu, Hongning Wang, Jiadai Sun,
  Jiajie Zhang, Jiale Cheng, Jiayi Gui, Jie Tang, Jing Zhang, Juanzi Li, Lei
  Zhao, Lindong Wu, Lucen Zhong, Mingdao Liu, Minlie Huang, Peng Zhang, Qinkai
  Zheng, Rui Lu, Shuaiqi Duan, Shudan Zhang, Shulin Cao, Shuxun Yang, Weng~Lam
  Tam, Wenyi Zhao, Xiao Liu, Xiao Xia, Xiaohan Zhang, Xiaotao Gu, Xin Lv,
  Xinghan Liu, Xinyi Liu, Xinyue Yang, Xixuan Song, Xunkai Zhang, Yifan An,
  Yifan Xu, Yilin Niu, Yuantao Yang, Yueyan Li, Yushi Bai, Yuxiao Dong, Zehan
  Qi, Zhaoyu Wang, Zhen Yang, Zhengxiao Du, Zhenyu Hou, and Zihan Wang.
\newblock Chatglm: A family of large language models from glm-130b to glm-4 all
  tools, 2024.

\bibitem[Goertzel(2014)]{goertzel}
Ben Goertzel.
\newblock {Artificial General Intelligence: Concept, State of the Art, and
  Future Prospects}.
\newblock \emph{Journal of Artificial General Intelligence}, 01 2014.
\newblock \doi{10.2478/jagi-2014-0001}.

\bibitem[Gu et~al.(2024)Gu, Rozière, Leather, Solar-Lezama, Synnaeve, and
  Wang]{gu2024cruxeval}
Alex Gu, Baptiste Rozière, Hugh Leather, Armando Solar-Lezama, Gabriel
  Synnaeve, and Sida~I. Wang.
\newblock Cruxeval: A benchmark for code reasoning, understanding and
  execution.
\newblock \emph{arXiv preprint arXiv:2401.03065}, 2024.

\bibitem[Guo et~al.(2025)Guo, Yang, Zhang, Song, Zhang, Xu, Zhu, Ma, Wang, Bi,
  et~al.]{guo2025deepseek}
Daya Guo, Dejian Yang, Haowei Zhang, Junxiao Song, Ruoyu Zhang, Runxin Xu,
  Qihao Zhu, Shirong Ma, Peiyi Wang, Xiao Bi, et~al.
\newblock Deepseek-r1: Incentivizing reasoning capability in llms via
  reinforcement learning.
\newblock \emph{arXiv preprint arXiv:2501.12948}, 2025.

\bibitem[Harnad(1991)]{harnad1991other}
Stevan Harnad.
\newblock Other bodies, other minds: A machine incarnation of an old
  philosophical problem.
\newblock \emph{Minds and Machines}, 1:\penalty0 43--54, 1991.

\bibitem[Harper \& Konstan(2015)Harper and Konstan]{harper2015movielens}
F~Maxwell Harper and Joseph~A Konstan.
\newblock The movielens datasets: History and context.
\newblock \emph{Acm transactions on interactive intelligent systems (tiis)},
  5\penalty0 (4):\penalty0 1--19, 2015.

\bibitem[Haykin(1994)]{haykin1994neural}
Simon Haykin.
\newblock \emph{Neural networks: a comprehensive foundation}.
\newblock Prentice Hall PTR, 1994.

\bibitem[He et~al.(2022)He, Chen, Xie, Li, Doll{\'a}r, and
  Girshick]{he2022masked}
Kaiming He, Xinlei Chen, Saining Xie, Yanghao Li, Piotr Doll{\'a}r, and Ross
  Girshick.
\newblock Masked autoencoders are scalable vision learners.
\newblock In \emph{Proceedings of the IEEE/CVF conference on computer vision
  and pattern recognition}, pp.\  16000--16009, 2022.

\bibitem[He \& McAuley(2016)He and McAuley]{he2016ups}
Ruining He and Julian McAuley.
\newblock Ups and downs: Modeling the visual evolution of fashion trends with
  one-class collaborative filtering.
\newblock In \emph{proceedings of the 25th international conference on world
  wide web}, pp.\  507--517, 2016.

\bibitem[Hendrycks et~al.(2021)Hendrycks, Burns, Kadavath, Arora, Basart, Tang,
  Song, and Steinhardt]{hendrycksmath2021}
Dan Hendrycks, Collin Burns, Saurav Kadavath, Akul Arora, Steven Basart, Eric
  Tang, Dawn Song, and Jacob Steinhardt.
\newblock Measuring mathematical problem solving with the math dataset.
\newblock \emph{NeurIPS}, 2021.

\bibitem[Hidasi \& Karatzoglou(2018)Hidasi and
  Karatzoglou]{hidasi2018recurrent}
Bal{\'a}zs Hidasi and Alexandros Karatzoglou.
\newblock Recurrent neural networks with top-k gains for session-based
  recommendations.
\newblock In \emph{Proceedings of the 27th ACM international conference on
  information and knowledge management}, pp.\  843--852, 2018.

\bibitem[Hidasi et~al.(2016)Hidasi, Karatzoglou, Baltrunas, and
  Tikk]{hidasi2015session}
Bal{\'a}zs Hidasi, Alexandros Karatzoglou, Linas Baltrunas, and Domonkos Tikk.
\newblock Session-based recommendations with recurrent neural networks.
\newblock 2016.

\bibitem[Hoogeveen et~al.(2015)Hoogeveen, Verspoor, and Baldwin]{hoogeveen2015}
Doris Hoogeveen, Karin~M. Verspoor, and Timothy Baldwin.
\newblock Cqadupstack: A benchmark data set for community question-answering
  research.
\newblock In \emph{Proceedings of the 20th Australasian Document Computing
  Symposium (ADCS)}, ADCS '15, pp.\  3:1--3:8, New York, NY, USA, 2015. ACM.
\newblock ISBN 978-1-4503-4040-3.
\newblock \doi{10.1145/2838931.2838934}.
\newblock URL \url{http://doi.acm.org/10.1145/2838931.2838934}.

\bibitem[Hwang et~al.(2020)Hwang, Kwon, Jeong, Kim, Kang, Park, Kim, Han, Lim,
  and An]{hwang2020immune}
Sohyun Hwang, Ah-Young Kwon, Ju-Yeon Jeong, Sewha Kim, Haeyoun Kang, Joonsuk
  Park, Joo-Hang Kim, Ok~Jin Han, Sun~Min Lim, and Hee~Jung An.
\newblock Immune gene signatures for predicting durable clinical benefit of
  anti-pd-1 immunotherapy in patients with non-small cell lung cancer.
\newblock \emph{Scientific reports}, 10\penalty0 (1):\penalty0 643, 2020.

\bibitem[Ilharco et~al.(2021)Ilharco, Wortsman, Wightman, Gordon, Carlini,
  Taori, Dave, Shankar, Namkoong, Miller, Hajishirzi, Farhadi, and
  Schmidt]{ilharco2021openclip}
Gabriel Ilharco, Mitchell Wortsman, Ross Wightman, Cade Gordon, Nicholas
  Carlini, Rohan Taori, Achal Dave, Vaishaal Shankar, Hongseok Namkoong, John
  Miller, Hannaneh Hajishirzi, Ali Farhadi, and Ludwig Schmidt.
\newblock Openclip, July 2021.
\newblock URL \url{https://doi.org/10.5281/zenodo.5143773}.
\newblock If you use this software, please cite it as below.

\bibitem[Iyer et~al.(2012)Iyer, Dandekar, and Csernai]{Iyer2022first}
Shankar Iyer, Nikhil Dandekar, and Kornél Csernai.
\newblock First quora dataset release: Question pairs, 2012.
\newblock URL
  \url{https://quoradata.quora.com/First-Quora-Dataset-Release-Question-Pairs}.

\bibitem[Izacard et~al.(2021)Izacard, Caron, Hosseini, Riedel, Bojanowski,
  Joulin, and Grave]{izacard2021unsupervised}
Gautier Izacard, Mathilde Caron, Lucas Hosseini, Sebastian Riedel, Piotr
  Bojanowski, Armand Joulin, and Edouard Grave.
\newblock Unsupervised dense information retrieval with contrastive learning.
\newblock \emph{arXiv preprint arXiv:2112.09118}, 2021.

\bibitem[Javaheripi et~al.(2023)Javaheripi, Bubeck, Abdin, Aneja, Bubeck,
  Mendes, Chen, Del~Giorno, Eldan, Gopi, et~al.]{javaheripi2023phi}
Mojan Javaheripi, S{\'e}bastien Bubeck, Marah Abdin, Jyoti Aneja, Sebastien
  Bubeck, Caio C{\'e}sar~Teodoro Mendes, Weizhu Chen, Allie Del~Giorno, Ronen
  Eldan, Sivakanth Gopi, et~al.
\newblock Phi-2: The surprising power of small language models.
\newblock \emph{Microsoft Research Blog}, 1\penalty0 (3):\penalty0 3, 2023.

\bibitem[Jiang et~al.(2023)Jiang, Sablayrolles, Mensch, Bamford, Chaplot,
  de~las Casas, Bressand, Lengyel, Lample, Saulnier, Lavaud, Lachaux, Stock,
  Scao, Lavril, Wang, Lacroix, and Sayed]{jiang2023mistral7b}
Albert~Q. Jiang, Alexandre Sablayrolles, Arthur Mensch, Chris Bamford,
  Devendra~Singh Chaplot, Diego de~las Casas, Florian Bressand, Gianna Lengyel,
  Guillaume Lample, Lucile Saulnier, Lélio~Renard Lavaud, Marie-Anne Lachaux,
  Pierre Stock, Teven~Le Scao, Thibaut Lavril, Thomas Wang, Timothée Lacroix,
  and William~El Sayed.
\newblock Mistral 7b, 2023.
\newblock URL \url{https://arxiv.org/abs/2310.06825}.

\bibitem[{Joshi} et~al.(2017){Joshi}, {Choi}, {Weld}, and
  {Zettlemoyer}]{2017arXivtriviaqa}
Mandar {Joshi}, Eunsol {Choi}, Daniel {Weld}, and Luke {Zettlemoyer}.
\newblock {triviaqa: A Large Scale Distantly Supervised Challenge Dataset for
  Reading Comprehension}.
\newblock \emph{arXiv e-prints}, art. arXiv:1705.03551, 2017.

\bibitem[Kaelbling et~al.(1996)Kaelbling, Littman, and
  Moore]{kaelbling1996reinforcement}
Leslie~Pack Kaelbling, Michael~L Littman, and Andrew~W Moore.
\newblock Reinforcement learning: A survey.
\newblock \emph{Journal of artificial intelligence research}, 4:\penalty0
  237--285, 1996.

\bibitem[Kalantidis et~al.(2020)Kalantidis, Sariyildiz, Pion, Weinzaepfel, and
  Larlus]{kalantidis2020hard}
Yannis Kalantidis, Mert~Bulent Sariyildiz, Noe Pion, Philippe Weinzaepfel, and
  Diane Larlus.
\newblock Hard negative mixing for contrastive learning.
\newblock \emph{Advances in neural information processing systems},
  33:\penalty0 21798--21809, 2020.

\bibitem[Kang \& McAuley(2018)Kang and McAuley]{kang2018self}
Wang-Cheng Kang and Julian McAuley.
\newblock Self-attentive sequential recommendation.
\newblock In \emph{2018 IEEE international conference on data mining (ICDM)},
  pp.\  197--206. IEEE, 2018.

\bibitem[Karpukhin et~al.(2020)Karpukhin, O{\u{g}}uz, Min, Lewis, Wu, Edunov,
  Chen, and Yih]{karpukhin2020dense}
Vladimir Karpukhin, Barlas O{\u{g}}uz, Sewon Min, Patrick Lewis, Ledell Wu,
  Sergey Edunov, Danqi Chen, and Wen-tau Yih.
\newblock Dense passage retrieval for open-domain question answering.
\newblock \emph{arXiv preprint arXiv:2004.04906}, 2020.

\bibitem[Kocijan et~al.(2023)Kocijan, Davis, Lukasiewicz, Marcus, and
  Morgenstern]{kocijan2023defeat}
Vid Kocijan, Ernest Davis, Thomas Lukasiewicz, Gary Marcus, and Leora
  Morgenstern.
\newblock The defeat of the winograd schema challenge.
\newblock \emph{Artificial Intelligence}, 325:\penalty0 103971, 2023.

\bibitem[Kwiatkowski et~al.(2019)Kwiatkowski, Palomaki, Redfield, Collins,
  Parikh, Alberti, Epstein, Polosukhin, Kelcey, Devlin, Lee, Toutanova, Jones,
  Chang, Dai, Uszkoreit, Le, and Petrov]{Kwiatkowski2019NaturalQ}
Tom Kwiatkowski, Jennimaria Palomaki, Olivia Redfield, Michael Collins, Ankur
  Parikh, Chris Alberti, Danielle Epstein, Illia Polosukhin, Matthew Kelcey,
  Jacob Devlin, Kenton Lee, Kristina~N. Toutanova, Llion Jones, Ming-Wei Chang,
  Andrew Dai, Jakob Uszkoreit, Quoc Le, and Slav Petrov.
\newblock Natural questions: a benchmark for question answering research.
\newblock \emph{Transactions of the Association of Computational Linguistics},
  2019.

\bibitem[Lee et~al.(2019)Lee, Chang, and Toutanova]{lee2019latent}
Kenton Lee, Ming-Wei Chang, and Kristina Toutanova.
\newblock Latent retrieval for weakly supervised open domain question
  answering.
\newblock In \emph{Proceedings of the 57th Annual Meeting of the Association
  for Computational Linguistics}, pp.\  6086--6096, 01 2019.
\newblock \doi{10.18653/v1/P19-1612}.

\bibitem[Legg \& Hutter(2007)Legg and Hutter]{legg2007universal}
Shane Legg and Marcus Hutter.
\newblock Universal intelligence: A definition of machine intelligence.
\newblock \emph{Minds and machines}, 17:\penalty0 391--444, 2007.

\bibitem[Levesque et~al.(2012)Levesque, Davis, and
  Morgenstern]{levesque2012winograd}
Hector Levesque, Ernest Davis, and Leora Morgenstern.
\newblock The winograd schema challenge.
\newblock In \emph{Thirteenth international conference on the principles of
  knowledge representation and reasoning}, 2012.

\bibitem[Lin et~al.(2014)Lin, Maire, Belongie, Hays, Perona, Ramanan,
  Doll{\'a}r, and Zitnick]{lin2014microsoft}
Tsung-Yi Lin, Michael Maire, Serge Belongie, James Hays, Pietro Perona, Deva
  Ramanan, Piotr Doll{\'a}r, and C~Lawrence Zitnick.
\newblock Microsoft coco: Common objects in context.
\newblock In \emph{Computer Vision--ECCV 2014: 13th European Conference,
  Zurich, Switzerland, September 6-12, 2014, Proceedings, Part V 13}, pp.\
  740--755. Springer, 2014.

\bibitem[Maia et~al.(2018)Maia, Handschuh, Freitas, Davis, McDermott, Zarrouk,
  and Balahur]{Maia2018FiQA}
Macedo Maia, Siegfried Handschuh, Andr\'{e} Freitas, Brian Davis, Ross
  McDermott, Manel Zarrouk, and Alexandra Balahur.
\newblock Www'18 open challenge: Financial opinion mining and question
  answering.
\newblock In \emph{Companion Proceedings of the The Web Conference 2018}, WWW
  '18, pp.\  1941–1942, Republic and Canton of Geneva, CHE, 2018.
  International World Wide Web Conferences Steering Committee.
\newblock ISBN 9781450356404.
\newblock \doi{10.1145/3184558.3192301}.
\newblock URL \url{https://doi.org/10.1145/3184558.3192301}.

\bibitem[Malamud et~al.(1998)Malamud, Morein, and Turcotte]{malamud1998forest}
Bruce~D Malamud, Gleb Morein, and Donald~L Turcotte.
\newblock Forest fires: an example of self-organized critical behavior.
\newblock \emph{Science}, 281\penalty0 (5384):\penalty0 1840--1842, 1998.

\bibitem[Mart{\'\i}nez-Jim{\'e}nez et~al.(2023)Mart{\'\i}nez-Jim{\'e}nez,
  Priestley, Shale, Baber, Rozemuller, and Cuppen]{martinez2023genetic}
Francisco Mart{\'\i}nez-Jim{\'e}nez, Peter Priestley, Charles Shale, Jonathan
  Baber, Erik Rozemuller, and Edwin Cuppen.
\newblock Genetic immune escape landscape in primary and metastatic cancer.
\newblock \emph{Nature Genetics}, 55\penalty0 (5):\penalty0 820--831, 2023.

\bibitem[Mei et~al.(2024)Mei, Xie, Yuan, and Jackson]{mei2024turing}
Qiaozhu Mei, Yutong Xie, Walter Yuan, and Matthew~O Jackson.
\newblock A turing test of whether ai chatbots are behaviorally similar to
  humans.
\newblock \emph{Proceedings of the National Academy of Sciences}, 121\penalty0
  (9):\penalty0 e2313925121, 2024.

\bibitem[Morris et~al.(2024)Morris, Sohl-dickstein, Fiedel, Warkentin, Dafoe,
  Faust, Farabet, and Legg]{morris2024levels}
Meredith~Ringel Morris, Jascha Sohl-dickstein, Noah Fiedel, Tris Warkentin,
  Allan Dafoe, Aleksandra Faust, Clement Farabet, and Shane Legg.
\newblock Levels of agi for operationalizing progress on the path to agi, 2024.
\newblock URL \url{https://arxiv.org/abs/2311.02462}.

\bibitem[Naudet(2024)]{HFforLegal2024}
Louis~Brulé Naudet.
\newblock The laws, centralizing legal texts for better use.
\newblock \url{https://huggingface.co/datasets/HFforLegal/laws}, 2024.

\bibitem[Nogueira \& Cho(2019)Nogueira and Cho]{nogueira2019passage}
Rodrigo Nogueira and Kyunghyun Cho.
\newblock Passage re-ranking with bert.
\newblock \emph{arXiv preprint arXiv:1901.04085}, 2019.

\bibitem[Oord et~al.(2018)Oord, Li, and Vinyals]{oord2018representation}
Aaron van~den Oord, Yazhe Li, and Oriol Vinyals.
\newblock Representation learning with contrastive predictive coding.
\newblock \emph{arXiv preprint arXiv:1807.03748}, 2018.

\bibitem[{OpenAI}(2018)]{openAICharter}
{OpenAI}.
\newblock {OpenAI Charter}, 2018.
\newblock URL \url{https://openai.com/charter}.
\newblock Accessed February 24, 2025.

\bibitem[Phillips(2014)]{phillips2014fractals}
JC~Phillips.
\newblock Fractals and self-organized criticality in proteins.
\newblock \emph{Physica A: Statistical Mechanics and Its Applications},
  415:\penalty0 440--448, 2014.

\bibitem[Plummer et~al.(2015)Plummer, Wang, Cervantes, Caicedo, Hockenmaier,
  and Lazebnik]{plummer2015flickr30k}
Bryan~A Plummer, Liwei Wang, Chris~M Cervantes, Juan~C Caicedo, Julia
  Hockenmaier, and Svetlana Lazebnik.
\newblock Flickr30k entities: Collecting region-to-phrase correspondences for
  richer image-to-sentence models.
\newblock In \emph{Proceedings of the IEEE international conference on computer
  vision}, pp.\  2641--2649, 2015.

\bibitem[Radford et~al.(2019)Radford, Wu, Child, Luan, Amodei, and
  Sutskever]{radford2019language}
Alec Radford, Jeff Wu, Rewon Child, David Luan, Dario Amodei, and Ilya
  Sutskever.
\newblock Language models are unsupervised multitask learners.
\newblock 2019.

\bibitem[Radford et~al.(2021)Radford, Kim, Hallacy, Ramesh, Goh, Agarwal,
  Sastry, Askell, Mishkin, Clark, et~al.]{radford2021learning}
Alec Radford, Jong~Wook Kim, Chris Hallacy, Aditya Ramesh, Gabriel Goh,
  Sandhini Agarwal, Girish Sastry, Amanda Askell, Pamela Mishkin, Jack Clark,
  et~al.
\newblock Learning transferable visual models from natural language
  supervision.
\newblock In \emph{International conference on machine learning}, pp.\
  8748--8763. PMLR, 2021.

\bibitem[Raffel et~al.(2020)Raffel, Shazeer, Roberts, Lee, Narang, Matena,
  Zhou, Li, and Liu]{raffel2020exploring}
Colin Raffel, Noam Shazeer, Adam Roberts, Katherine Lee, Sharan Narang, Michael
  Matena, Yanqi Zhou, Wei Li, and Peter~J Liu.
\newblock Exploring the limits of transfer learning with a unified text-to-text
  transformer.
\newblock \emph{Journal of machine learning research}, 21\penalty0
  (140):\penalty0 1--67, 2020.

\bibitem[Reimers \& Gurevych(2019)Reimers and Gurevych]{reimers2019sentence}
Nils Reimers and Iryna Gurevych.
\newblock Sentence-bert: Sentence embeddings using siamese bert-networks.
\newblock In \emph{Proceedings of the 2019 Conference on Empirical Methods in
  Natural Language Processing}. Association for Computational Linguistics, 11
  2019.
\newblock URL \url{https://arxiv.org/abs/1908.10084}.

\bibitem[Robertson \& Jones(1976)Robertson and Jones]{robertson1976relevance}
Stephen~E Robertson and K~Sparck Jones.
\newblock Relevance weighting of search terms.
\newblock \emph{Journal of the American Society for Information science},
  27\penalty0 (3):\penalty0 129--146, 1976.

\bibitem[Searle(1999)]{searle1999chinese}
John Searle.
\newblock The chinese room, 1999.

\bibitem[Shrivastava et~al.(2016)Shrivastava, Gupta, and
  Girshick]{shrivastava2016training}
Abhinav Shrivastava, Abhinav Gupta, and Ross Girshick.
\newblock Training region-based object detectors with online hard example
  mining.
\newblock In \emph{Proceedings of the IEEE conference on computer vision and
  pattern recognition}, pp.\  761--769, 2016.

\bibitem[Suleyman(2023)]{suleyman2023coming}
Mustafa Suleyman.
\newblock \emph{The coming wave: technology, power, and the twenty-first
  century's greatest dilemma}.
\newblock Crown, 2023.

\bibitem[Touvron et~al.(2023{\natexlab{a}})Touvron, Lavril, Izacard, Martinet,
  Lachaux, Lacroix, Rozi{\`e}re, Goyal, Hambro, Azhar,
  et~al.]{touvron2023llama1}
Hugo Touvron, Thibaut Lavril, Gautier Izacard, Xavier Martinet, Marie-Anne
  Lachaux, Timoth{\'e}e Lacroix, Baptiste Rozi{\`e}re, Naman Goyal, Eric
  Hambro, Faisal Azhar, et~al.
\newblock Llama: Open and efficient foundation language models.
\newblock \emph{arXiv preprint arXiv:2302.13971}, 2023{\natexlab{a}}.

\bibitem[Touvron et~al.(2023{\natexlab{b}})Touvron, Martin, Stone, Albert,
  Almahairi, Babaei, Bashlykov, Batra, Bhargava, Bhosale,
  et~al.]{touvron2023llama2}
Hugo Touvron, Louis Martin, Kevin Stone, Peter Albert, Amjad Almahairi, Yasmine
  Babaei, Nikolay Bashlykov, Soumya Batra, Prajjwal Bhargava, Shruti Bhosale,
  et~al.
\newblock Llama 2: Open foundation and fine-tuned chat models.
\newblock \emph{arXiv preprint arXiv:2307.09288}, 2023{\natexlab{b}}.

\bibitem[Turcotte et~al.(1985)Turcotte, Smalley~Jr, and
  Solla]{turcotte1985collapse}
DL~Turcotte, RF~Smalley~Jr, and Sara~A Solla.
\newblock Collapse of loaded fractal trees.
\newblock \emph{Nature}, 313\penalty0 (6004):\penalty0 671--672, 1985.

\bibitem[Turing(1950)]{turing1950computing}
A.~M. Turing.
\newblock Computing machinery and intelligence.
\newblock \emph{Mind}, 59\penalty0 (236):\penalty0 433--460, 1950.
\newblock ISSN 00264423.
\newblock URL \url{http://www.jstor.org/stable/2251299}.

\bibitem[Wang(2023)]{wang2023chinese}
Tao Wang.
\newblock Chinese law and regulations.
\newblock
  \url{https://huggingface.co/datasets/twang2218/chinese-law-and-regulations},
  2023.

\bibitem[Wang \& Isola(2020)Wang and Isola]{wang2020understanding}
Tongzhou Wang and Phillip Isola.
\newblock Understanding contrastive representation learning through alignment
  and uniformity on the hypersphere.
\newblock In \emph{International conference on machine learning}, pp.\
  9929--9939. PMLR, 2020.

\bibitem[Wang et~al.(2024)Wang, Ma, Zhang, Ni, Chandra, Guo, Ren, Arulraj, He,
  Jiang, et~al.]{wang2024mmlu}
Yubo Wang, Xueguang Ma, Ge~Zhang, Yuansheng Ni, Abhranil Chandra, Shiguang Guo,
  Weiming Ren, Aaran Arulraj, Xuan He, Ziyan Jiang, et~al.
\newblock Mmlu-pro: A more robust and challenging multi-task language
  understanding benchmark.
\newblock \emph{arXiv preprint arXiv:2406.01574}, 2024.

\bibitem[Wei et~al.(2023)Wei, Luan, Liu, Dong, and Wang]{wei2023cmath}
Tianwen Wei, Jian Luan, Wei Liu, Shuang Dong, and Bin Wang.
\newblock Cmath: Can your language model pass chinese elementary school math
  test?, 2023.

\bibitem[Xiao et~al.(2023)Xiao, Liu, Zhang, and Muennighoff]{bge_embedding}
Shitao Xiao, Zheng Liu, Peitian Zhang, and Niklas Muennighoff.
\newblock C-pack: Packaged resources to advance general chinese embedding,
  2023.

\bibitem[Xie et~al.(2023)Xie, Dong, Wang, Lv, Yao, Gan, Wu, Li, Li, Liu,
  et~al.]{xie2023t2ranking}
Xiaohui Xie, Qian Dong, Bingning Wang, Feiyang Lv, Ting Yao, Weinan Gan,
  Zhijing Wu, Xiangsheng Li, Haitao Li, Yiqun Liu, et~al.
\newblock T2ranking: A large-scale chinese benchmark for passage ranking.
\newblock In \emph{Proceedings of the 46th International ACM SIGIR Conference
  on Research and Development in Information Retrieval}, pp.\  2681--2690,
  2023.

\bibitem[Xiong et~al.(2021)Xiong, Xiong, Li, Tang, Liu, Bennett, Ahmed, and
  Overwijk]{xiong2021approximate}
Lee Xiong, Chenyan Xiong, Ye~Li, Kwok-Fung Tang, Jialin Liu, Paul~N. Bennett,
  Junaid Ahmed, and Arnold Overwijk.
\newblock Approximate nearest neighbor negative contrastive learning for dense
  text retrieval.
\newblock In \emph{International Conference on Learning Representations}, 2021.
\newblock URL \url{https://openreview.net/forum?id=zeFrfgyZln}.

\bibitem[Yang et~al.(2024{\natexlab{a}})Yang, Yang, Hui, Zheng, Yu, Zhou, Li,
  Li, Liu, Huang, Dong, Wei, Lin, Tang, Wang, Yang, Tu, Zhang, Ma, Xu, Zhou,
  Bai, He, Lin, Dang, Lu, Chen, Yang, Li, Xue, Ni, Zhang, Wang, Peng, Men, Gao,
  Lin, Wang, Bai, Tan, Zhu, Li, Liu, Ge, Deng, Zhou, Ren, Zhang, Wei, Ren, Fan,
  Yao, Zhang, Wan, Chu, Liu, Cui, Zhang, and Fan]{qwen2}
An~Yang, Baosong Yang, Binyuan Hui, Bo~Zheng, Bowen Yu, Chang Zhou, Chengpeng
  Li, Chengyuan Li, Dayiheng Liu, Fei Huang, Guanting Dong, Haoran Wei, Huan
  Lin, Jialong Tang, Jialin Wang, Jian Yang, Jianhong Tu, Jianwei Zhang,
  Jianxin Ma, Jin Xu, Jingren Zhou, Jinze Bai, Jinzheng He, Junyang Lin, Kai
  Dang, Keming Lu, Keqin Chen, Kexin Yang, Mei Li, Mingfeng Xue, Na~Ni, Pei
  Zhang, Peng Wang, Ru~Peng, Rui Men, Ruize Gao, Runji Lin, Shijie Wang, Shuai
  Bai, Sinan Tan, Tianhang Zhu, Tianhao Li, Tianyu Liu, Wenbin Ge, Xiaodong
  Deng, Xiaohuan Zhou, Xingzhang Ren, Xinyu Zhang, Xipin Wei, Xuancheng Ren,
  Yang Fan, Yang Yao, Yichang Zhang, Yu~Wan, Yunfei Chu, Yuqiong Liu, Zeyu Cui,
  Zhenru Zhang, and Zhihao Fan.
\newblock Qwen2 technical report.
\newblock \emph{arXiv preprint arXiv:2407.10671}, 2024{\natexlab{a}}.

\bibitem[Yang et~al.(2024{\natexlab{b}})Yang, Yang, Zhang, Hui, Zheng, Yu, Li,
  Liu, Huang, Wei, Lin, Yang, Tu, Zhang, Yang, Yang, Zhou, Lin, Dang, Lu, Bao,
  Yang, Yu, Li, Xue, Zhang, Zhu, Men, Lin, Li, Xia, Ren, Ren, Fan, Su, Zhang,
  Wan, Liu, Cui, Zhang, and Qiu]{qwen2.5}
An~Yang, Baosong Yang, Beichen Zhang, Binyuan Hui, Bo~Zheng, Bowen Yu,
  Chengyuan Li, Dayiheng Liu, Fei Huang, Haoran Wei, Huan Lin, Jian Yang,
  Jianhong Tu, Jianwei Zhang, Jianxin Yang, Jiaxi Yang, Jingren Zhou, Junyang
  Lin, Kai Dang, Keming Lu, Keqin Bao, Kexin Yang, Le~Yu, Mei Li, Mingfeng Xue,
  Pei Zhang, Qin Zhu, Rui Men, Runji Lin, Tianhao Li, Tingyu Xia, Xingzhang
  Ren, Xuancheng Ren, Yang Fan, Yang Su, Yichang Zhang, Yu~Wan, Yuqiong Liu,
  Zeyu Cui, Zhenru Zhang, and Zihan Qiu.
\newblock Qwen2.5 technical report.
\newblock \emph{arXiv preprint arXiv:2412.15115}, 2024{\natexlab{b}}.

\bibitem[Yang et~al.(2017)Yang, Fang, and Lin]{yang2017anserini}
Peilin Yang, Hui Fang, and Jimmy Lin.
\newblock Anserini: Enabling the use of lucene for information retrieval
  research.
\newblock In \emph{Proceedings of the 40th international ACM SIGIR conference
  on research and development in information retrieval}, pp.\  1253--1256,
  2017.

\bibitem[Zachariou et~al.(2015)Zachariou, Expert, Takayasu, and
  Christensen]{zachariou2015generalised}
Nicky Zachariou, Paul Expert, Misako Takayasu, and Kim Christensen.
\newblock Generalised sandpile dynamics on artificial and real-world directed
  networks.
\newblock \emph{PloS One}, 10\penalty0 (11):\penalty0 e0142685, 2015.

\bibitem[Zhan et~al.(2021)Zhan, Mao, Liu, Guo, Zhang, and
  Ma]{zhan2021optimizing}
Jingtao Zhan, Jiaxin Mao, Yiqun Liu, Jiafeng Guo, Min Zhang, and Shaoping Ma.
\newblock Optimizing dense retrieval model training with hard negatives.
\newblock In \emph{Proceedings of the 44th International ACM SIGIR Conference
  on Research and Development in Information Retrieval}, pp.\  1503--1512,
  2021.

\bibitem[Zhang et~al.(2022)Zhang, Roller, Goyal, Artetxe, Chen, Chen, Dewan,
  Diab, Li, Lin, et~al.]{zhang2022opt}
Susan Zhang, Stephen Roller, Naman Goyal, Mikel Artetxe, Moya Chen, Shuohui
  Chen, Christopher Dewan, Mona Diab, Xian Li, Xi~Victoria Lin, et~al.
\newblock Opt: Open pre-trained transformer language models.
\newblock \emph{arXiv preprint arXiv:2205.01068}, 2022.

\bibitem[Zhang et~al.(2019)Zhang, Miao, Qian, Wang, Qi, Zhang, Dang, Wu, and
  Wang]{zhang2019molecular}
Zhen-Ye Zhang, Ling-Feng Miao, Ling-Ling Qian, Ning Wang, Miao-Miao Qi, Yu-Min
  Zhang, Shi-Peng Dang, Ying Wu, and Ru-Xing Wang.
\newblock Molecular mechanisms of glucose fluctuations on diabetic
  complications.
\newblock \emph{Frontiers in endocrinology}, 10:\penalty0 640, 2019.

\bibitem[Zhu et~al.(2019)Zhu, Chen, Wang, Liu, and Zheng]{zhu2019dtcdr}
Feng Zhu, Chaochao Chen, Yan Wang, Guanfeng Liu, and Xiaolin Zheng.
\newblock Dtcdr: A framework for dual-target cross-domain recommendation.
\newblock In \emph{Proceedings of the 28th ACM international conference on
  information and knowledge management}, pp.\  1533--1542, 2019.

\bibitem[Zhu et~al.(2020)Zhu, Wang, Chen, Liu, and Zheng]{zhu2020graphical}
Feng Zhu, Yan Wang, Chaochao Chen, Guanfeng Liu, and Xiaolin Zheng.
\newblock A graphical and attentional framework for dual-target cross-domain
  recommendation.
\newblock In \emph{IJCAI}, volume~21, pp.\ ~39, 2020.

\end{thebibliography}

\end{document}